\documentclass{article}

\PassOptionsToPackage{numbers, sort&compress}{natbib}

\usepackage[final]{neurips_2023}

\usepackage{amsmath,amsfonts,bm}

\def\eqref#1{equation~\ref{#1}}

\def\1{\bm{1}}

\DeclareMathAlphabet{\mathsfit}{\encodingdefault}{\sfdefault}{m}{sl}
\SetMathAlphabet{\mathsfit}{bold}{\encodingdefault}{\sfdefault}{bx}{n}

\newcommand{\E}{\texorpdfstring{\mathbb{E}}{ℰ}}

\DeclareMathOperator*{\argmin}{arg\,min}

\usepackage[T1]{fontenc}    
\usepackage{nicefrac}       
\usepackage{microtype}      
\usepackage{xcolor}         

\usepackage{hyperref}
\usepackage{url}

\usepackage{microtype}
\usepackage{graphicx}
\usepackage{subfigure}
\usepackage{booktabs} 
\usepackage{caption}
\usepackage{amsmath}
\usepackage{amssymb}
\usepackage{mathtools}
\usepackage{amsthm}
\usepackage{soul}
\usepackage{paralist}
\usepackage{algorithm}
\usepackage{algpseudocode}
\usepackage[capitalize,noabbrev]{cleveref}
\usepackage[acronym]{glossaries}
\usepackage[utf8]{inputenc}

\usepackage{enumitem}
\usepackage{wrapfig}

\usepackage{todonotes}
\newcommand{\note}[1]{\todo[inline]{#1}}
\renewcommand{\note}[1]{}  

\newcommand{\codeURL}
{\url{https://github.com/automl/mf-prior-exp}}

\newcommand{\algo}{\texttt{PriorBand}} 

\newcommand{\mfpbench}{\texttt{mf-prior-bench}\ }
\newcommand{\yahpogym}{Yahpo-Gym\ }
\newcommand{\jahsbench}{JAHS-Bench-201\ }
\newcommand{\hyperbo}{HyperBO\ }
\newcommand{\nan}{\texttt{NaN}\ }
\newcommand{\pibo}{$\pi$BO}

\newcommand{\hp}{\text{HP}}
\newcommand{\hps}{\text{HPs}}
\newcommand{\dl}{\text{DL}}

\newcommand{\hpo}{\text{HPO}}

\newcommand{\hb}{\text{HB}} 
\newcommand{\hblong}{\texttt{HyperBand}} 
\newcommand{\sh}{\text{SH}}
\newcommand{\shlong}{\texttt{SuccessiveHalving}}
\newcommand{\bo}{\text{BO}}
\newcommand{\bolong}{\texttt{Bayesian Optimization}}
\newcommand{\bohb}{\text{BOHB}}
\newcommand{\asha}{\text{ASHA}}
\newcommand{\mobster}{\text{Mobster}}
\newcommand{\esp}{\text{ESP}}

\def \wi {\texorpdfstring{$w_{\hat{\lambda}}$}{wₓ}}
\def \pinc {\texorpdfstring{$p_{\hat{\lambda}}$}{wₓ}}
\def \inc {\texorpdfstring{$\hat{\lambda}(\cdot)$}{x*(·)}}

\def \E {\texorpdfstring{$\mathcal{E}_{\pi}$}{ℰπ}}

\def \pu {\texorpdfstring{$p_\mathcal{U}$}{wᵤ}}
\def \pp {$p_\pi$}

\def \config {\bm{\lambda}}
\def \cspace {\Lambda}
\def \fid {\bm{z}}
\def \min { _\text{min} }
\def \max { _\text{max} }

\def \prior {$\pi(\cdot)$}
\def \uniform {$\mathcal{U}(\cdot)$}

\def \model {$\mathcal{M}(\cdot)$}

\def \state {$s_t$}

\def \actionset {$\mathcal{A}$}

\def \rung {$r$}

\def \topeta {$\text{top-performing }(1/\eta$)}

\theoremstyle{plain}

\theoremstyle{definition}

\theoremstyle{remark}

\newcommand{\xmark}[0]{\textcolor{red}{\textbf{\large\texttimes}}}

\title{PriorBand: Practical Hyperparameter Optimization in the Age of Deep Learning}

\author{%
    Neeratyoy Mallik \\
    University of Freiburg\\
    \href{mailto:mallik@cs.uni-freiburg.de}{\texttt{mallik@cs.uni-freiburg.de}} 
    \And 
    Edward Bergman\\
    University of Freiburg\\
    \href{mailto:bergmane@cs.uni-freiburg.de}{\texttt{bergmane@cs.uni-freiburg.de}} 
    \And 
    Carl Hvarfner\\
    Lund University\\
    \href{mailto:carl.hvarfner@cs.lth.se}{\texttt{carl.hvarfner@cs.lth.se}}  
    \And 
    Danny Stoll\\
    University of Freiburg\\
    \href{mailto:stolld@cs.uni-freiburg.de}{\texttt{stolld@cs.uni-freiburg.de}}
    \And 
    Maciej Janowski\\
    University of Freiburg\\
    \href{mailto:janowski@cs.uni-freiburg.de}{\texttt{janowski@cs.uni-freiburg.de}}
    \And 
    Marius Lindauer\\
    Leibniz University Hannover\\
    \href{mailto:m.lindauer@ai.uni-hannover.de}{\texttt{m.lindauer@ai.uni-hannover.de}}
    \And 
    Luigi Nardi\\
    Lund University\\
    Stanford University\\
    DBtune\\
    \href{mailto:luigi.nardi@cs.lth.se}{\texttt{luigi.nardi@cs.lth.se}}
    \And 
    Frank Hutter \\
    University of Freiburg\\
    \href{mailto:fh@cs.uni-freiburg.de}{\texttt{fh@cs.uni-freiburg.de}}
}

\begin{document}

\maketitle

\begin{abstract}
Hyperparameters of Deep Learning (\dl{}) pipelines are crucial for their downstream performance. 
While a large number of methods for Hyperparameter Optimization (\hpo{}) have been developed, their incurred costs are often untenable for modern \dl{}.
Consequently, manual experimentation is still the most prevalent approach to optimize hyperparameters, relying on the researcher's intuition, domain knowledge, and cheap preliminary explorations.
To resolve this misalignment between \hpo{} algorithms and \dl{} researchers, we propose \algo{}, an \hpo{} algorithm tailored to \dl{}, able to utilize both expert beliefs and cheap proxy tasks. 
Empirically, we demonstrate \algo{}'s efficiency across a range of \dl{} benchmarks and show its gains under informative expert input and robustness against poor expert beliefs. 
\end{abstract}


\section{Introduction} \label{sec:intro}


The performance of Deep Learning (\dl{}) models crucially depends on dataset-specific settings of their hyperparameters (\hps{})~\citep{goodfellow-16a,loshchilov-iclr19a}. 
Therefore, Hyperparameter Optimization (\hpo{}) is an integral step in the development of \dl{} models \citep{chen-arxiv18a, melis-iclr18a,zhang-aistats21a,wightman-neurips21a}.
\hpo{} methods 
have typically been applied to traditional machine learning models (including shallow neural networks) that focus on fairly small datasets~\citep{snoek-nips12a, feurer-jmlr22a}. 
Current \dl{} practitioners, however, often utilize much larger models and datasets (e.g., a single training of GPT-3~\citep{brown-neurips20a} was estimated to require $3\cdot10^{23}$ FLOPS, i.e., months on a thousand V100 GPUs~\citep{LambdaLabs}).
As recent \hpo{} practices continue to apply simple techniques like grid, random, or manual search~\citep{bouthillier-hal20a, hasebrook-arxiv22a, schneider-hity2022a, vanderblom-automl21a}, existing \hpo{} methods seem misaligned with DL practice.


To make the misalignment between existing HPO methods and DL practice explicit, we identify the following desiderata for an efficient, scalable \hpo{} method suitable for current DL: 
\begin{enumerate} [itemsep=-0.005mm]
    \item \textbf{Strong performance under low compute budgets}: As \dl{} pipelines become larger and more complex, only a few model training can be afforded to find performant configurations.
    \item \textbf{Integrate cheap proxy tasks}: To save resources, cheap evaluations must be used to find promising configurations early while maintaining robustness to uninformative proxies.
    \item \textbf{Integrate expert beliefs}: \dl{} models often come with a competitive default setting or an expert may have prior beliefs on what settings to consider for \hpo{}. Incorporating this information is essential while maintaining robustness to such beliefs being inaccurate.
    \item \textbf{Handle mixed type search spaces}: Modern DL pipelines tend to be composed of many kinds of hyperparameters,
    e.g. categorical, numerical, and log hyperparameters.
    \item \textbf{Simple to understand and implement}: The HPO method should be easy to apply to different \dl{} pipelines and the behavior should be conceptually easy to understand.
    \item \textbf{Parallelism}: Scaling to modern compute resources must be simple and effective.
\end{enumerate}
%

\begin{table}[t] 
\small
\caption[Caption for LOF]{
Comparison with respect to identified desiderata for DL pipelines. 
Shown is our algorithm, PriorBand, along with Grid Search, multi-fidelity random search (MF-RS), e.g. \hblong{}, \asha{}, \bolong{} (\bo{}), \bo{} with Expert Priors (\pibo{}), multi-fidelity \bo{} (MF-BO), e.g. \bohb{}, \mobster{}. A checkmark and red cross indicates whether they satisfy the corresponding desideratum. Parenthesized checkmark indicates the desideratum is partly fulfilled, or requires additional modifications to be fulfilled. Model-based methods require customized kernels for discrete hyperparameters and asynchronous parallel approaches~\cite{pmlr-v51-gonzalez16a, snoek-nips12a} to achieve speed-up under (asynchronous) parallelism. Multi-fidelity algorithms' final performance is contingent on the budget spent on HPO.
}
\label{table:desiderata}
\begin{tabular}{ p{3.5cm}cccccc }
\toprule
Criterion & Grid Search & MF-RS & BO & BO with Priors & MF-BO & PriorBand  \\
\midrule
Good anytime performance & \xmark & \checkmark & \xmark  & \xmark & \checkmark & \checkmark \\
Good final performance & \xmark & (\checkmark) & \checkmark  & \checkmark & \checkmark & \checkmark \\
Proxy task integration & \xmark & \checkmark & \xmark  & \xmark & \checkmark & \checkmark \\
Integrate expert beliefs & \xmark & \xmark & \xmark  & \checkmark & \xmark & \checkmark \\
Mixed search spaces & \checkmark & \checkmark & (\checkmark)  & (\checkmark) & (\checkmark) & \checkmark \\
Model free & \checkmark & \checkmark & \xmark  & \xmark & \xmark & \checkmark \\
Speedup under parallelism & \checkmark & \checkmark & (\checkmark) & (\checkmark)  & (\checkmark) & \checkmark \\
\bottomrule
\end{tabular}
\end{table}

\begin{figure*}
    \centering
    \begin{tabular}{c}
        \includegraphics[width=0.9\textwidth]{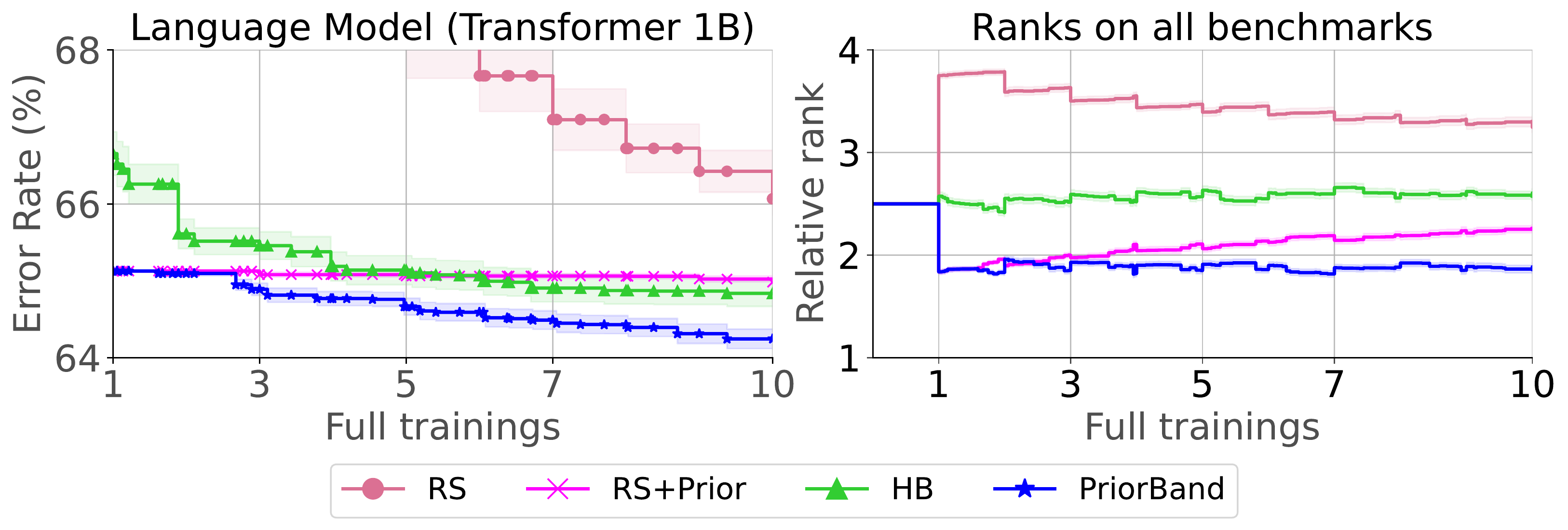}
    \end{tabular}
    \caption{
        Both plots compare Random Search (RS), sampling from prior (RS+Prior), HyperBand (HB), and our method \algo{} which utilizes a \textit{good} prior as defined by an expert.
        [\textbf{Left}] Tuning a large transformer on the 1B word benchmark. \algo{} leverages the prior, achieving strong anytime performance. [\textbf{Right}] Ranks on all our 12 benchmarks; \algo{} consistently ranks best.
    }
    \label{fig:contributions-plot}
\end{figure*}
Existing HPO algorithms satisfy subsets of these desiderata (see Table~\ref{table:desiderata}). 
For example, methods that utilize cheaper proxy tasks exploit training data subsets~\citep{klein-aistats17a}, fewer epochs/updates~\citep{swersky-arxiv14a,domhan-ijcai15a},
and proxy performance predictors~\citep{white-neurips21a,abdelfattah-iclr21a}.
Previous works also consider how to combine expert knowledge into the optimization procedure in the form of the optimal value~\citep{nguyen-icml20a}, meta or transfer learned knowledge~\citep{feurer-aaai15a,perrone-neurips19a,feurer-arxiv22a,wistuba-iclr21a} or configurations and regions that are known to the expert to have worked well previously~\citep{stoll-neurips20a,souza-ecmlpkdd21a,hvarfner-iclr22a}.
Nonetheless, no \hpo{} method exists that meets all of the desiderata above, especially, a simple model-free approach that can run cheaper proxy tasks, letting the DL expert integrate their domain knowledge for \hpo{}.

We propose \algo{}, the first method to fulfill all desiderata. 
Our \textbf{contributions} are as follows:
\begin{enumerate}
    \item We are the first to develop an approach to leverage cheap proxy tasks with an expert prior input~(Section~\ref{sec:problem_statement}), and we show the need beyond a naive solution to do so (Section~\ref{sec:beyond-naive}).
    \item We contribute a general \hpo{} algorithm with \algo{}, fulfilling all desiderata (Table~\ref{table:desiderata}) required for application to \dl{} (Section~\ref{sec:method}; also see Figure~\ref{fig:contributions-plot})
    \item We demonstrate the efficiency and robustness of \algo{} on a wide suite of \dl{} tasks under practically feasible compute resources (Section~\ref{sec:results}). 
    \item We highlight the flexibility of the method behind \algo{} by improving different multi-fidelity algorithms and their model-based extensions (Section~\ref{sec:exp-generality} \& ~\ref{sec:exp-model}), which highlights the flexible, modular design and possible extensions of \algo{}.
\end{enumerate}

We conclude, highlight limitations, and discuss further research directions in Section~\ref{sec:conclusion}.
Our code for reproducing the experiments is open-sourced at \codeURL. 




\section{Problem statement} \label{sec:problem_statement}

We consider the problem of \textit{minimizing} an expensive-to-evaluate objective function $f: \cspace \to \mathbb{R}$, i.e., \mbox{$\config^* \in \argmin_{\config \in \cspace} f(\config)$}, where a configuration $\config$ comes from a search space $\cspace$ which may consist of any combination of continuous, discrete and categorical variables. 
In \hpo{} particularly, $f$ refers to training and validating some model based on hyperparameters defined by $\config$.

Multi-fidelity optimization of $f(\config)$ requires a proxy function, namely $\hat{f}(\config, \fid)$, that provides a cheap approximation of $f$ at a given fidelity $\fid$, e.g., the validation loss of a model only trained for $\fid$ epochs. Our methodology considers a
fidelity scale $\fid \in [\fid\min, \fid\max]$ such that evaluating at $\fid\max$ coincides with our true objective: $f(\config) = \hat{f}(\config, \fid\max)$.

To take all desiderata into account, we additionally include a user-specified belief $\pi: \cspace \to \mathbb{R}$, where $\pi (\config) = \mathbb{P}\left(\config = \config{}^*\right)$ represents a subjective probability that a given configuration $\config$ is optimal. 
Thus, the overall objective is
\begin{equation} \label{eq:mf-prior-obj}
    \config^* \in \argmin_{\config \in \cspace} \hat{f}(\config, \fid\max), \quad \text{guided by } \pi(\config).
\end{equation} 
Our problem is to efficiently solve Equation~\ref{eq:mf-prior-obj} under a constrained budget, e.g., $10$ full trainings of a \dl{} model. 
 Given DL training can diverge, the \textit{incumbent} we return is the configuration-fidelity pair ($\config, \fid$) that achieved the lowest error while optimizing for Equation~\ref{eq:mf-prior-obj}. 

\section{Background}
\label{sec:bg}
While we are the first to target Equation~\ref{eq:mf-prior-obj} in its full form, below, we introduce the necessary background on methods that use either multi-fidelity optimization \emph{or} expert beliefs.

\textbf{Successive Halving (\sh{})} 
optimizes Equation~\ref{eq:mf-prior-obj} (sans $\pi$) as a best-arm identification problem in a multi-armed bandit setting~\citep{jamieson-aistats16a}. 
Given a lower and upper fidelity bound ($[\fid\min, \fid\max]$) and a reduction factor ($\eta$), \sh{} discretizes the fidelity range geometrically with a log-factor of $\eta$. 
For example, if the inputs to \sh{} are $\fid \in [3, 81]$ and $\eta = 3$, the fidelity levels in \sh{} are $\fid=\{ 3, 9, 27, 81 \}$ with accompanying enumeration called rungs, $\rung{}=\{0, 1, 2, 3\}$. 
Any $\fid < \fid\max$ is called a lower fidelity, a cheaper proxy task in \dl{} training.
Every iteration of each round of \sh{} involves uniform random sampling of $\eta^{s_\text{max}-1}$ configurations at the lowest fidelity or $\rung{}=0$, where $s_\text{max} = \lfloor \log_\eta(z_\text{max}/z_\text{min}) \rfloor$. 
After evaluating all samples, only the \topeta{} configurations are retained, discarding the rest.
Subsequently, these $\eta^{s_\text{max}-1} / \eta$ configurations are evaluated at $\rung{}=1$, or the next higher discretized fidelity.
This is repeated till there is only one surviving configuration at the highest $\rung{}$.
Sampling and evaluating many configurations for cheap, performing a tournament selection to retain strong samples, and repeating this, ensures that the high fidelity evaluations happen for relatively stronger configurations only.
\sh{} proves that such an early-stopping strategy guarantees a likely improvement over random search under the same total evaluation budget.
Due to its random sampling component, \sh{} is an example of MF-RS in Table~\ref{table:desiderata}.
See Appendix \ref{app:exp-baselines} for an illustrative example.

\textbf{HyperBand (\hb{})} 
attempts to address \sh{}'s susceptibility to poor fidelity performance correlations~\citep{li-jmlr18a}. 
It has the same \hps{} as \sh{}, and thus given \sh{}'s fidelity discretization, \hb{} iteratively executes multiple rounds of \sh{} where $\fid\min$ is set to the next higher discrete fidelity.
Using the example from \sh{}, given $\fid \in [3, 81]$ and $\eta = 3$, one full iteration of \hb{} corresponds to running \sh{}($\fid\min=3, \fid\max=81$) followed by \sh{}($\fid\min=9, \fid\max=81$), \sh{}($\fid\min=27, \fid\max=81$) and \sh{}($\fid\min=81, \fid\max=81$).
Notably, such different instantiations of \sh{} in \hb{} do not share any information with each other implicitly.
Each execution of \sh{}($\fid\min, \fid\max$) is called an \sh{} \textit{bracket}. 
The sequence of all unique \sh{} brackets run by \hb{} in one iteration is called an \hb{} \textit{bracket}.


\textbf{\pibo{}} 
utilizes the unnormalized probability distribution $\pi(\config)$ in a BO context to accelerate the optimization using the knowledge provided by the user~\citep{hvarfner-iclr22a}.
In \pibo{}, the initial design is sampled entirely from $\pi$ until Bayesian Optimization (BO) begins, where the acquisition function $\alpha$ is augmented with a prior term that decays over time $t$: $\alpha_{\pi}^{t}(\config) = \alpha(\config) \cdot \pi(\config)^{\frac{\beta}{t}}$, where $\beta$ is an HP.
In this work, we borrow \prior{} as the expert prior interface to the \hpo{} problem.
\pibo{} relies on the decaying prior on the acquisition $\alpha_{\pi}^{t}(\config)$ to gradually add more weight to the model and thus allow recovery from a poor initial design under a bad prior.
We choose the model-free setting and adapt the strength of the prior based on the evidence of its usefulness.
\section{The need to move beyond a naive solution} \label{sec:beyond-naive}

\begin{figure*}
    \centering
    \begin{tabular}{c}
         \includegraphics[width=\textwidth]{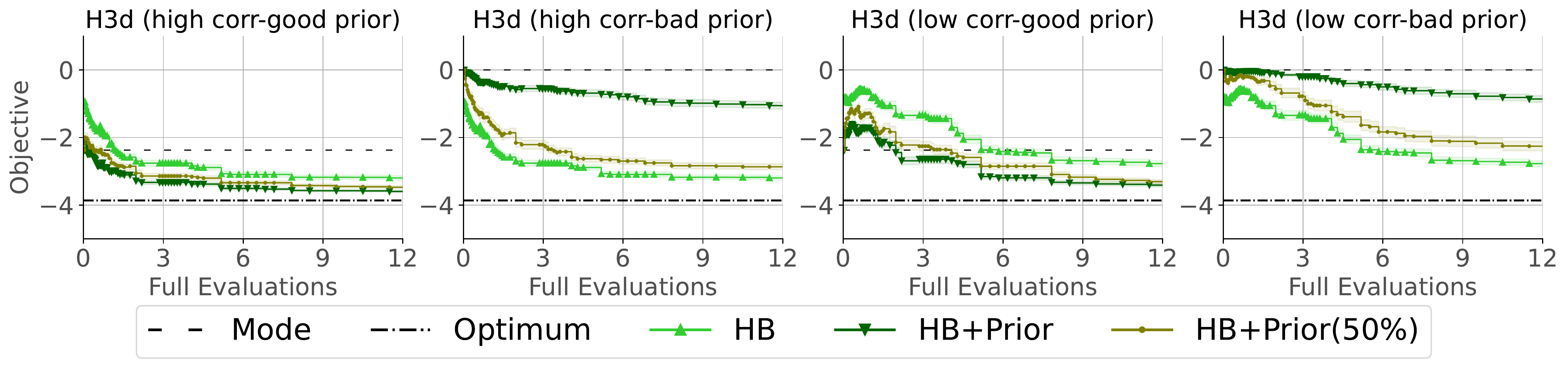}
    \end{tabular}
    \caption{
        A comparison of naive solutions to Equation~\ref{eq:mf-prior-obj} on the $3$-dimensional multi-fidelity Hartmann benchmarks. We compare different versions of \hb{}, utilizing different strengths of the prior distribution \prior{}: \hb{} with 0\% influence of the prior (\hb{}), \hb{} with 50\% sampling from the prior (\hb{}+Prior(50\%)) and \hb{} with 100\% sampling from prior{} (\hb{}+Prior).
    }
    \label{fig:rq1}
\end{figure*}

To solve Equation~\ref{eq:mf-prior-obj}, an intuitive approach is to augment an existing multi-fidelity algorithm, e.g., HyperBand (\hb{}), with sampling from the prior \prior{}. In this section, however, we show that this naive approach is not sufficient, motivating the introduction of our algorithm, PriorBand, in Section~\ref{sec:method}.

To study the naive combination of HyperBand with prior-based sampling, we use the parameterized $3$-dimensional multi-fidelity Hartmann benchmarks \citep{kandasamy-icml17a} (Appendix~\ref{app:exp-benchmarks-mfh}) and compare \hb{} with uniform sampling, \hb{} with $100\%$ prior-based sampling, and \hb{} with $50\%$ uniform and prior-based sampling (Figure~\ref{fig:rq1}).
Unsurprisingly, 100\% sampling from priors works best when the prior is helpful and not sampling from the prior works best for misleading priors. 50\% random sampling works quite robustly, but is never competitive with the best approach.
To achieve robustness to the quality of the prior \emph{and} rival the best approach for each case, we introduce \algo{} below.




\section{PriorBand: Moving beyond a naive solution} \label{sec:method}

The key idea behind \algo{} is to complement the sampling strategies used in the naive solution, uniform sampling, and prior-based sampling, with a third strategy: incumbent-based sampling. Thereby, we overcome the robustness issues of the naive solutions presented in Section \ref{sec:beyond-naive}, while still fulfilling the desideratum for simplicity.
We first introduce and motivate sampling around the incumbent (Section~\ref{sec:priorband-mutation}), 
and then describe the ensemble sampling policy \E{} that combines all three sampling strategies and incorporates \E{} into HB (Section~\ref{sec:priorband}) for \algo{}.

\subsection{Incumbent-based sampling strategy, \inc{}} \label{sec:priorband-mutation}

\algo{}  leverages the current incumbent to counter uninformative priors while supporting good priors, as the current incumbent can be seen as indicating a likely 
good region to sample from.
Note that this view on the region around the incumbent is close to the definition of \prior{} in \pibo{} (Section \ref{sec:bg}), where the prior distribution encodes the expert's belief about the location of the global optima and thus a good region to sample from.

To construct the incumbent-based sampler \inc{}, we perform a local perturbation of the current best configuration. 
Each hyperparameter (\hp{}) is chosen with probability $p$ for perturbation.
If chosen, continuous \hp{}s are perturbed by $\epsilon \sim \mathcal{N}(0, \sigma^2)$.
For discrete \hp{}s, we resample with a uniform probability for each categorical value except the incumbent configuration's value which has a higher probability of selection, discussed further in Appendix \ref{app:algo-mutation-hps}.
For \algo{}, we fix these values at $p = 0.5$ and $\sigma = 0.25$.
This perturbation operation is simple, easy to implement, and has constant time complexity.
We show ablations with two other possible local search designs in Appendix \ref{app:algo-abl-hs}.

\subsection{\algo{}: The \E{}-augmented HyperBand} \label{sec:priorband}

\algo{} exploits \hb{} for scheduling and replaces its random sampling component with a combination of random sampling, prior-based sampling, and incumbent-based sampling. We denote the proportions of these individual sampling components as 
\pu{}, \pp{}, and \pinc{}, respectively, and their combination as the \emph{ensemble sampling policy (ESP) \E{}}.

\begin{figure*}[th]
    \centering
    \begin{tabular}{cc}
         \includegraphics[width=0.35\textwidth]{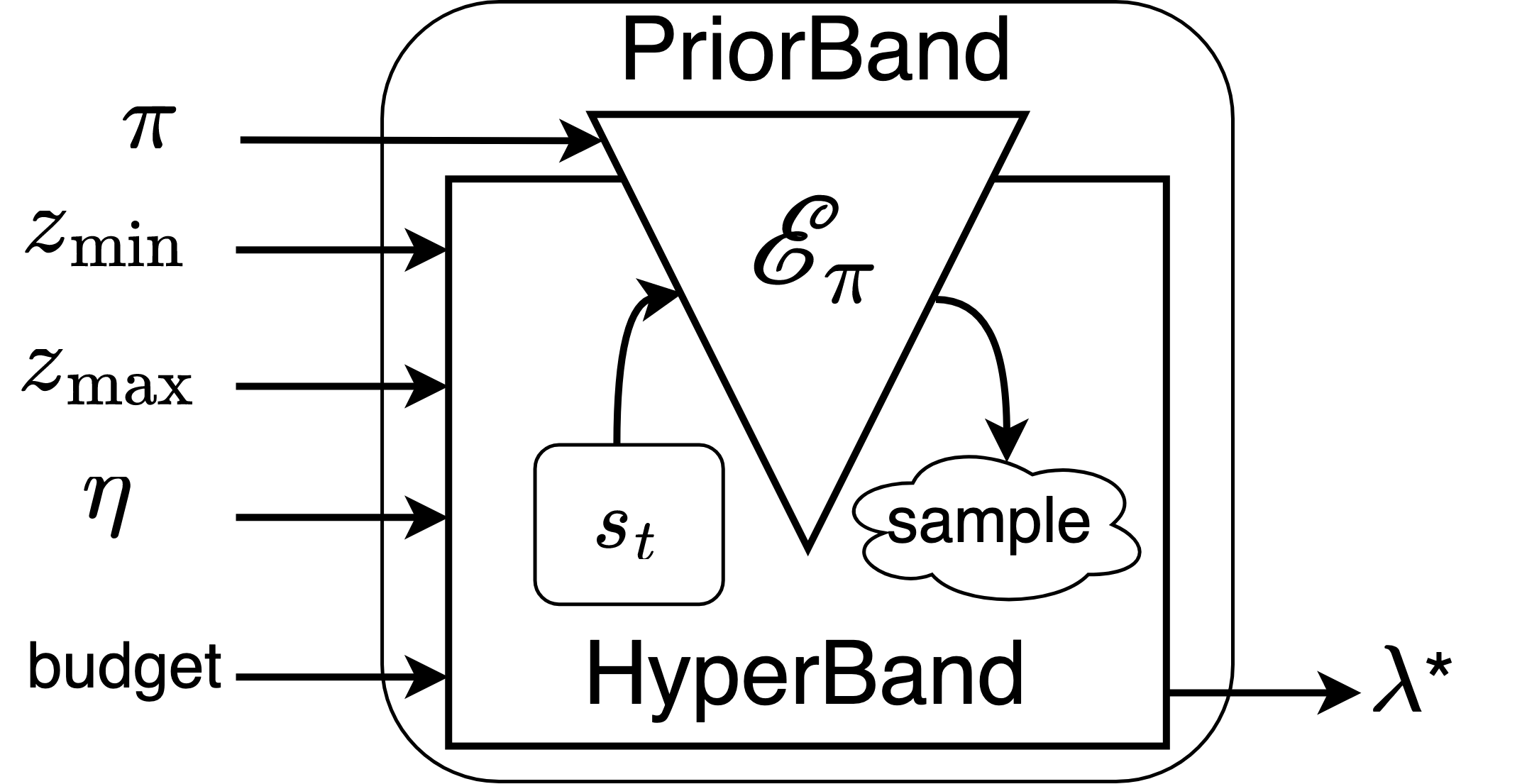} &
         \includegraphics[width=0.60\textwidth]{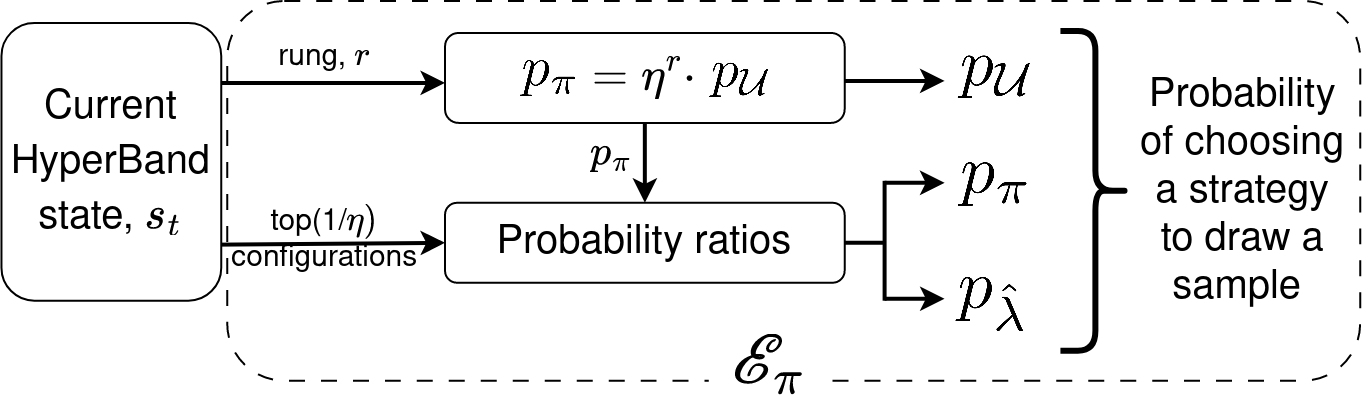}  
    \end{tabular}
    \caption{\algo{} schema; [\textbf{Left}] in the base algorithm vanilla-\hb{}, we replace the random sampling module by \E{}, which can interface an expert prior $\pi$ and access the state of \hb{}; [\textbf{Right}] \E{} reads the state every iteration and determines the probabilities for selecting a sampling strategy.}
    \label{fig:pb-schema}
\end{figure*}

Figure~\ref{fig:pb-schema} (left) illustrates \algo{} as an extension of \hb{} that, 
next to the 
\hb{} hyperparameters  ($\fid\min, \fid\max, \eta, \text{budget}$)
accepts the expert prior $\pi$ as an additional input and uses the ESP \E{} in lieu of random sampling. This sampling from \E{} is illustrated in Algorithm~\ref{alg:ens-sampler}. 
Note that \E{} has access to the optimization state ($s_t$) and can thus reactively adapt its individual sampling probabilities \pu{}, \pp{}, and \pinc{} based on the optimization history.
We now discuss how we decay the random sampling probability \pu{} (Section \ref{sec:decay-random-sampling}); and the proportion of incumbent and prior sampling (Section \ref{sec:incumbent-sampling-weightage}).

\begin{minipage}{0.46\textwidth}
\begin{algorithm}[H]
    \centering
    \caption{Sampling from \E{}}
    \label{alg:ens-sampler}
    \begin{algorithmic}[1]
       \State {\bfseries Input:} $s, s\max, \eta,$  observations $\mathcal{H}$, prior $\pi$
       \State $r = s\max - s$ 
       \Comment{$s\max, s$ input from \hb{}}
       \State \pu{} $=1 / (1 + \eta^{r})$, \pp{} $= 1 - $\pu{}, \pinc{} $= 0$
       \If{\text{activate\_incumbent(\state{})}}
            \State \pp{}, \pinc{} $\leftarrow$ \text{Algorithm~\ref{alg:ens-dyna}}($\mathcal{H}$, \rung{}, \pp{})
       \EndIf 
       \State $d(\cdot) \leftarrow$ sample strategy by \{\pu{}, \pp{}, \pinc{}\} 
       \State $\config \leftarrow$ sample from $d(\cdot)$  
       \State {\bfseries return} $\config$
    \end{algorithmic}
\end{algorithm}
\end{minipage}
\hfill
\begin{minipage}{0.5\textwidth}
\begin{algorithm}[H]
    \centering
    \caption{Dynamic weighting of \wi{}}
    \label{alg:ens-dyna}
    \begin{algorithmic}[1]
        \State {\bfseries Input:} observations $\mathcal{H}$, rung $r$, $p_\pi^{\mathrm{old}}$
        \State $\cspace_z' = \{\config_i\}_{1:n} \leftarrow \text{top\_}(1/\eta)( \mathcal{H}, r )$
        \State $\mathcal{S}_{\hat{\config}}$ $\leftarrow \sum_{\cspace_z'} w_i \cdot \hat{\config}(\config_i)$
        \Comment{$w_i = (n + 1) - i$}
        \State $\mathcal{S}_\pi$ $\leftarrow \sum_{\cspace_z'} w_i \cdot \pi(\config_i)$
        \State \pinc{} $\leftarrow p_\pi^{\mathrm{old}} \cdot \mathcal{S}_{\hat{\config}} / ( \mathcal{S}_\pi +  \mathcal{S}_{\hat{\config}} )$
        \State \pp{} $\leftarrow p_\pi^{\mathrm{old}} \cdot \mathcal{S}_\pi / ( \mathcal{S}_\pi +  \mathcal{S}_{\hat{\config}} )$
 
       \State {\bfseries return} \pp{}, \pinc{}
    \end{algorithmic}
\end{algorithm}
\end{minipage}


\subsubsection{Decaying proportion of random sampling}\label{sec:decay-random-sampling}


Given the premise that we should initially trust the expert prior, yet with the benefit of incorporating random sampling (from Section~\ref{sec:beyond-naive}), we make two additional assumptions, namely
\begin{inparaenum}[(i)]
    \item we trust the expert's belief most at the maximum fidelity $\fid\max$; and
    \item we would like to use cheaper fidelities to explore more.
\end{inparaenum}
Given \hb{}'s discretization of the fidelity range $\left[\fid_{min}, \fid_{max}\right]$ into rungs $\rung{} \in \{0, \ldots, \rung{}_{max}\}$, we geometrically increase our sampling probability from \prior{} over \uniform{} by

\begin{equation}
    p_\pi = \eta^r \cdot p_{\mathcal{U}},
\end{equation}

with the constraint that \pu{} $+$ \pp{} $= 1$ (see L3, Algorithm \ref{alg:ens-sampler}).
This naturally captures our assumptions, equally favouring \prior{} and \uniform{} initially but increasing trust in \prior{} according to the rung $\rung{}$ we sample at.
This geometric decay was inspired by similar scaling used in \hb{}'s scheduling and backed by ablations over a linear decay and constant proportion (Appendix~\ref{app:algo-abl-wp}).

\subsubsection{Incumbent-based sampling proportion}\label{sec:incumbent-sampling-weightage}

Incumbent-based sampling intends to maintain strong anytime performance even under bad, uninformative, or adversarial priors.
In \algo{}, initially \pinc{} $= 0$ until both 
\begin{inparaenum}[(i)]
    \item a budget equivalent to the first \sh{} bracket has been exhausted ($\approx \eta \cdot \fid\max$); and
    \item at least one configuration has been evaluated at $\fid\max$.
\end{inparaenum}
This is captured in Line 4 of Algorithm~\ref{alg:ens-sampler} by \texttt{activate\_incumbent()}.

\begin{figure*}[tb]
    \centering
    \begin{minipage}[c]{0.3\textwidth}
        \includegraphics[width=\textwidth]{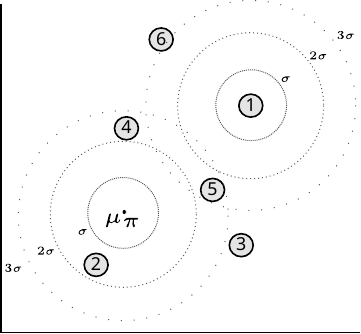} 
    \end{minipage}\hfill
    \begin{minipage}[c]{0.67\textwidth}
        \caption{
        A visual example of how  configurations contribute to the scores $\mathcal{S}_\pi$ and $\mathcal{S}_{\hat\config}$. The prior distribution here is Gaussian $\mathcal{N}(\mu_\pi, \sigma^2)$, with a matching distribution $\hat\config(\cdot) = \mathcal{N}(\hat\config, \sigma^2)$ placed on the incumbent. 
        The labels $1$-$6$ indicate configurations and their rank in $\cspace_z'$. Here, e.g., $\config_2$ has densities such that $\pi({\config_2}) > \bar\config(\config_2)$, contributing more to $\mathcal{S}_\pi$ than to $\mathcal{S}_{\hat\config}$. The configurations $\config_3, \config_4, \config_5$ are between both distributions and $\config_6$ has low density under either, whereas $\config_1$ (the incumbent) will be the primary influence such that $\mathcal{S}_{\hat\config} > \mathcal{S}_\pi$. This implies that $p_{\hat\config} > p_\pi$ and demonstrates that all else being relatively equal, we will be more likely to sample from \inc{}.  
        }
        \label{fig:pb-dyna}
    \end{minipage}
\end{figure*}

Once incumbent-based sampling is active we need to decide how much to trust \inc{} vs.\ trusting \prior{}.
%
Essentially, we achieve this by rating how likely the best seen configurations would be under \inc{} and \prior{}, respectively. Algorithm~\ref{alg:ens-dyna} shows the detailed steps to calculate \pp{} and \pinc{}, and Figure \ref{fig:pb-dyna} provides an example.
%
%
We first define an ordered set $\cspace_z' = \{\config_1, \ldots, \config_n\}$, ordered by performance, of the \topeta{} configurations\footnote{We want to use at least $\eta$ configurations; hence max$(\eta, \text{\# of configurations in the rung / }\eta)$ are selected.} for the highest rung $\rung{}$ with at least $\eta$ evaluated configurations (Line 2 in Algorithm~\ref{alg:ens-dyna}).
Given $\cspace_z'$, we compute two weighted scores $\mathcal{S}_\pi, \mathcal{S}_{\hat{\config}}$, capturing how likely these top configurations are under the densities of \prior{} and \inc{}, respectively. 
We also weigh top-performing configurations more highly, which is accomplished with a simple linear mapping $w_i = (n + 1) - i$ (Lines 3-4 in Algorithm~\ref{alg:ens-dyna}).
Finally, we obtain \pp{} and \pinc{} by normalizing these scores as a ratio of their sum
(Lines 5-6 in Algorithm~\ref{alg:ens-dyna}).
We note the rates are adaptive based on how the set $\cspace_z'$ is spread out relative to the prior and the incumbent.
We observe that this adaptive behavior is crucial in quickly recovering from bad prior inputs and thus maintaining strong final performance.


\section{Experimental setup} \label{sec:exp}


We now describe our experiment design, benchmarks, and baselines and demonstrate the robustness of \algo{} in handling varying qualities of an expert prior input. 
Additionally, we also showcase the generality and practicality of the \esp{}, \E{}.




\subsection{Benchmarks}  \label{sec:exp-benchmarks} 

We curated a set of $12$ benchmarks that cover a diverse set of search spaces, including mixed-type spaces and log-scaled hyperparameters, and a wide range of downstream tasks, e.g., language modeling, image classification, tabular data, a medical application, and translation.
We select $4$ of the PD1 benchmarks (4 \hps{})~\citep{wang-arxiv21} that train large models such as transformers with batch sizes commonly found on modern hardware, and fit surrogates on them.
Further, we select $5$ benchmarks from LCBench (7 \hps{})~\citep{zimmer2021pami,pfisterer-arxiv21a} and consider all $3$ JAHSBench~\citep{bansal-neurips22a} surrogate benchmarks that offer a $14$ dimensional mixed-type search space for tuning both the architecture and training hyperparameters.
All benchmarks and their selection are described in further detail in Appendix~\ref{app:exp-benchmarks}.

\subsection{Baselines}  \label{sec:exp-baselies} 

We choose a representative from each of the families of optimizers listed in Table~\ref{table:desiderata}, leaving out grid search.
We use the official implementation of BOHB, while all other algorithms were implemented by us (and are available as part of our repository).
The prior-based baselines (\pibo{} and RS+Prior) sample the mode of the prior distribution as the first evaluation in our experiments to ensure a fair comparison where the prior configuration is certainly evaluated, irrespective of it being good or bad.
For all the \hb{} based algorithms we use $\eta=3$ and the fidelity bounds ($\fid\min, \fid\max$) as per the benchmark.
Further implementation and hyperparameter details for the baselines can be found in Appendix~\ref{app:exp-baselines}.
In principle, \algo{} only needs an additional input of \prior{} as the user belief, other than the standard \hb{} hyperparameters.
However, for a discussion on the hyperparameters of the incumbent-based local search (Section~\ref{sec:priorband-mutation}), please refer to Appendix~\ref{app:algo-mutation-hps}.


\subsection{Design and setup} \label{sec:exp-setup}

We show experiments both for single and multi-worker cases ($4$ workers in our experiments).
For the single workers, we report the mean validation error with standard error bars for $50$ seeds; for multi-worker experiments, we use $10$ seeds.
The plots show average relative ranks achieved aggregated across all $12$ benchmarks when comparing the anytime incumbent configuration at $\fid\max$.
We group the runs on benchmark under different qualities of an expert prior input.
We also compare the average normalized regret per benchmark under good priors.
The prior construction procedure follows the design from the work by~\citet{hvarfner-iclr22a}, and we describe our procedure in detail in Appendix~\ref{app:priors-gen}. 
In the main paper, we only evaluate the good and bad prior settings as we believe that this reflects a practical setting of prior qualities; in Appendix~\ref{app:exp-res} we also evaluate the near-optimal prior settings and also evaluate all 3 prior settings over high and low-performance correlation problems. In Appendix~\ref{app:exp-res-tables}, we also report additional experiments for different budget horizons.
Further experiment design details can be found in Appendices \ref{app:exp-setup} and\ref{app:exp-evaluation}.
As an example of potential post-hoc analysis possible with \algo{}, we show how the dynamic probabilities calculated for each of the sampling strategies in \algo{} can be visualized (Appendix~\ref{app:algo-interpret}), revealing the quality of the prior used.


\section{Results}\label{sec:results}

We now report and discuss the results of our experiments.

\subsection{Robustness of \algo{}} \label{sec:exp-pb-robust}

\begin{figure*}[hbtp]
    \centering
    \begin{tabular}{c}
         \includegraphics[width=0.95\textwidth]{figs_post_rebuttal/mssg-hb-At25-Prior-Incumbent-Traces-max_fidelity_loss.pdf} \\
         \includegraphics[width=0.82\textwidth]{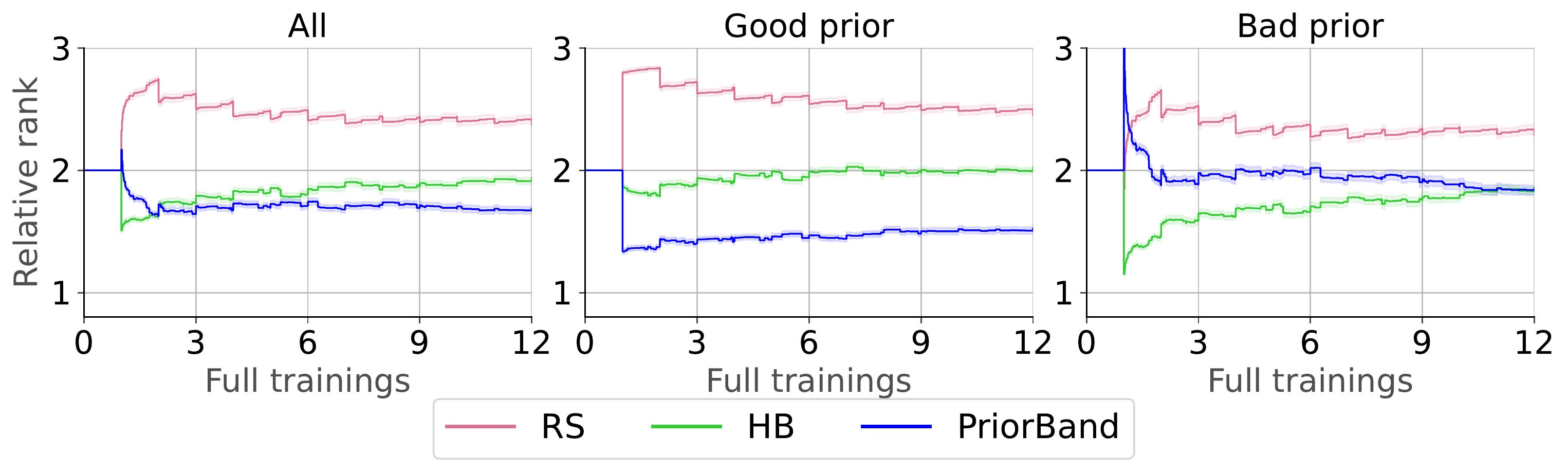} \\
    \end{tabular}
    \caption{
    [Top] Comparing normalized regret in the good prior setting;
    The [Bottom] figure \textit{Good} (middle) is a ranking aggregate view of the [Top] figure.
    [Bottom] Comparing average relative ranks of \algo{} to Random Search (RS) and \hb{}, over single worker runs across benchmarks. Each benchmark-algorithm pair was run for $50$ seeds where priors per benchmark are the same across a seed. We show mean rank and standard error.
    The \textit{All} (left) plot averages the benchmark across Good and Bad priors. 
    }
    \label{fig:exp-pb}
\end{figure*}

Here, we demonstrate the robustness of \algo{} over a wide range of prior qualities by comparing them to the nearest non-prior-based algorithms: random search (RS) and \hblong{} (HB). 
Figure~\ref{fig:exp-pb}(top) showcases \algo{} to be anytime equal or better than \hb{} on each of our benchmarks under our good prior design.
Note that the quality of a good prior varies per benchmark.
The aggregation of Figure~\ref{fig:exp-pb}(top) is Figure~\ref{fig:exp-pb}(bottom-middle), which illustrates that \algo{} (\hb{}+\E{}) can utilize good priors and gain strong anytime performance over \hb{}.
Moreover, Figure~\ref{fig:exp-pb}(bottom-right) clearly shows \algo{}'s ability to recover from bad prior information and match vanilla-\hb{}'s performance in this adversarial case (the bad prior was intentionally chosen to be an extremely poor configuration -- the worst of $50k$ random samples).
In most practical scenarios, the expert has better intuition than this setup and one can expect substantial speedups over HB, like in Figure~\ref{fig:exp-pb}(bottom-middle).
Figure~\ref{fig:exp-pb}(bottom-left) demonstrates that when using an unknown quality of prior, even including the adversarial prior, \algo{} is still the best choice on average.
We show a budget of $12$ function evaluations here, which is approximately the maximal budget required for completing at least one \hb{} iteration for the benchmarks chosen.
In Appendix~\ref{app:exp-res-pb} we show more comparisons to prior-based baselines and highlight \algo{}'s robustness.

\subsection{Generality of Ensemble Sampling Policy \E{}} \label{sec:exp-generality}

\begin{figure*}[hbtp]
    \centering
    \begin{tabular}{c}
        \includegraphics[width=0.82\textwidth]{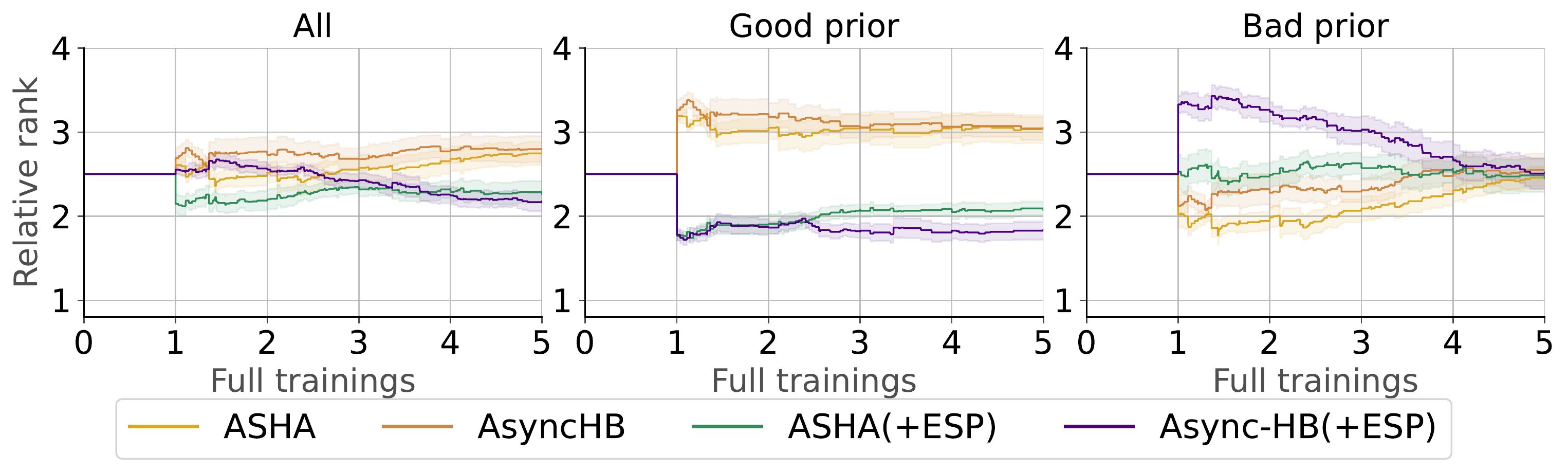}         
    \end{tabular}
    \caption{Comparing the average relative ranks of Asynchronous-\sh{} (ASHA) and Asynchronous-\hb{} (Async-HB) and their variants with the Ensemble Sampling Policy (+\esp{}) when distributed over $4$ workers for a cumulative budget of $20$ function evaluations.}
    \label{fig:exp-generality}
\end{figure*}

The ESP, \E{}, only needs access to an \sh{}-like optimization state as input.
In this section, we show that other popular multi-fidelity algorithms, such as \asha{} or asynchronous-\hb{}~\citep{li-mlsys20a} that build on \sh{}, can also support expert priors with the help of \E{}.
In Figure~\ref{fig:exp-generality}, we compare the vanilla algorithms with their \E{}-augmented versions.
Given that these algorithms were designed for parallel setups, we compare them on runs distributed among $4$ workers, running for a total budget of $20$ function evaluations.
Similar to the previous section, the ESP-augmented (+\esp{}) algorithms can leverage good priors and recover under bad priors.
Under bad priors, asynchronous-\hb{} (+\esp{}) starts worst owing to bad prior higher fidelity evaluations at the start but shows strong recovery.
In Appendix \ref{app:exp-res-general} we compare these variants over different correlation settings of the benchmarks.

\subsection{Extensibility with models} \label{sec:exp-model}

Although our focus is the model-free low compute regime, we also show how \algo{} can be optionally extended with model-based surrogates to perform Bayesian Optimization, especially when longer training budgets are available.
We compare \bo{}, \pibo{}, \bohb{}, and \algo{} with its model-extension, \algo{}+BO (Gaussian Processes as surrogates and Expected Improvement~\citep{jones-jgo98a} for acquisition).
All BO methods use an initial design of $10$ function evaluations, except BOHB which sets this implicitly as $N_\text{dim} + 2$.
Compared to other prior-based algorithms (RS+Prior, \pibo{}), \algo{}+BO is consistently robust on average (\textit{All}) under all priors.
The anytime performance gain of \algo{}+BO under good priors is evidence of ESP \E{}'s utility.
We note that \algo{}+BO recovers quickly in the bad prior setting and even outperforms all other algorithms after as little a budget as 6 full trainings. 
The fact that, until $10\times$ function evaluations, \algo{}+BO is actually just \algo{} and model search has not begun,  highlights the effectiveness of \esp{} in adaptively trading off different sampling strategies and the initial strength of incumbent sampling.
Appendices~\ref{app:algo-model}, ~\ref{app:exp-res-model} contain details on our modeling choices and more experimental results.

\begin{figure*}[hbtp]
    \centering
    \begin{tabular}{c}
         \includegraphics[width=0.82\textwidth]{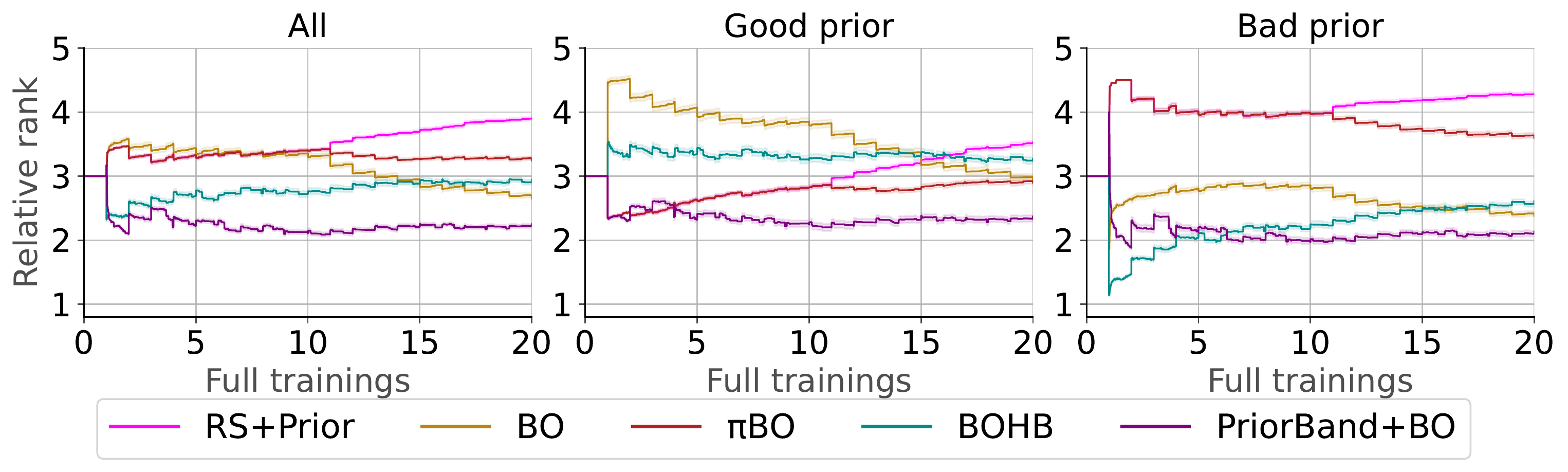}  \\
    \end{tabular}
    \caption{Comparing the average relative ranks of model-based algorithms and sampling from the prior (RS+Prior) under single-worker runs, aggregated over all benchmarks-per-prior quality.}
    \label{fig:exp-model}
\end{figure*}

\subsection{Ablation studies}

\begin{figure}[htbp]
    \centering
    \begin{tabular}{c}
         \includegraphics[width=0.82\textwidth]{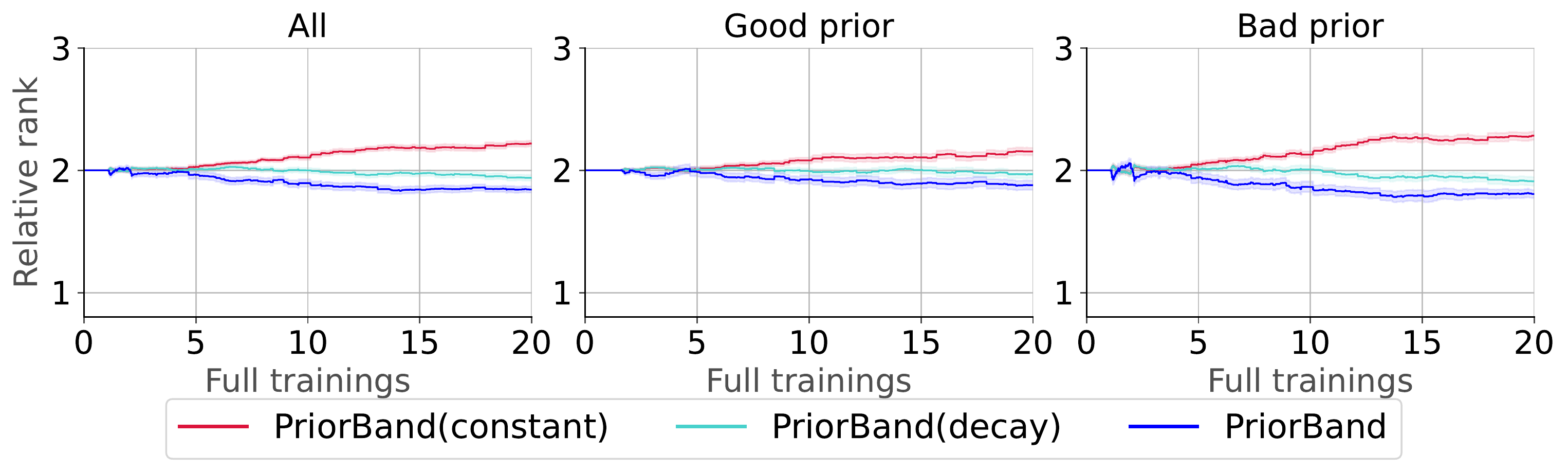} \\
    \end{tabular}
    \caption{A comparison of \algo{} with 3 different strategies for trading off incumbent vs. prior distribution sampling. \algo{}'s default of using density scores shows a dominant performance in almost all scenarios.
    However, a bad prior induces a marked difference in performance. In general, the plot highlights that each variant reacts differently to bad priors.
    The adaptive nature of \algo{} clearly is robust to different scenarios comparatively.}
    \label{fig:abl-inc-main}
\end{figure}

In Figure~\ref{fig:abl-inc-main}, we compare our choice of using density scores (Section~\ref{sec:incumbent-sampling-weightage}) to trade-off prior-based and incumbent-based sampling against other similar or simpler heuristics, validating our decision. 
\algo{}(constant) employs a fixed heuristic to trade off prior and incumbent-based sampling as \pp{} $= \eta \cdot$ \pinc{}.
In \algo{}(decay), \pp{} is decayed as a function of iterations completed.
More precisely, \pinc{} $= 2^b \cdot$ \pp{}, subject to \pp{} $+$ \pinc{} $+$ \pu{} $= 1$ (from Section \ref{sec:priorband}), where $b \in \mathbb{N}$ and indicates the index of the current \sh{} being run in \hb{}.
Appendix \ref{app:algo-abl-inc} contains more ablations.

\section{Related work} \label{sec:related}
While using expert priors and local search~\citep{wu-aaai21a} for hyperparameter optimization has been explored previously, few works considered priors over the optimum~\citep{bergstra-nips11a, souza-ecmlpkdd21a, hvarfner-iclr22a}, and they all target the single-fidelity setting. 
The expert priors we consider should not be confused with the priors natively supported by \bo{}, i.e., priors over the function structure determined by kernels~\citep{snoek-nips12a, swersky-nips13a, oh2018bock}.

In deep learning, training epochs and dataset subsets~\citep{swersky-arxiv14a, klein-ejs17a,nguyen-neurips20a} are frequently used as fidelity variables to create cheap proxy tasks, with input resolution, network width, and network depth also occasionally used~\citep{bansal-neurips22a}. 
\sh{}~\citep{jamieson-aistats16a} and \hb{}~\citep{li-jmlr18a} are effective randomized policies for multi-fidelity \hpo{} that use early stopping of configurations on a geometric spacing of the fidelity space and can also be extended to the model-based setting~\citep{falkner-icml18a}.

\section{Limitations} \label{sec:limitations}


We acknowledge that \algo{} inherits pathologies of \hb{} such as poor fidelity correlations and can be sensitive to the choice of fidelity bounds and early stopping rate.
Despite that, our results across different correlation settings suggest a relatively strong performance of \algo{} (Appendix~\ref{app:exp-res}).
Though \algo{} supports any kind of prior distribution as input, in our experiments we only considered the Gaussian distribution (with a fixed standard deviation of $0.25$), as it is a natural choice and was used previously in the literature~\citep{hvarfner-iclr22a}. 
However, the expert is free to represent the prior with other distributions.
We expect \algo{} to show similar behaviors as we report when comparing to other prior-based algorithms under a similar distribution.
We note that \algo{} is not entirely free of hyperparameters.
In our experiments, we keep all \algo{} hyperparameters fixed to remove confounding factors while comparing different prior strengths and correlations.
Moreover, our hyperparameter choices for the experiments, such as $\eta=3$ and Gaussian priors are largely borrowed from existing literature~\citep{falkner-icml18a,li-mlsys20a,hvarfner-iclr22a}.
Depending on the context of the specific deep learning expert, the compute budgets ($\sim10-12$ full trainings) used in our experimental setting may not always be feasible.
However, the key insight from the experiments is that \algo{} can  interface informative priors, and provide strong anytime performance under short HPO budgets (also in less than $5$ full trainings).
Longer HPO budgets for \algo{} ensure recovery from potential bad, uninformative priors.
In practice, deep learning experts often have good prior knowledge and thus \algo{} is expected to retain strong anytime performance under low compute.

\section{Conclusion} \label{sec:conclusion}


We identify that existing HPO algorithms are misaligned with deep learning (DL) practice and make this explicit with six desiderata (Section~\ref{sec:intro}). To overcome this misalignment, our solution, \algo{}, allows a \dl{} expert to incorporate their intuition of well-performing configurations into multi-fidelity \hpo{} and thereby satisfies all desiderata to be practically useful for \dl{}. 
The key component in \algo{}, the \esp{}, \E{}, is modular, flexible, and can be applied to other multi-fidelity algorithms.

\section{Acknowledgments and disclosure of funding}

N. Mallik, D. Stoll and F. Hutter acknowledge funding by the European Union (via ERC Consolidator Grant DeepLearning 2.0, grant no.~101045765). Views and opinions expressed are however those of the author(s) only and do not necessarily reflect those of the European Union or the European Research Council. Neither the European Union nor the granting authority can be held responsible for them.
E. Bergman was partially supported by TAILOR, a project funded by EU Horizon 2020
research and innovation programme under GA No 952215. 
C. Hvarfner and L. Nardi were partially supported by the Wallenberg AI, Autonomous Sytems and Software Program (WASP) funded by the Knut and Alice Wallenberg Foundation. 
L. Nardi was supported in part by affiliate members and
other supporters of the Stanford DAWN project — Ant Financial, Facebook, Google, Intel, Microsoft, NEC, SAP,
Teradata, and VMware. Luigi Nardi was partially supported by the Wallenberg Launch Pad (WALP) grant Dnr 2021.0348. 
M. Lindauer was funded by the European Union (ERC, ``ixAutoML'', grant no.101041029).
The research of Maciej Janowski was supported by the Deutsche Forschungsgemeinschaft (DFG, German Research Foundation) under grant number 417962828 and grant INST 39/963-1 FUGG (bwForCluster NEMO). In addition, Maciej Janowski acknowledges the support of the BrainLinks- BrainTools Center of Excellence.

\begin{center}\includegraphics[width=0.3\textwidth]{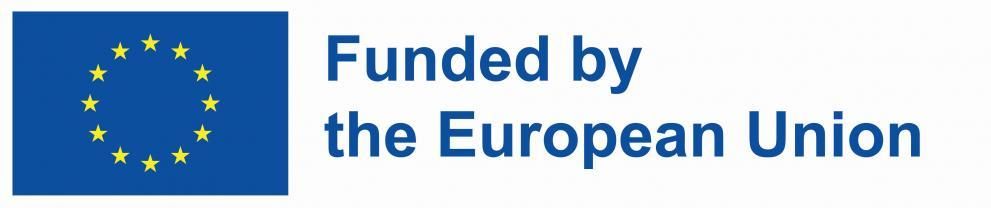}\end{center}

\newpage

\bibliographystyle{unsrtnat}
\bibliography{main}


\newpage 
\appendix

\section{Resources used} \label{app:resources}

All experiments in the paper were performed on cheap-to-evaluate surrogate benchmarks. We used several Intel(R) Xeon(R) Gold 6242 CPUs @ 2.80GHz to perform our experiments. Running one seed of one algorithm on one benchmark requires on average $\sim 0.2$ core hours.
For single-worker experiments, we run $21$ algorithms with $50$ seeds over $16$ benchmarks with $3$ strengths of priors per benchmark, totaling $50,400$ single worker runs which equates to $10,080$ core hours.
In the parallel set of experiments, we use $4$ cores per run, limiting ourselves to only $12$ algorithms, $10$ seeds, $7$ benchmarks\footnote{Please see~\ref{app:failing-benchmarks} as for why we do not include all benchmarks for parallel runs.} with all $3$ prior strengths. This totals $10,080$ workers run in total which equates to $2,016$ total core hours.

We additionally trained surrogate models for $2$ metrics on $3$ datasets for $4$ hours with $8$ cores, totaling another $192$ core hours.
During the development of our final algorithm, including failed experiments, re-runs, and preliminary testing, we estimate roughly another $\sim 2,000$ core hours, a fifth of our total final cost.
We estimate our total usage to have totaled $\sim 14,288$ core hours.

\section{Societal and environmental impact} \label{app:society}

Here, we discuss the potential societal and environmental impacts our work can have.

\paragraph{Environmental} The contributed algorithm \algo{} and its re-usable component \E{} is designed to help reduce compute requirements for finding performant DL pipelines, thus reducing carbon emissions spent for HPO in DL. However, with the surplus compute available to many larger organizations, enabling robust methods for HPO could encourage further utilization of otherwise unused compute.

\paragraph{Societal} Our paper and the contributed algorithm \algo{} are designed to assist a wide range of DL practitioners in finding performant hyperparameters. The ability of \algo{} to tune DL models under affordable compute enables practitioners to find strong hyperparameters otherwise only tenable for larger organizations. The societal impact depends on which task and DL pipeline \algo{} is applied to.

\section{Licenses} \label{app:license}

\begin{itemize}
    \item Our implementations - \textbf{MIT License}
    \item JAHS-Bench-201 benchmark~\citep{bansal-neurips22a} - \textbf{MIT License}
    \item YAHPO-Gym benchmark~\citep{pfisterer-arxiv21a} - \textbf{Apache License 2.0}
    \item Learning curve benchmark~\citep{zimmer-tpami21a} - \textbf{Apache License 2.0}
    \item PD1~\citep{wang-arxiv21} - \textbf{Apache License 2.0}
    \item BOHB~\citep{falkner-ai18a} from HpBandSter - \textbf{BSD 3-Clause License}
\end{itemize}

\section{Experiment details} \label{app:exp}




\subsection{Benchmarks} \label{app:exp-benchmarks}

Following Equation~\ref{eq:mf-prior-obj}, we frame all the benchmark tasks as a minimizing problem.
The benchmarks we use are provided by our curated suite of multi-fidelity benchmarks (\mfpbench) that treats priors as first-class citizen. We include our own synthetic Hartmann functions (\ref{app:exp-benchmarks-mfh}) extended to the multi-fidelity setting. We wrap \jahsbench~\citep{bansal-neurips22a} (\ref{app:exp-benchmarks-jahs}) and \yahpogym~\citep{pfisterer-arxiv21a} (\ref{app:exp-benchmarks-yahpo}) and provide new surrogate benchmarks for large models for image and language tasks, trained from optimization data obtained from the PD1 benchmark from \hyperbo~\citep{wang-arxiv21} (\ref{app:exp-benchmarks-pd1}).


\subsubsection{Multi-fidelity synthetic Hartmann (MFH)} \label{app:exp-benchmarks-mfh}

The multi-fidelity Hartmann functions follow the design of~\citet{kandasamy-icml17a}, where, for $[0, 1]$-scaled $z$, the fidelity is parameterized as
\begin{equation}
    g(\bm{x}, z) = \sum_{i=1}^4(\alpha_i - \alpha^{'}_i(z; b))\exp{\left(-\sum_{j=1}^D A_{ij}(x_j - P_{ij})^2\right)}
\end{equation}
where for Hartmann-3, 
\begin{equation*}
A =     
\begin{bmatrix}
3 & 10 & 30\\
0.1 & 10 & 35\\
3 & 10 & 30\\
0.1 & 10 & 35\\
\end{bmatrix},
\quad
P = 10^{-4} \times    
\begin{bmatrix}
3689 &1170 &2673\\
4699 &4387 &7470\\
1091 &8732 &5547\\
381  &5743 &8828
\end{bmatrix},
\end{equation*}
and for Hartmann-6, 
\begin{equation*}
A =     
\begin{bmatrix}
10 &3 &17 &3.5 &1.7 &8\\
0.05 &10 &17 &0.1 &8 &14\\
3 &3.5 &1.7 &10 &17 &8\\
17 &8 &0.05 &10 &0.1 &14\\
\end{bmatrix},
\quad
P = 10^{-4} \times
\begin{bmatrix}
1312 &1696 &5569  &124 &8283 &5886\\
2329 &4135 &8307 &3736 &1004 &9991\\
2348 &1451 &3522 &2883 &3047 &6650\\
4047 &8828 &8732 &5743 &1091 &381
\end{bmatrix}
\end{equation*}
where $\alpha^{'}_i (z; b)) = b(1 - z_i)$. In the original paper, the variable which accounts for the bias between fidelities, $b$, is set to $0.1$. To account for the fact that we only consider a single fidelity variable, we set $z_i = z, \forall i$, and increase the bias terms significantly to create realistic task correlations.

For the \textbf{good correlation} version used in our evaluations, we set $b = 2.5$ with half-normally distributed noise of $\sigma = 2 (1-z)$. \\
The \textbf{bad correlation} version uses $b = 4$ with a noise of $\sigma = 5 (1-z)$.

Tables~\ref{hps:mfh3},~\ref{hps:mfh6} show the search space for this synthetic benchmark.

\begin{table}[htbp] 
    \parbox{0.4\textwidth}{
        \centering
        \begin{tabular}{l l l l} 
            \toprule
            name & type & values & info \\
            \midrule
                X\_0 & continuous & $[0.0, 1.0]$ &  \\
                X\_1 & continuous & $[0.0, 1.0]$ &  \\
                X\_2 & continuous & $[0.0, 1.0]$ &  \\
            \midrule
                z & log integer & $[3, 100]$ & fidelity \\
            \bottomrule
            \\
        \end{tabular}
        \caption{Synthetic Multi-Fidelity Hartmann search space in 3 dimensions.}
        \label{hps:mfh3}
    }
    \hspace{4em}
    \parbox{0.4\textwidth}{
        \centering
        \begin{tabular}{l l l l}    
            \toprule
            name & type & values & info \\
            \midrule
                X\_0 & continuous & $[0.0, 1.0]$ &  \\
                X\_1 & continuous & $[0.0, 1.0]$ &  \\
                X\_2 & continuous & $[0.0, 1.0]$ &  \\
                X\_3 & continuous & $[0.0, 1.0]$ &  \\
                X\_4 & continuous & $[0.0, 1.0]$ &  \\
                X\_5 & continuous & $[0.0, 1.0]$ &  \\
            \midrule
                z & log integer & $[3, 100]$ & fidelity \\
            \bottomrule
            \\
        \end{tabular}
        \caption{Synthetic Multi-Fidelity Hartmann search space in 6 dimensions.}
        \label{hps:mfh6}
    }

\end{table}

The global minimums of these functions are known:
\begin{itemize}
    \item Hartmann-$3$d: $f(x^*) = -3.86278$ at $x^* =$ ($0.114614$, $0.555649$, $0.852547$)
    \item Hartmann-$6$d: $f(x^*) = -3.32237$ at $x^* =$ ($0.20169$, $0.150011$, $0.476874$, $0.275332$, $0.311652$, $0.6573$)
\end{itemize}

\subsubsection{\jahsbench} \label{app:exp-benchmarks-jahs}
\jahsbench~\citep{bansal-neurips22a} is a benchmark consisting of surrogates trained on $140$ million data points of Neural Networks trained on $3$ datasets, namely CIFAR10, Colorectal-Histology, and Fashion-MNIST. They extend the search space beyond the original tabular search space of NAS-Bench-201~\citep{dong-arxiv20a} consisting of purely discrete architectural choices, introducing both hyperparameters and multiple fidelities to create the first multi-multi-fidelity benchmark for deep learning hyperparameter optimization (Table~\ref{hps:jahs}).

Each of the three datasets shares equal search spaces while we fix the fidelity parameters, depth \texttt{N} and width \texttt{W}, to their maximum. We further limit \texttt{Resolution} to a fixed value of $1.0$ out of the three original choices $\{0.25, 0.5, 1.0\}$.
The surrogates provided by \jahsbench do not explicitly model the monotonic constraint that as \texttt{epoch} increase, so should the training cost. In practice, this was found to be insignificant but we state so for completeness.
For these benchmarks, optimizers minimize \texttt{1 - valid\_acc}. 

\begin{table}[htbp] 
    \centering
    \begin{tabular}{l l l l}
    \toprule
    name & type & values & info \\
    \midrule
        Activation & categorical & \{ReLU,Hardswish,Mish\} &  \\
        LearningRate & continuous & $[0.001, 1.0]$ & log \\
        N & constant & $5$ &  \\
        Op1 & categorical & \{0,1,2,3,4\} &  \\
        Op2 & categorical & \{0,1,2,3,4\} &  \\
        Op3 & categorical & \{0,1,2,3,4\} &  \\
        Op4 & categorical & \{0,1,2,3,4\} &  \\
        Op5 & categorical & \{0,1,2,3,4\} &  \\
        Op6 & categorical & \{0,1,2,3,4\} &  \\
        Optimizer & constant & SGD &  \\
        Resolution & constant & $1.0$ &  \\
        TrivialAugment & categorical & \{True,False\} &  \\
        W & constant & $16$ &  \\
        WeightDecay & continuous & $[1e\mbox{-}05, 0.01]$ & log \\
    \midrule
        epoch & integer & $[3, 200]$ & fidelity \\
    \bottomrule
    \\
    \end{tabular}
    \caption{The \jahsbench search space for all 3 datasets, CIFAR10, Colorectal-Histology and Fashion-MNIST.}
    \label{hps:jahs}
\end{table}

\subsubsection{PD1 (\hyperbo{})}  \label{app:exp-benchmarks-pd1}
The PD1 benchmarks consist of surrogates trained on the learning curves of large architectures, spanning both natural language and computer vision tasks.
The original tabular data is obtainable from the output generated by \hyperbo~\citep{wang-arxiv21} using the dataset and training setup of~\cite{gilmer-github2021}, enabling us to test our methods and baselines for low-budget settings, where multi-fidelity methods are most applicable. The hyperparameters considered for the optimization runs were for Nesterov Momentum~\cite{nesterov-smd83a} which constitutes our search space (Table~\ref{hps:pd1-lm1b}-\ref{hps:pd1-imagenet}). All other hyperparameters were fixed according to their training setup and provided data. 

This tabular data consists of 4 collections of optimization records, a grid-like spread of configurations and also those chosen by their optimizer, in both an initial testing phase and a later full experiment phase. To maximize the data available to the surrogate, we utilize all of this data but take care to drop duplicated runs from their test runs. We select these 4 benchmarks out of the available 24, opting to have a variety of tasks, favoring larger models where possible, or tasks that use big batch sizes such as 2048. 

For these benchmarks, each optimizer aims to minimize the \texttt{valid\_error\_rate}.
The hyperparameters listed
are based on the minimum and maximum values found within the original tabular data, rather than the reported ranges by the authors of HyperBO~\cite{wang-arxiv21}. This was to prevent surrogates from being required to extrapolate outside of their training domain.

\begin{itemize}
    \setlength\itemsep{1em}
    
    \item \textbf{lm1b\_transformer\_2048} derives from the optimization trace of a transformer model~\citep{roy-acl21} with batch size of 2048 on the \texttt{l1mb} statistical language modelling dataset~\citep{chelba-corr13}.

    \begin{table}[hbtp]
        \centering
        \begin{tabular}{l l l l}
            \toprule
            name & type & values & info \\
            \midrule
                lr\_decay\_factor & continuous & $[0.010543, 0.9885653]$ &  \\
                lr\_initial & continuous & $[1e\mbox{-}05, 9.986256]$ &  log \\
                lr\_power & continuous & $[0.100811, 1.999659]$ &  \\
                opt\_momentum & continuous & $[5.9e\mbox{-}05, 0.9989986]$ & log \\
            \midrule
                epoch & integer & $[1, 74]$ & fidelity \\
            \bottomrule
            \\
        \end{tabular}
        \caption{The \texttt{lm1b\_transformer\_2048} search space.}
        \label{hps:pd1-lm1b}
    \end{table}
    
    \item \textbf{translatewmt\_xformer\_64} derives from the optimization trace of an \texttt{xformer}~\citep{lefaudeux-github22} with batch size of 64 on the  WMT15 German-English~\citep{bojar-wmt15}.
    
    \begin{table}[hbtp]
        \centering
        \begin{tabular}{l l l l}
            \toprule
            name & type & values & info \\
            \midrule
                lr\_decay\_factor & continuous & $[0.0100221257, 0.988565263]$ &  \\
                lr\_initial & continuous & $[1.00276e\mbox{-}05, 9.8422475735]$ &  log \\
                lr\_power & continuous & $[0.1004250993, 1.9985927056]$ &  \\
                opt\_momentum & continuous & $[5.86114e\mbox{-}05, 0.9989999746]$ & log  \\
            \midrule
                epoch & integer & $[1, 19]$ & fidelity \\
            \bottomrule
            \\
        \end{tabular}
        \caption{The \texttt{translatewmt\_xformer\_64} search space.}
        \label{hps:pd1-wmt}
    \end{table}

    \item \textbf{cifar100\_wideresnet\_2048} dervies from the optimization trace of a wideresnet model~\citep{zagoruyko-arxiv16} with batch size 2048 on the cifar100 dataset~\citep{krizhevsky-tech09a}.
    \begin{table}[hbtp]
        \centering
        \begin{tabular}{l l l l}
            \toprule
            \textbf{name} & \textbf{type} & \textbf{values} & \textbf{info} \\
            \midrule
                lr\_decay\_factor & continuous & $[0.010093, 0.989012]$ &  \\
                lr\_initial & continuous & $[1e\mbox{-}05, 9.779176]$ &  log \\
                lr\_power & continuous & $[0.100708, 1.999376]$ &  \\
                opt\_momentum & continuous & $[5.9e\mbox{-}05, 0.998993]$ &  log \\
            \bottomrule        
        \end{tabular}
        \caption{The \texttt{cifar100\_wideresenet\_2048} search space.}
        \label{hps:pd1-cifar}
    \end{table}
    
    \item \textbf{imagenet\_resnet\_512} dervies from the optimization trace of a resnet model~\citep{he-cvpr16a} with batch size of 512 on the imagenet dataset~\citep{russakovsky-ijcv15a}.
    \begin{table}[hbtp]
        \centering
        \begin{tabular}{l l l l}
            \toprule
            \textbf{name} & \textbf{type} & \textbf{values} & \textbf{info} \\
            \midrule
                lr\_decay\_factor & continuous & $[0.010294, 0.989753]$ &  \\
                lr\_initial & continuous & $[1e\mbox{-}05, 9.774312]$ &  log \\
                lr\_power & continuous & $[0.100225, 1.999326]$ &  \\
                opt\_momentum & continuous & $[5.9e\mbox{-}05, 0.998993]$ & log  \\
            \bottomrule
        \end{tabular}
        \caption{The \texttt{imagenet\_resnet\_512} search space.}
        \label{hps:pd1-imagenet}
    \end{table}

\end{itemize}

\textbf{Training surrogates on the PD1 tabular data} The original data is a mix of several datasets, models, and their parameters for which we do some preprocessing. All data-preprocessing is available as part of \mfpbench and consists of:
\begin{enumerate}
    \item Splitting the raw data by all available $\{dataset name, model, batch size\}$ subsets.
    \item Identify which columns are hyperparameters by those being marked as such and consist of more than one unique value.
    \item Drop all columns which are not hyperparameters or metrics.
    \item Drop all \nan values for which no metrics are recorded.
    \item Drop configurations that recorded \textit{divergent} training costs.
    \item Drop duplicated entries, preferring to keep those from their full experimental runs.
\end{enumerate}

To decide if a configuration diverged was to find outliers that reported extreme outlier costs, with a cutoff applied heuristically to each individual dataset, the details of which can be found within \mfpbench. This was done to ease the learning process of the surrogate model and to remove emphasis on these outliers. The resulting surrogate is no longer aware of these divergences and offers smooth interpolation for the training cost for these configurations. 
 
Once the datasets are prepared, we then train a single surrogate XGBoost model~\citep{chen-kdd16a} per metric recorded.
This training was performed using DEHB~\citep{awad-ijcai21a}, optimizing for the mean R2 loss of 5-fold cross-validation for a total of 4 hours, 8 CPU cores, and the seed set to $1$.  All surrogate models were found to converge in their R2 loss. These surrogates are available as part of \mfpbench for further inspection. While certainly improvements can be made in this modeling phase, for the purpose of our experiments, they offer a good approximation of the entire optimization landscape.

\subsubsection{\yahpogym}  \label{app:exp-benchmarks-yahpo}
The \yahpogym~\citep{pfisterer-arxiv21a} collection is a large collection of multi-fidelity surrogates across a wide range of tasks, including traditional machine learning models with dataset size as a fidelity, as well as Neural Network, benchmarks such as LCBench and NAS-Bench-301~\citep{siems-arxiv20a}.

\yahpogym provides surrogates trained on a shared search space between all of these tasks.
We ignore the rest of the available benchmarks from \yahpogym as the others consist of non-DL tasks or contain conditional search spaces which are not suitable for most of our baselines. 
For these benchmarks, the optimization objective was to minimize \texttt{1 - val\_balanced\_accuracy}.
Table~\ref{hps:lcbench} shows the search space for these $5$ benchmarks.

\begin{table}[htbp]
    \centering
    \begin{tabular}{l l l l}
    \toprule
    name & type & values & info \\
    \midrule
        batch\_size & integer & $[16, 512]$ & log \\
        learning\_rate & continuous & $[0.0001, 0.1]$ & log \\
        max\_dropout & continuous & $[0.0, 1.0]$ &  \\
        max\_units & integer & $[64, 1024]$ & log \\
        momentum & continuous & $[0.1, 0.99]$ &  \\
        num\_layers & integer & $[1, 5]$ &  \\
        weight\_decay & continuous & $[1e\mbox{-}05, 0.1]$ &  \\
    \midrule
        epoch & integer & $[1, 52]$ & fidelity \\
    \bottomrule
    \\
    \end{tabular}
    \caption{The \texttt{lcbench} search space.}
    \label{hps:lcbench}
\end{table}

\begin{figure}[htbp]
    \centering
    \includegraphics[width=0.9\textwidth]{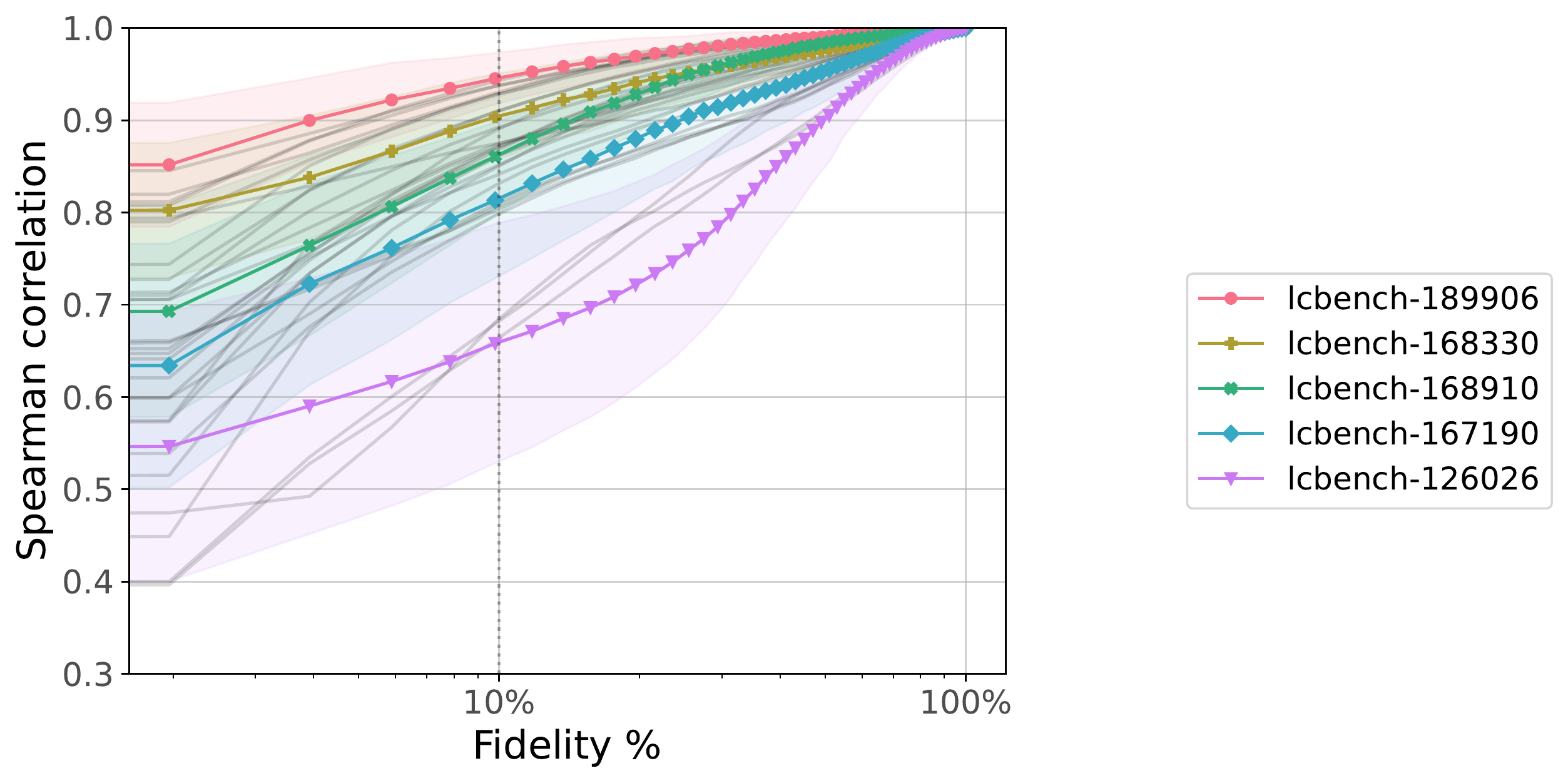}
    \caption{Each curve shows the spearman-correlation of $25$ configurations from the given fidelity along the x-axis to the last fidelity, where the standard deviation is estimated with repeated samples until the mean curve converges to within a $0.001$ Euclidean distance update to the previous mean. The highlighted lines are the LCBench tasks selected by taking the $(0, 0.25, 0.5, 0.75, 1)$ quantiles of the correlations at $10\%$ of the full fidelity range. The faded gray lines with no markers represent LCBench tasks not selected.}
    \label{fig:correlation_curves_lcbench}
\end{figure}

For our experiments, we choose 5 LCBench tasks from $34$ from OpenML~\citep{bischl-arxiv17a} according to the Spearman rank correlation of configurations at the $10\%$ epoch and the final epoch $z_{\text{max}}$. The chosen tasks, 126026, 167190, 168330, 168910, and 189906 are equally spaced according to this correlation and include the task featuring the least (126026) and most (189906) correlation. The correlation for all LCBench tasks can be seen in Figure~\ref{fig:correlation_curves_lcbench}.  

\subsubsection{Classifying benchmarks into high-low correlation} \label{app:cluster-benchmarks}
\begin{figure}[htbp]
    \centering
    \includegraphics[width=0.9\textwidth]{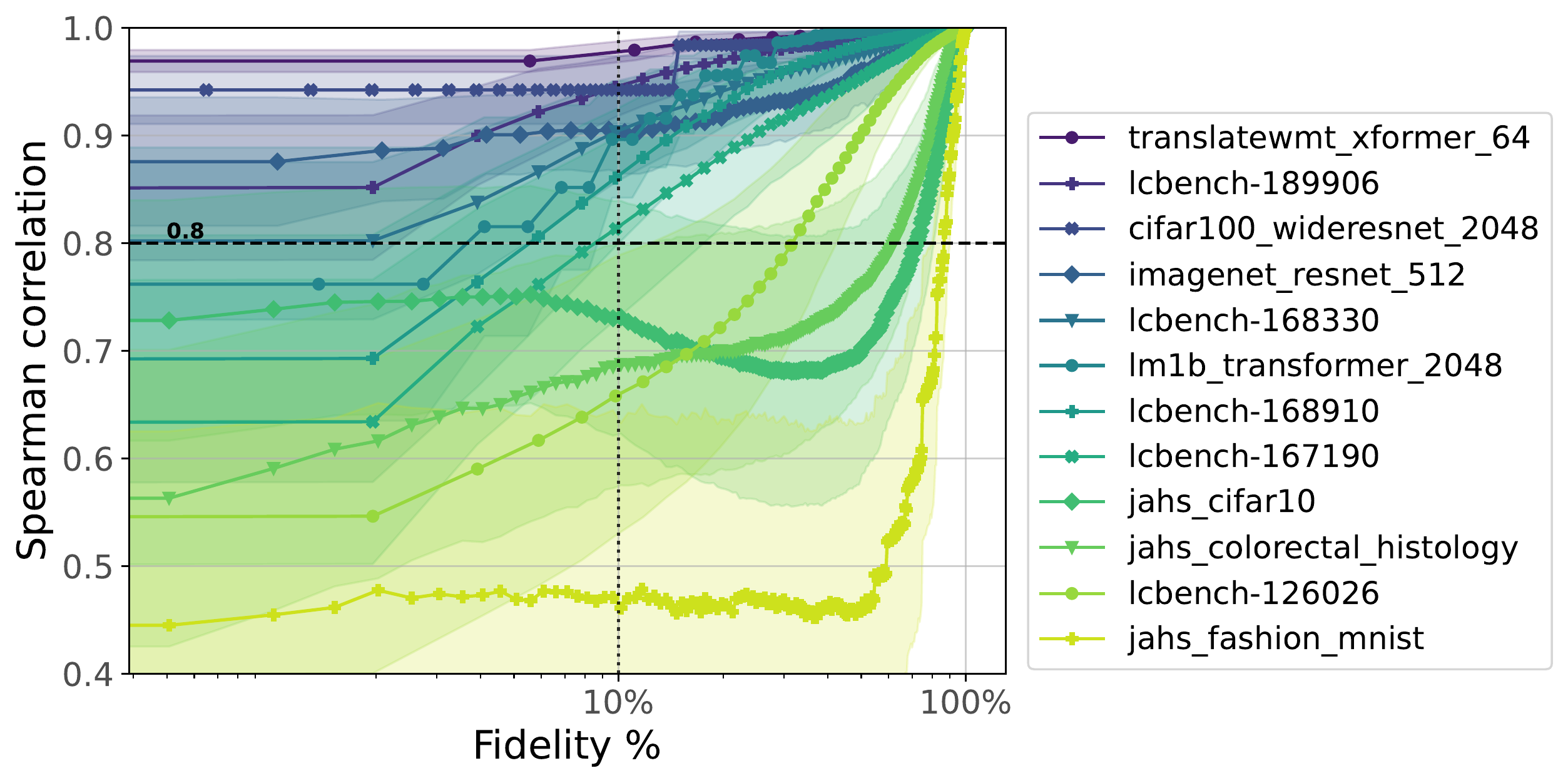}
    \caption{Each curve shows the spearman-correlation of $25$ configurations from the given fidelity along the x-axis to the last fidelity, where the standard deviation is estimated with repeated samples until the mean curve converges to within a $0.001$ Euclidean distance update to the previous mean. Our good/bad correlation definition corresponds to a spearman-correlation cutoff of $0.8$ at $10\%$ of the maximum budget. The legend is sorted by their correlation at $10\%$ fidelity and corresponds to the order of benchmarks in Figure~\ref{fig:rs-hb-corr}.}
    \label{fig:correlation-curves-good-bad}
\end{figure}

Our definition of a \textit{high} or a \textit{low} correlation benchmark is that the benchmark must have $0.8$ spearman correlation of rankings at $10\%$ of the maximum fidelity to be classified as \textit{high}, otherwise it is classified as \textit{low}. We depict this in Figure~\ref{fig:correlation-curves-good-bad}, showing the spearman correlation between each fidelity available and the final full fidelity. While other cutoffs and classifications are possible, given our suite of benchmarks, we find this to be a reasonable separation given the data.

\begin{figure}[htbp]
    \centering
    \includegraphics[width=\textwidth]{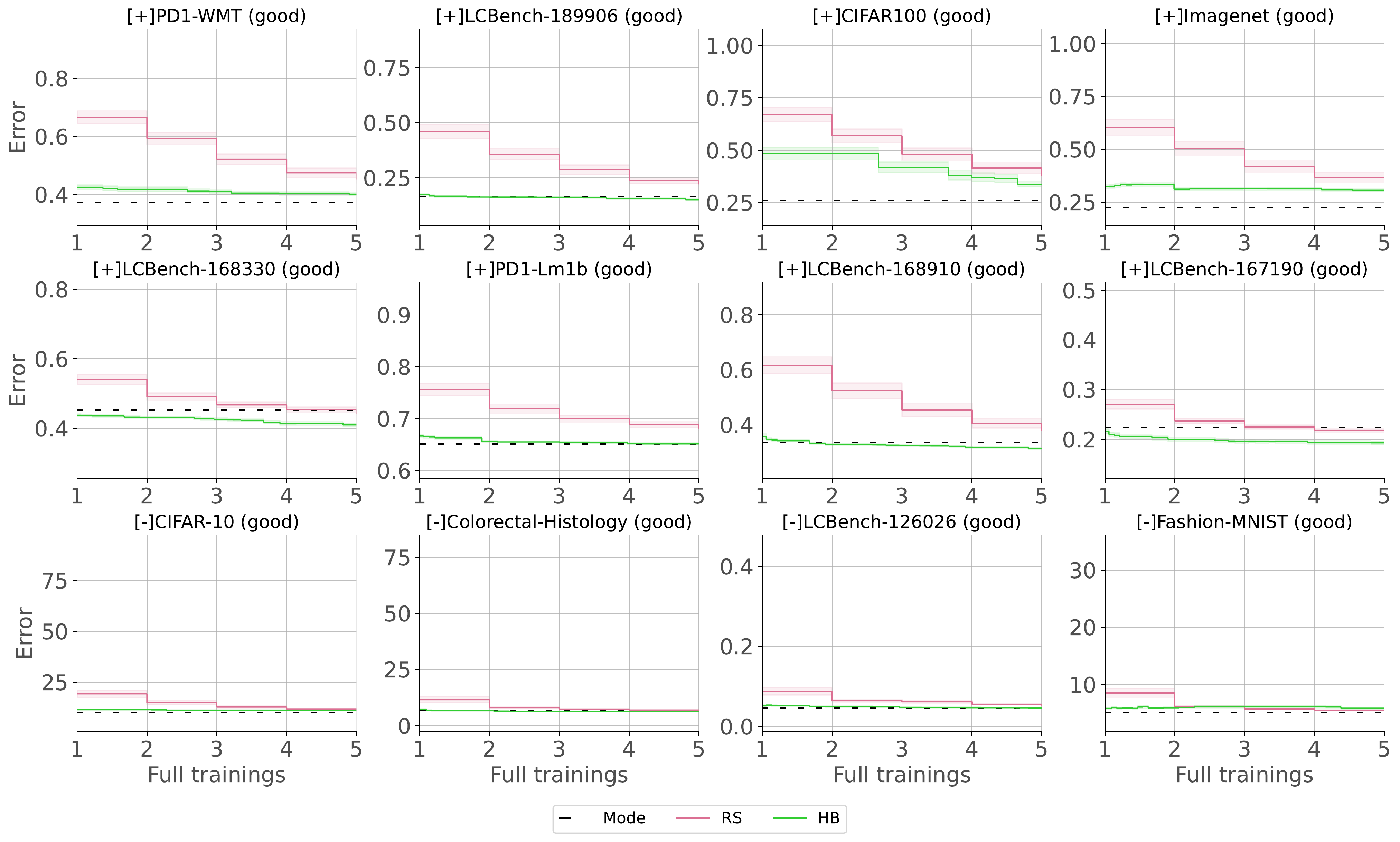}
    \caption{Comparing Random Search and HyperBand over $5 \times$ budget to gauge quality of performance correlation across fidelities. We classify benchmarks marked with $[+]$ as \textit{high correlation} benchmarks (Row 1 and 2). The benchmarks marked as $[-]$ are classified as \textit{low correlation} benchmarks (Row 3).}
    \label{fig:rs-hb-corr}
\end{figure}

To further motivate this choice, Figure~\ref{fig:rs-hb-corr} shows the performance of random search and HyperBand across these $12$ benchmarks. 
We choose HyperBand as under the hood, it runs different instantiations of \sh{}. 
Under a single worker-run, \hb{} becomes a sequential run of \sh{} with increasing $\fid\min$.
Strong performance of \sh{} requires a high correlation of performance across fidelities.
Thus in $5\times$ in Figure~\ref{fig:rs-hb-corr}, wherever \hb{} the performance gap between random search and hyperband is not pronounced in the short budget regime shown, it can be inferred that the lower fidelities do not provide reliable prediction when early stopping configurations.
Based on the classification strategy derived from Figure~\ref{fig:correlation-curves-good-bad}, we denote high correlation benchmarks with a $(+)$ and low correlation benchmarks with a $(-)$ in Figure~\ref{fig:rs-hb-corr}.
The comparison of random search and \hb{} in this figure reflects the reasonable correctness of our classification strategy.

\subsubsection{Issues with Benchmarks in parallel setting} \label{app:failing-benchmarks}
During our parallel worker runs, we noticed that workers running on LCBench (see~\ref{app:exp-benchmarks-yahpo}) would silently drop out. This is an artifact of \yahpogym, where the shared loading of system resources does not play nicely with workers being started in parallel. The optimization runs would still progress without issue but only utilize 1-3 processes instead of the deployed 4. The degree to which the problem occurred was minimal but likely to impact aggregated results. As a safety precaution, we remove LCBench from our parallel algorithm evaluations to prevent undue biases from leaking into our evaluations.


\subsection{Baselines} \label{app:exp-baselines}

For fair comparison, customizability and certain technical constraints, we reimplement all the baselines listed below, other than BOHB. 

\begin{figure}
    \centering
    \includegraphics[width=0.75\textwidth]{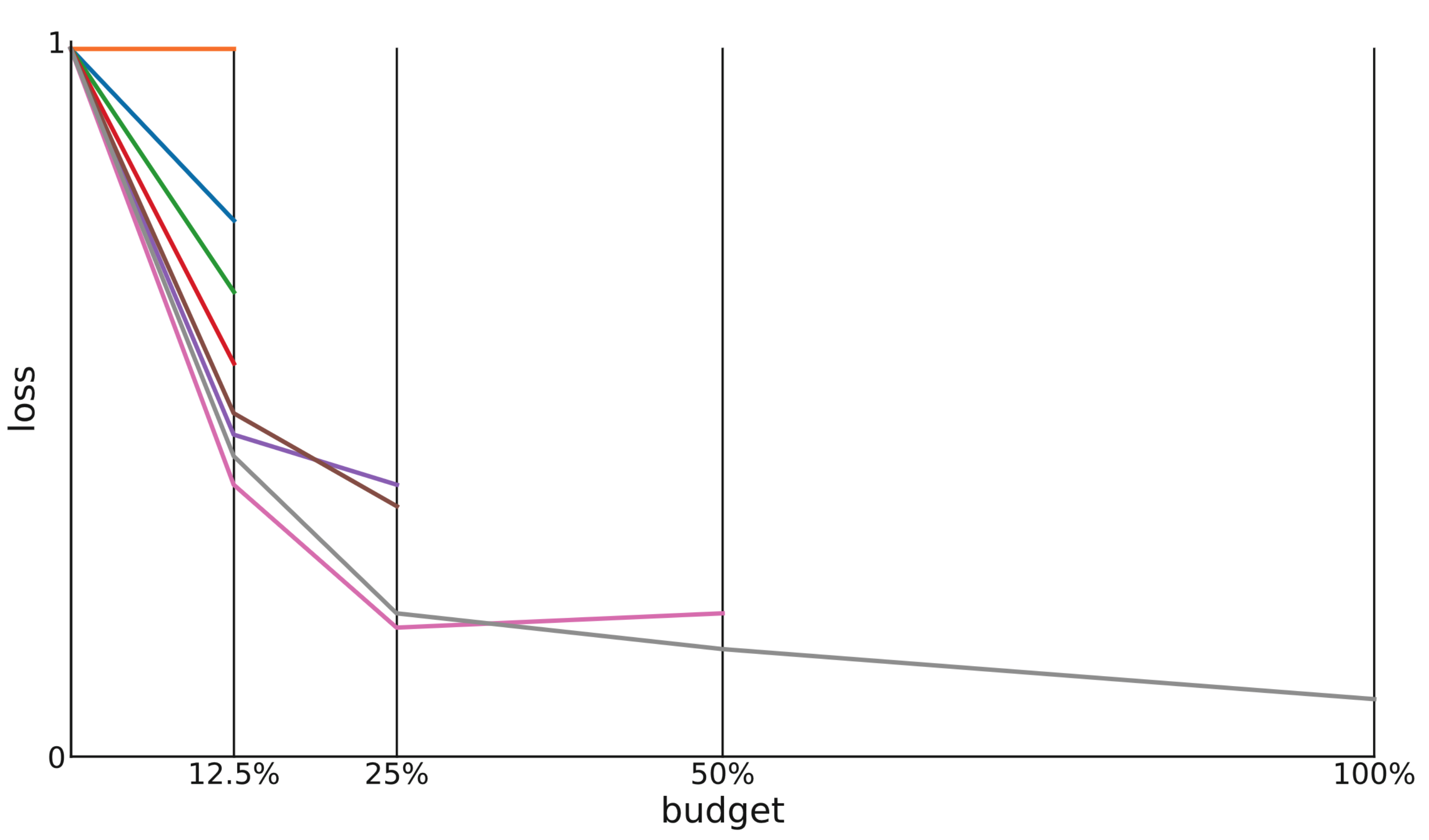}
    \caption{Illustration of \shlong{} for eight algorithms/configurations. 
    After evaluating all algorithms on $1/8$ of the total budget, half of them are dropped and the budget given to the remaining algorithms are doubled.}
    \label{fig:sh-book}
\end{figure}

\textbf{HyperBand} We implement our own version of \hb{}\citep{li-jmlr18a} to allow for the input of priors. 
We verified our implementation with the popular \hb{} implementation provided in BOHB~\citep{falkner-icml18a}. 
We use $\eta=3$ for all experiments with the minimum and maximum budget coming as an input from the problem to solve, in this case, benchmarks.
As described in Section~\ref{sec:bg}, \hb{} iterates over different instantiations of \shlong{} (\sh{}).
In Figure~\ref{fig:sh-book} we show an example\footnote{image and caption sourced under CC-BY-4.0 from \url{https://www.automl.org/wp-content/uploads/2019/05/AutoML_Book.pdf}} for an \sh{} run under $\eta=2$.

\textbf{Bayesian optimization} is a prominent framework for Hyperparameter Optimization~\citep{snoek-nips12a,kiili-neurips20a}, hence we choose its Gaussian Processes (GP) implementation as a model-based competitor. 
We incorporate expert priors to the optimization following the~$\pi$BO algorithm~\citep{hvarfner-iclr22a}. 
In a low-budget setting, model-based search proves challenging for high-dimensional search spaces (e.g. \jahsbench) as common practices require the number of initial random observations equal to the search space dimensionality. 
To allow model-based search in BO and $\pi$BO we set their initial design size to $10$, to allow model fitting under our experiment budgets.
Expected Improvement is used as the acquisition function across all BO algorithms.

\textbf{BOHB}~\citep{falkner-icml18a}\footnote{Implementation from HpBandSter: \url{https://github.com/automl/HpBandSter}} incorporates multi-fidelity \hb{} into the Bayesian optimization framework by building KDE models on each fidelity level to efficiently guide the search. 
The official implementation has no direct way of accepting a prior distribution over optimal configurations and incorporating it into search. 
We keep the other default settings intact. 

\textbf{ASHA}~\citep{li-mlsys20a} was developed as an extension to \shlong{} that can run on massively parallel systems, designed to minimize idle workers. 
ASHA modified \sh{} and \hb{} to allow for a configuration to be promoted as soon as $\eta$ configurations have been seen at a fidelity, calling such a step an \textit{asynchronous promotion}.
Thereby, a free worker need not remain idle till an entire predefined number of configurations have finished evaluation at a fidelity, like vanilla-\sh{} or \hb{}.
Since \hb{} effectively runs multiple \sh{} brackets, asynchronous-\hb{} can be designed as \hb{} running multiple ASHA brackets.
However, for this work, we chose the sampling distribution for the brackets as used in~\citet{klein-arxiv20a} for asynchronous-\hb{}.

\textbf{Mobster}~\citep{klein-arxiv20a} extended asynchronous-\hb{} with a surrogate model to create an asynchronous version of BOHB.
The asynchronous design aside, Mobster is different from BOHB as it uses Gaussian Processes as the surrogate model, unlike BOHB which uses Tree Parzen Estimators (TPE).

\paragraph{ESP based baselines.} We construct multiple baselines that do not explicitly exist in the literature but serve as important baselines for a fair comparison and a deeper understanding of the problem and the method.

\begin{itemize}
    \item \textit{[X]+Prior} methods: for such a method, the uniform random search component in X is replaced with sampling from the prior. If accompanied by a Y\%, that indicates the percentage of prior-based sampling, with (1 - Y)\% for random sampling. For example, \hb{}+Prior is \hb{} but with only sampling from prior and \hb{}+Prior($30\%$) would indicate that sampling from prior happens with a probability of $0.3$ and random sampling with a probability of $0.7$.
    \item \textit{[X]+\esp{}} methods: for such a method, the uniform random search component in X is replaced with sampling from the ESP, \E{}.
    \item \algo{}+BO: this runs \algo{} as described in Section~\ref{sec:method} for a budget equivalent to an initial design of $10$, and then switching to sampling a configuration from the acquisition in a BO loop, disabling the ESP.
\end{itemize}

\paragraph{Parallel runs.} We run each benchmark-optimizer-seed combination over $4$ workers. 
For \hb{} based algorithms, if a worker is free, an evaluation from the next \sh{} bracket is already started, in order to maximize worker efficiency. However, a pending evaluation from the earliest active \sh{} bracket has the highest priority.
For BO algorithms, under our parallel setup for multi-fidelity optimization, we did not require batch acquisitions. However, we needed to account for incomplete evaluations from configurations that are still training when fitting the surrogate. We do this simply by fitting the surrogate on the finished evaluations and predicting the mean for the pending evaluations, before retraining on the required set of configurations\footnote{unlike typical BO, multi-fidelity BO may have different data subsets to fit the surrogate over}.

\subsubsection{GP kernels} \label{app:algo-gp}

\paragraph{Numerical kernel.} In the numerical continuous and discrete domain, we use the standard~(\cite{snoek-nips12a, misc-botorch-balandat2020}) \textit{Matérn}-5/2 kernel~\citep{matern2013spatial}. For $D$-dimensional numerical inputs $x$ and $x'$,
\begin{equation}
    \label{eq:matern52}
    k_{M5/2}(x, x') = \theta_{0} \left(1 + \sqrt{5r^{2}(x, x')} + \frac{5}{3}r^{2}(x, x')\right) \exp\left(-\sqrt{5r^{2}(x, x')}\right),
\end{equation}
where $r^{2}(x, x') = \sum_{d=1}^{D}(x_{d} - x'_{d})^2 / \theta^{2}_{d}$ denotes a scaled Euclidean distance between points, and $\theta_{d}$ is a dimension-specific lengthscale.

\paragraph{Categorical kernel.} A straightforward extension of a Mat\'{e}rn kernel for a categorical domain is to use \textit{1-in-K encoding}. However, this solution increases the dimensionality of the input, which might result in poor regression performance. Instead, we follow~\citep{bo-kernels-hutter2013} who propose direct handling of categorical inputs by computing a weighted Hamming distance:
\begin{equation}
   k_{HM}(x, x') = \exp \left(\sum_{d=1}^{D} \left(-\theta_{d} \cdot \mathbb{1}_{x_d \neq x'_d}\right)\right)
\end{equation}


\subsection{Generating priors} \label{app:priors-gen}

We borrow the prior generation procedure in~\citet{hvarfner-iclr22a} to generate the \textit{near optimum} and \textit{bad} priors.
Additionally, we construct another class of \textit{good} priors to reflect a different strategy that may be closer to practice, since for \dl{}, the optimum is generally not known.
We thus generate three kinds of priors, \textit{near optimum}, \textit{bad}, and \textit{good}, the first two to simulate boundary conditions of priors that may be received and \textit{good} to simulate a practitioner with some prior knowledge of a good configuration to choose. In each of these cases, the prior distribution is a normal $\mathcal{N}(\config, \sigma^{2})$, with $\sigma=0.25$, where the configuration $\config$ is generated by the following processes: 

\paragraph{Near optimum:} Given we don't know the optimum configuration for a benchmark, we approximate this by taking the best of $50,000$ configurations according to their observed loss value.
During each seeded run, we perturb this configuration by Gaussian noise with a $\sigma = 0.25$ for numerical values, while using this same $0.25$ for uniformly selecting a different categorical value. This is done to ensure there is still some room for improvement possible over the initial prior configuration. All algorithms receive the same prior configuration as defined by the seed.
In the case of the Multi Fidelity Hartmann benchmarks, we can access the optimum of the function analytically and so we take this optimum as the configuration which is to be perturbed.
    
\paragraph{Good:} We define a \textit{good} prior as one that is suggested by limited prior evaluations done by a practitioner.
For this, we take the best of $25$ configurations and apply no further noise modifications per seed.
This means each run will see the same prior irrespective of the prior.
We choose a budget of $25$ random samples since in our experimental setup our focus is strictly on HPO budgets under $20\times$.
As we show later, the relative quality of this class of priors varies per benchmark, compared to the near-optimum priors.

\paragraph{Bad:} As we don't know the worst configuration for each benchmark, we simulate this process by taking the worst configuration of $50,000$ samples. No additional noise is added per seed, to ensure we always treat the worst known configuration as the prior input.

\begin{figure}[htbp]
    \centering
    \includegraphics[width=\textwidth]{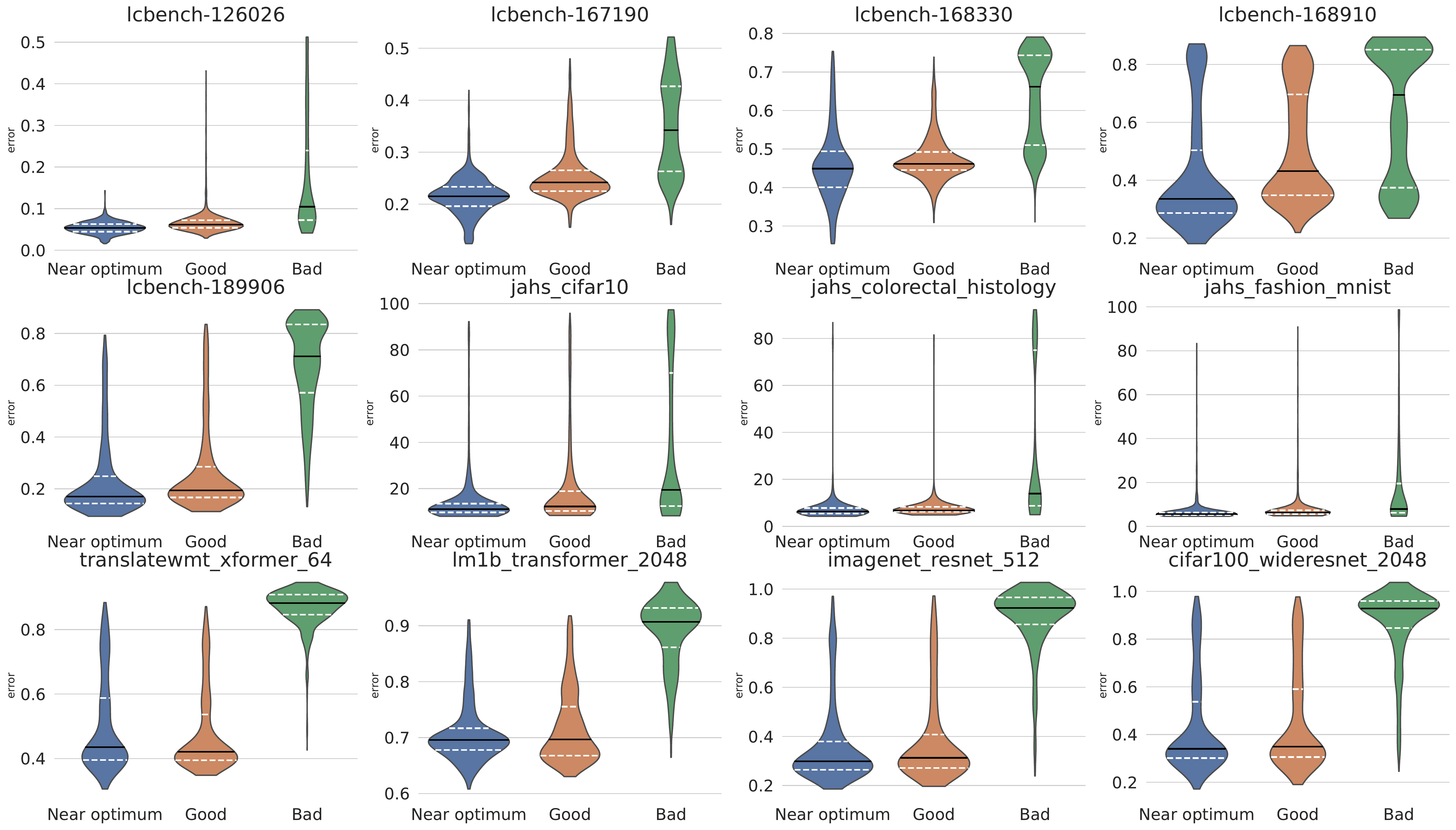}
    \caption{Each violin shows the density of performances achieved by random search on a distribution using the given prior. 
    The black line in the violin is the mean with the dashed white lines indicating the 25\% and 75\% quartiles. 
    In all cases, we see \textit{near optimum} and \textit{good} outperform the \textit{bad} prior, as expected. 
    In the case of the PD1 datasets (bottom row), some \textit{good} priors achieve a better mean error than \textit{near optimum} but their best configurations found are not as strong as they are when using \textit{near optimum} priors. 
    This could be a result of the near optimum prior being perturbed each run.
    Some iterations may put the prior at a slightly worse region than the fixed good prior.
    However, the near-optimum priors generally achieve the best-seen score.
    }
    \label{fig:prior-violin-plots}
\end{figure}

It is worth noting that while \textit{near optimum} and \textit{bad} are the best and worst of $50,000$ samples, respectively, it need not be that these configurations are also the best and worst for search. For benchmarks with a high categorical count, such as JAHS Bench as described in~\ref{app:exp-benchmarks-jahs} (middle row in Figure~\ref{fig:prior-violin-plots}), simply switching a categorical value can drastically alter the performance of the configuration. More generally, a single point of the search space does not immediately provide information about the performance in its locality. As such, bad performing configurations could be reached with the noise perturbations of the \textit{near optimum} prior.

To investigate the performance impacts of each prior, in Figure~\ref{fig:prior-violin-plots} we plot the distribution of performances of $25$ configurations sampled from the prior distribution used. This is additionally done for $50$ seeds, giving a total of $1,250$ configurations plotted per violin. 

\subsection{Experimental setup} \label{app:exp-setup}

Generally, the \hpo{} landscape for a new task is unknown and the quality of the expert prior input cannot be gauged without previous run data available.
In order to simulate how an expert may use \hpo{} in their task, we design and present the experiments to represent such scenarios.
We show the aggregated results (\textit{All}) over different qualities of prior input to highlight robustness across prior qualities, which may be unknown at the beginning of the problem. 
Subsequently, we break the results down into performances under the \textit{near optimum} priors, \textit{good} prior, and \textit{bad} prior.
This setup intends to highlight that \algo{}'s performance benefits are more substantial if the quality of the prior is better.
Our experiment design substantiates the hypothesis that \algo{} and algorithms extended with \E{} are robust across any given prior input, a property not held by our baseline algorithms.  

The metrics reported for single-worker runs are for $50$ seeds across $12$ benchmarks, $5$ from LCBench (\ref{app:exp-benchmarks-yahpo}), $3$ from \jahsbench (\ref{app:exp-benchmarks-jahs}) and $4$ from PD1 (\ref{app:exp-benchmarks-pd1}).
For the parallel case, we run $4$ workers for each algorithm, $10$ seeds, for only $7$ \footnote{We exclude 5 LCBench benchmarks due to parallelism issues as described in~\ref{app:failing-benchmarks}} of these benchmarks. 
We chose $10$ seeds due to the increased cost of running many workers\footnote{asynchronity is simulated for the queries to the benchmarks by sleeping for \textit{epoch} seconds}. 

\subsection{Evaluation protocol}  \label{app:exp-evaluation} 

For the reduction factor $\eta$, in \hb{} and thereby \algo{}, we chose $\eta=3$.
For the various $[\fid\min, \fid\max]$ coming from the $12$ benchmarks chosen, there are $2$, $3$, or $4$ levels of fidelity created (rungs) in \hb{} for our set of experiments. 
Therefore, for the single worker case, we show a budget of $12$ full function evaluations, which is approximately equivalent to the budget of one complete \hb{} iteration with $4$ fidelity levels ($12 \times \fid\max$).
For other cases, we show a budget of $20\times$ and report under $5\times$ for multi-worker runs.

The evaluation setup is motivated by how an HPO algorithm might be used in practice. 
The HPO algorithm stopped anytime, will return the best-seen configuration, i.e., will return the current incumbent. 
To compare algorithms, we plot the incumbent over the budget spent in epochs. 
The incumbent is simply chosen as the configuration with the lowest error across all fidelities.
The average relative rank plots are computed over the validation error of the incumbent at $\fid\max$, as obtained from the benchmark. 
For each seed, we obtain the ranks of algorithms for each benchmark, averaging them to get an estimate of each algorithm's robustness across tasks.
By computing the mean and standard error across seeds, we obtain an expectation of the algorithm's performance with respect to its stochastic components and other variations derived from seeding.
We show all relative ranks only after the $1\times$ budget has been exhausted.

We compute ranks over the error or loss on the validation sets, where the choice of benchmark determines the exact metric to be minimized.
Given we aggregate over the average ranks on a benchmark, the choice of per benchmark metric is irrelevant.
For all prior-based baselines, we evaluate the mode of the prior distribution at $\fid\max$ as the first evaluation.

\section{Algorithm details} \label{app:algo}


\subsection{Pseudocode} \label{app:algo-pseudo}

Algorithm~\ref{alg:hb} is the main \hb{} loop and the optimizer interface for \algo{}.
Depending on the actual framework where this is implemented, Lines 8-9 may be arranged differently.
L8 is an asynchronous call that allows \hb{} and by extensions \algo{} to be parallel since different SH brackets can be scheduled simultaneously.
That is, it can begin a new \sh{} bracket even if L8 has not returned the sampled and evaluated base rung of the current \sh{} bracket.
L4 in Algorithm~\ref{alg:samp-eval} can also be an asynchronous call, allowing multiple configurations from L3 to be evaluated in parallel.
The pseudo-code for \hb{} (Algorithm~\ref{alg:hb}) thus presents itself in a way where the ensemble sampling policy \E{} replaces vanilla random search, for the lowest fidelity (base rung) in the current \sh{} bracket in \hb{} (L3 in Algorithm~\ref{alg:samp-eval}).
Once the samples are collected and evaluated, vanilla-\sh{}'s promotion can be run normally to survive and continue training chosen configurations.


\begin{algorithm}[H]
  \caption{\hb{} base for \algo{}}
  \label{alg:hb}
  \begin{algorithmic}[1]

	\State {\bfseries Input:} Distribution over optimum $\pi(\cdot)$, halving parameter $\eta$, resource bounds $[z_\text{min}, z_\text{max}]$. 
	\State $s_\text{max} \xleftarrow{}   \lfloor \log_\eta(z_\text{max}/z_\text{min}) \rfloor$
        \State $\mathcal{H} \leftarrow \emptyset$

        \While{$budget\ remains$}
        
    	\For {$s \in \{s_\text{max},\, \ldots,\, 0\}$}
     
         	\State {$n \xleftarrow{} \left\lceil\frac{(s_\text{max} + 1)}{s+1} \cdot {\eta^s}\right\rceil$,\quad $\fid \xleftarrow{} z_\text{max} \cdot \eta^{-s}$} 
                \State \rung{} $= s_\text{max} - s$
                \State $\mathcal{H} \leftarrow $ Alg.~\ref{alg:samp-eval}($s, s_\text{max}, \eta, n, \fid, \pi, \mathcal{H}$) 
                \Comment{sample and evaluate $n$ times}
                \State $\mathcal{H} \leftarrow \text{Alg.~\ref{alg:sh}}(s, s_\text{max}, \eta, \fid, \mathcal{H})$  
                \Comment{run \sh{}-based promotions}
            \EndFor

        \EndWhile
        
\end{algorithmic}
\end{algorithm}


\begin{algorithm}[H]
   \caption{Sample and evaluate $n$ configurations}
   \label{alg:samp-eval}
   
    \begin{algorithmic}[1]
        \State {\bfseries Input:} $s, s_\text{max}, \eta, \text{number of samples } n, \text{evaluation resource } \fid,  \text{prior}\pi(\cdot), \text{observations } \mathcal{H}$
        \For {$i \in \{1, 2, \ldots, n\}$}
    
            \State $\config \leftarrow $ Alg.~\ref{alg:ens-sampler}$(s, s_\text{max}, \eta, \mathcal{H}, \pi(\cdot) )$
            \Comment{sampling from \uniform{} here runs vanilla-\hb{}}
            \State $y \leftarrow evaluate(\config, \fid)$
            \State $\mathcal{S} \leftarrow \mathcal{S} \cup \{(\config, y, \fid)\}$
        \EndFor
        \State {\bfseries return} $\mathcal{S}$
        
    \end{algorithmic}
\end{algorithm}


   
    

        


\begin{algorithm}[H]
   \caption{Perform a \sh{} iteration given the base rung}
   \label{alg:sh}
   
    \begin{algorithmic}[1]
        \State {\bfseries Input:} $s, s_\text{max}, \eta, \text{minimum evaluation resource } \fid, \mathcal{H}$
    
        \State $rung = s_\text{max} - s$
        \State $(\hat{X}, \hat{Y}) \leftarrow$ retrieve all observations from $r$ in $\mathcal{H}$
        \For {$\hat{s} \in \{ s-1, \ldots, 0 \}$}
            \State $\fid = \eta \cdot \fid$
            \Comment{the next higher fidelity}
            \State $\hat{X} \leftarrow top\_1/\eta(\hat{X}, \hat{Y})$
            \Comment{the \topeta{} configurations in $\hat{X}$}
            \State $\hat{y} \leftarrow evaluate(\config, \fid)\text{, } \forall \config \in \hat{X}$
            \State $\mathcal{S} \leftarrow \mathcal{S} \cup \{(\hat{X}, \hat{Y}, \fid)\}$
        \EndFor 
        \State {\bfseries return} $\mathcal{S}$
    
    \end{algorithmic}
\end{algorithm}


\subsection{Ablations} \label{app:algo-ablations}

This section contains ablations over the various design choices in \algo{} (Section~\ref{sec:priorband}).
All ablations are compared over different qualities of prior strength.
Additionally, the set of benchmarks is also grouped into combinations of high-low correlation of lower fidelity to the $\fid_\text{max}$ and good-bad priors as discussed in Appendix~\ref{app:cluster-benchmarks} and~\ref{app:priors-gen}.
In Sections~\ref{app:algo-abl-wp}-\ref{app:algo-abl-hs} the next $3$ sections we show ablations over different choices for the main design components in ESP for \algo{}.
In Appendix~\ref{app:algo-abl-to-inc-or-not-to-inc} we show the benefits of incumbent sampling in \algo{}.

\subsubsection{Random sampling proportions} \label{app:algo-abl-wp}

\begin{figure}
    \centering
    \begin{tabular}{c}
         \includegraphics[width=0.9\textwidth]{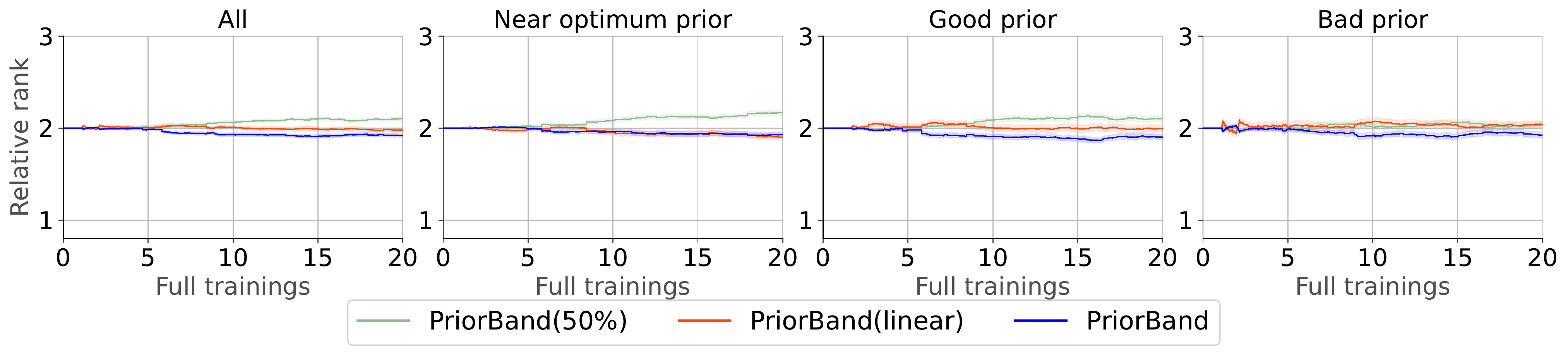} \\
         \includegraphics[width=0.9\textwidth]{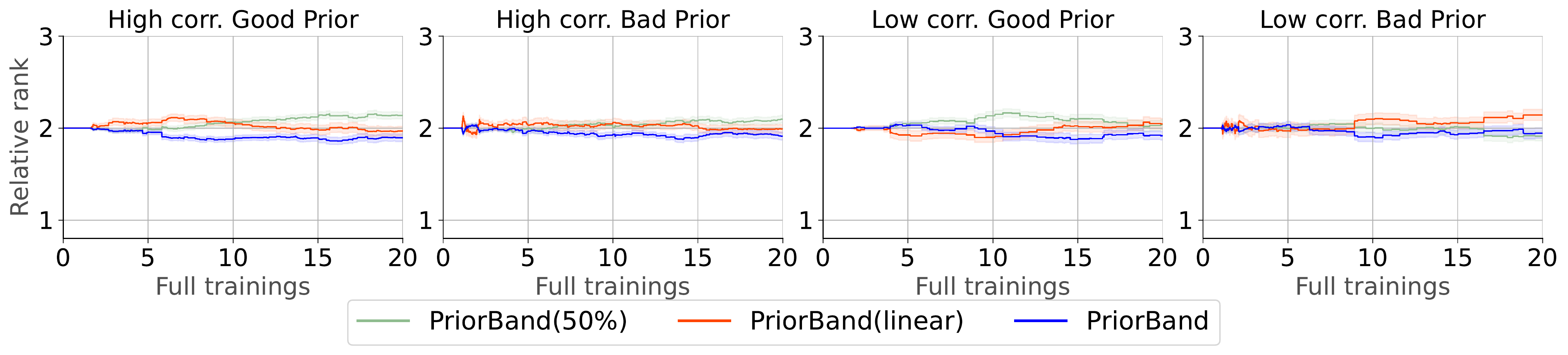}
    \end{tabular}
    \caption{A comparison of \algo{} with 3 different strategies for setting the proportion of random samples in ESP as a fixed function of the rung.
    PriorBand(50\%) fails to meaningfully utilize the \textit{near optimum} and \textit{good} prior when compared to both linear and geometric decay, with no aggregated benefit seen at any budget. 
    PriorBand(linear) is an equivalent heuristic to PriorBand(linear) in intuition and simplicity, but overall \algo{}'s geometric decay comparatively performs robustly across all scenarios.
    }
    \label{fig:abl-prior}
\end{figure}

Section~\ref{sec:decay-random-sampling} describes how the proportion of random samples is traded-off with the proportion of prior sampling across fidelities.
In Figure~\ref{fig:abl-prior} we compare $2$ other designs for how the proportion of random and prior samples can be determined.
\algo{}($50\%$) uses a fixed heuristic where $50\%$ of the samples at each rung is going to be a random sample.
\algo{}(linear) is similar to \algo{}, where instead of a geometric decay (Section~\ref{sec:decay-random-sampling}) of random sample proportions like the latter, a linear decay is applied.
That is, at the lowest rung, \pp{} $=$ \pu{}, and at the highest rung, \pp{}~$= \eta \cdot$~\pu{}. 
For all intermediate rungs, the relationship is derived from this interpolated line.

\subsubsection{Prior and incumbent-based sampling trade-off}  \label{app:algo-abl-inc}

The choice of incumbent-prior trade-off strategy (Section~\ref{sec:incumbent-sampling-weightage}) is crucial to the robustness of \algo{}.
As the simplest design, we kept the proportion of incumbent sampling to prior sampling a fixed constant in \algo{}(constant), \pp{} $= \eta \cdot$ \pinc{}.
\algo{}(decay)\footnote{this \algo{}(decay) is different from the decay version in Figure~\ref{fig:abl-prior} in Appendix~\ref{app:algo-abl-wp}} in Figure~\ref{fig:abl-inc} was designed such that the prior is decayed over time, similar to \pibo{}.
After every \sh{} bracket, the proportion of prior samples with respect to the incumbent proportion was halved.
As Figure~\ref{fig:abl-inc} highlights, the data-driven likelihood score-based method described in Section~\ref{sec:incumbent-sampling-weightage} is essential for robust performance across all cases.
This further supports our justification for the need for a non-naive method to achieve robustness, in Section~\ref{sec:beyond-naive}.

\begin{figure}
    \centering
    \begin{tabular}{c}
         \includegraphics[width=0.9\textwidth]{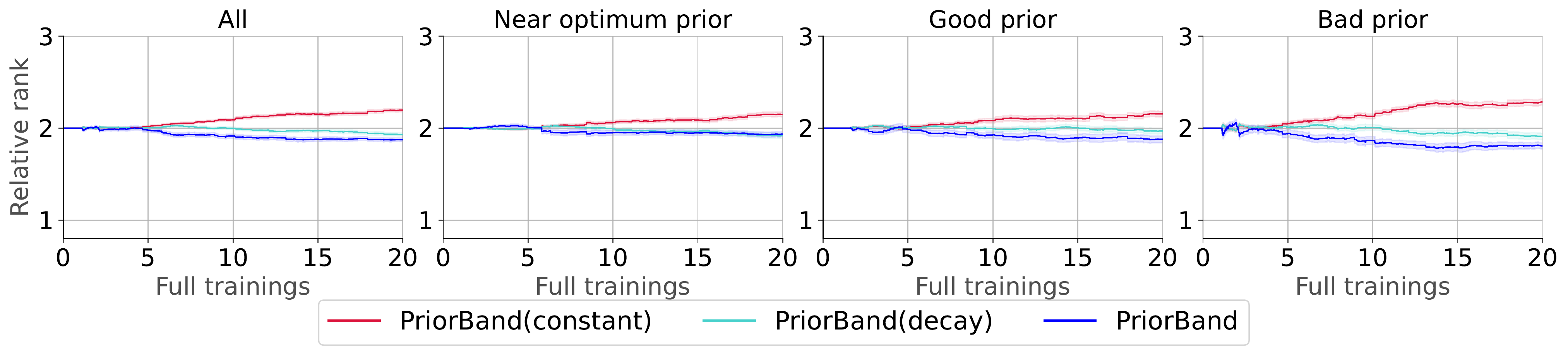} \\
         \includegraphics[width=0.9\textwidth]{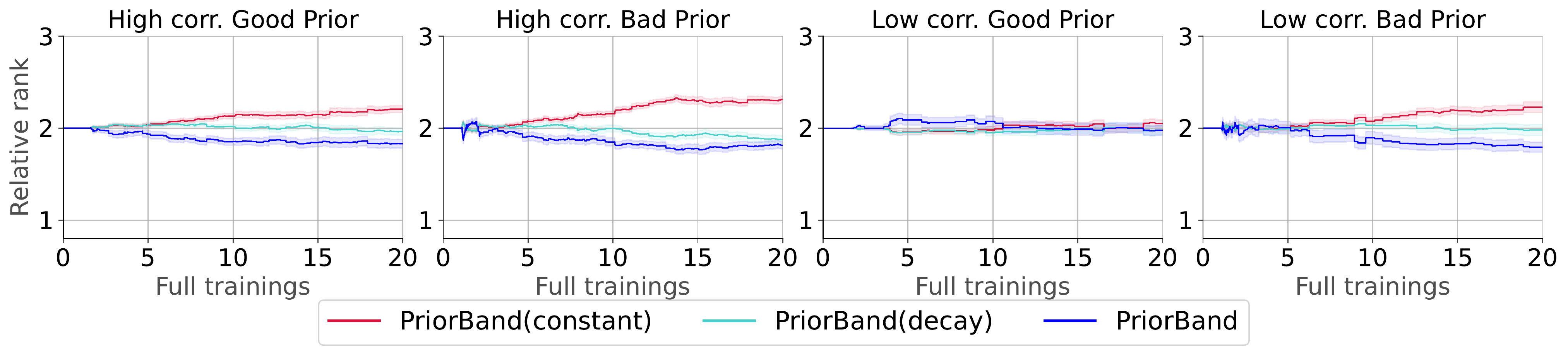}
    \end{tabular}
    \caption{A comparison of \algo{} with 3 different strategies for trading off between incumbent sampling and prior distribution sampling. \algo{}'s default of using likelihood scores for weighting shows a dominant performance in almost all scenarios. The one exception is in low correlation settings with a good prior where all versions seem to suffer under low correlations.
    However, under the same set of tasks, a bad prior induces a marked difference in performance. In general, the plot highlights that each variant reacts differently to bad priors.
    The adaptive nature of \algo{} clearly is robust to different scenarios comparatively.}
    \label{fig:abl-inc}
\end{figure}

\subsubsection{Choice of incumbent-based sampling} \label{app:algo-abl-hs}

\begin{figure}
    \centering
    \begin{tabular}{c}
         \includegraphics[width=0.9\textwidth]{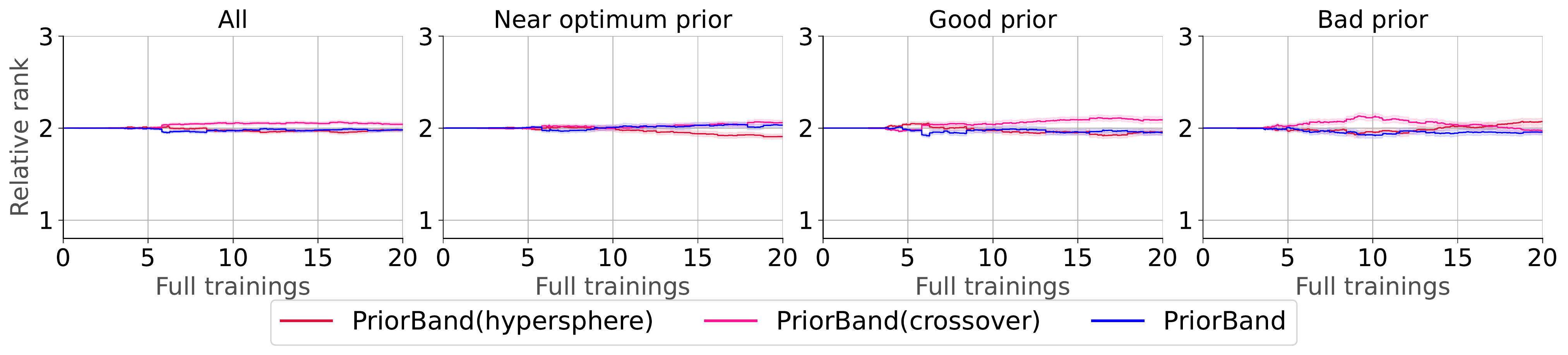} \\
         \includegraphics[width=0.9\textwidth]{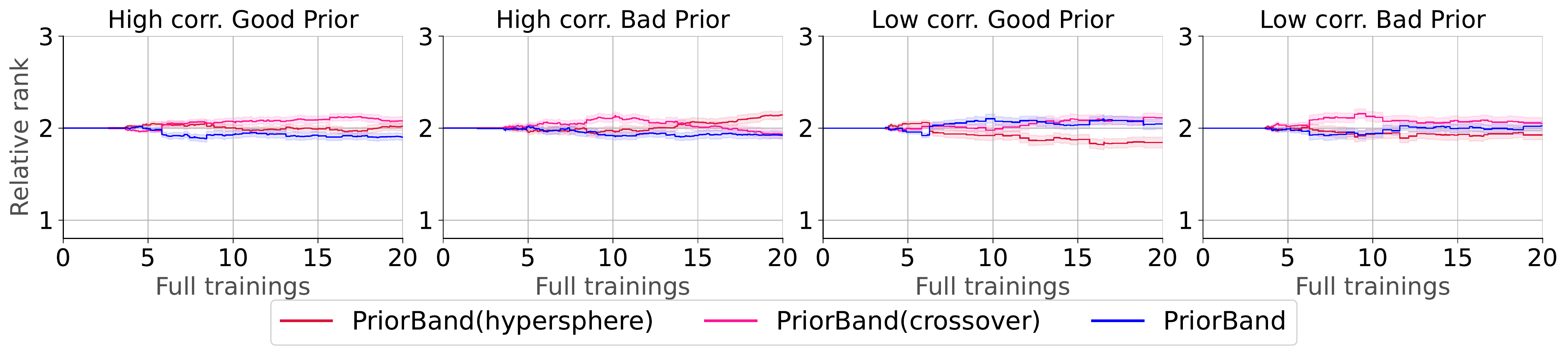}
    \end{tabular}
    \caption{A comparison of 3 different methods in \algo{} to sample based on the incumbent configuration. No single method dominates in terms of performance. The hypersphere sampling method shows the most variation across scenarios which is undesirable for robustness while uniform crossover requires a less intuitive and complex operation. \algo{} uses local perturbation which is more intuitive, a much simpler operation, and shows similar performance to the crossover variant.}
    \label{fig:abl-hs}
\end{figure}

In this section, we look at different methods of defining an incumbent-based sampler.
The role of the incumbent sampler is to allow exploitation by performing a local search around the best-seen configuration.
\algo{}(hypersphere) does so by sampling uniformly from a hypersphere around the incumbent, of a radius equivalent to the distance of the incumbent to its nearest neighbor.
This method is intuitive, however, in practice, numerical issues may arise if the incumbent and its nearest neighbor are close to each other.
This can make the sampling procedure extremely slow as the radius will keep shrinking, the more local search is performed.
Moreover, the distance measure for different hyperparameter types can be an extra design choice.
\algo{}(crossover) leverages the likelihood scores computed in Section~\ref{sec:incumbent-sampling-weightage} to choose if a random sample or a sample from the prior should participate in a simple uniform crossover with the incumbent.
Though this alleviates the computation issue of the \textit{hypersphere} method, it adds more complexity, is more difficult to interpret, and may not be as exploitative as desired of local search.
Our final choice described in Section~\ref{sec:priorband-mutation} is not only simple, fast, and easy to implement but also shows competitive performance with the other methods, as shown in Figure~\ref{fig:abl-hs}.

\begin{figure}
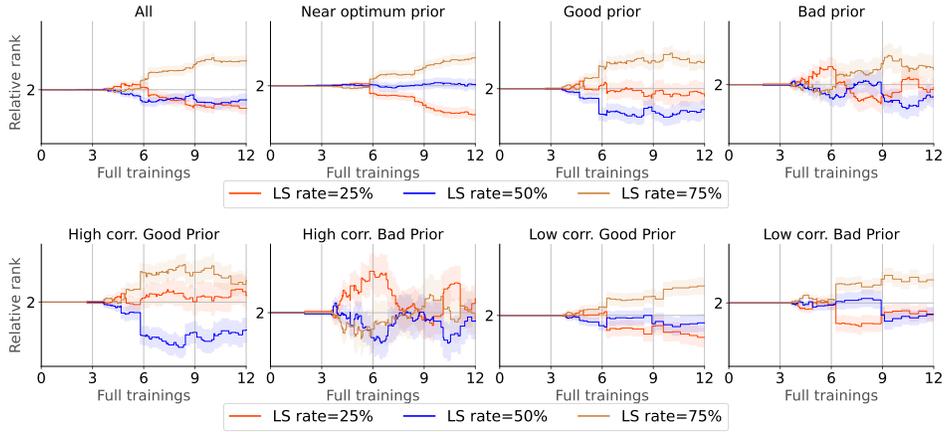

    \centering
    \begin{tabular}{c}
         \includegraphics[width=0.9\textwidth]{figs_rebuttal/abl-mut-rate-Relative-Ranking-by-Priors-1x4-max_fidelity_loss.pdf} \\
         \includegraphics[width=0.9\textwidth]{figs_rebuttal/abl-mut-rate-Relative-Ranking-by-Correlation-1x4-max_fidelity_loss.pdf}
    \end{tabular}
    \caption{
    An ablation study over the hyperparameter $p$ in the incumbent-based sampling in \algo{} which chooses the probability of selection of an HP to be perturbed for search.
    \algo{} chose $p=0.5$ which corresponds to $50\%$ in the figure.
    This setting appears to be never the worst method.
    A local search (LS) rate of $25\%$ is exploitative and can yield better performance under certain scenarios.
    However, our choice of $50\%$ was chosen to be generally balanced.
    A high rate of $75\%$ is too explorative and does not thus utilize the gains that the incumbent-based search can provide.}
    \label{fig:abl-mut-rate}
\end{figure}

\begin{figure}
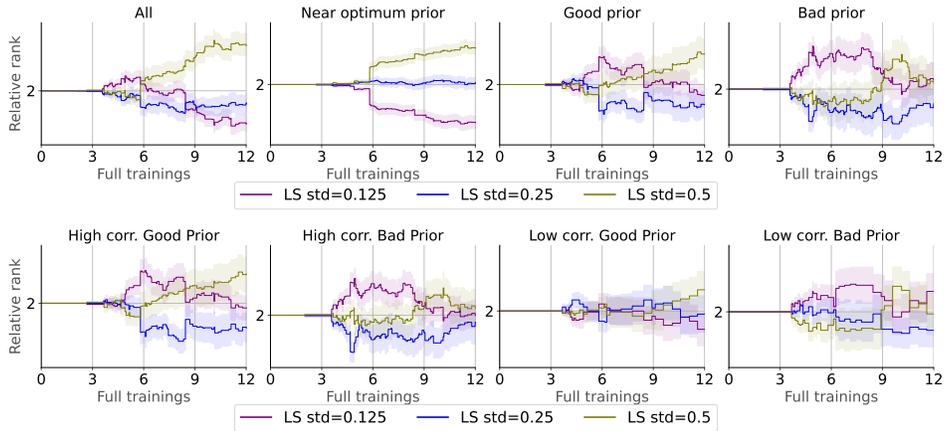

    \centering
    \begin{tabular}{c}
         \includegraphics[width=0.9\textwidth]{figs_rebuttal/abl-mut-std-Relative-Ranking-by-Priors-1x4-max_fidelity_loss.pdf} \\
         \includegraphics[width=0.9\textwidth]{figs_rebuttal/abl-mut-std-Relative-Ranking-by-Correlation-1x4-max_fidelity_loss.pdf}
    \end{tabular}
    \caption{
    An ablation study over the hyperparameter $\sigma$ in the incumbent-based sampling in \algo{} which chooses the standard deviation of the Gaussian that will be centered around the HP to be perturbed for local search (LS).
    \algo{} chose $\sigma=0.25$.
    This setting appears to be never the worst method.
    As expected, a peaker distribution under $\sigma=0.125$ leads to more exploitation and thus strong performance under near-optimum priors. 
    However, the quality of an incumbent improves over time. 
    Being too exploitative, too early, could hurt optimization which is seen under the varying quality of good priors.
    The broad distribution under $\sigma=0.5$ is much more explorative and is thus relatively worst under good priors and competitive under bad priors.
    }
    \label{fig:abl-mut-std}
\end{figure}

We additionally ablate over the chosen local mutation operation hyperparameters in Figure~\ref{fig:abl-mut-rate} and Figure~\ref{fig:abl-mut-std}.

\subsubsection{Importance of incumbent-based sampling} \label{app:algo-abl-to-inc-or-not-to-inc}

\begin{figure}
    \centering
    \begin{tabular}{c}
         \includegraphics[width=0.9\textwidth]{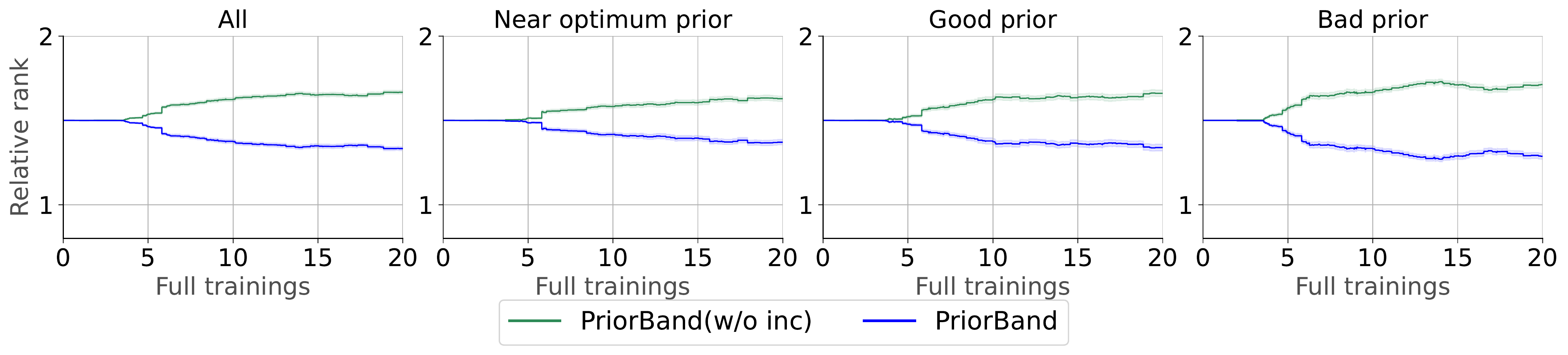} \\
         \includegraphics[width=0.9\textwidth]{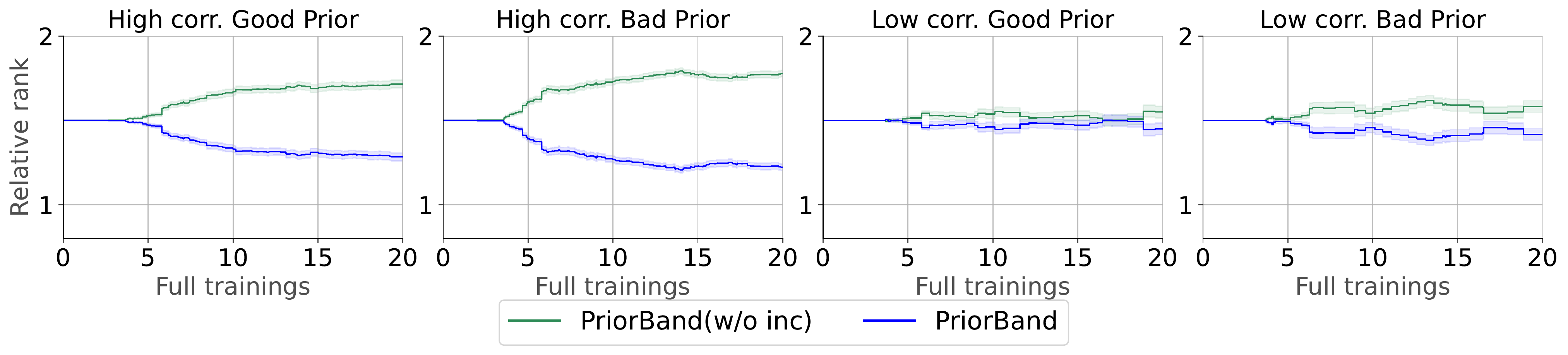}
    \end{tabular}
    \caption{The top row shows different qualities of priors with the leftmost subplot being their aggregation. The bottom row shows different combinations of low and high correlation across fidelities in both the good and bad prior setting. The plot evidently shows the strength of incumbent-based sampling in all presented scenarios. The top row demonstrates how incumbent-based sampling helps escape the prior, which has less effect in the near optimum setting but a much stronger effect in the bad prior setting, where incumbent-based sampling is essential. In the bottom row, we see that incumbent-based sampling is more important in high correlation settings, where incumbents in low fidelities are likely to carry to later fidelities. The low correlation setting contrasts this effect, where spending the budget on configurations near the incumbent at low fidelities is not likely to translate to high-performing configurations at the highest fidelity.}
    \label{fig:abl-to-inc-or-not-to-inc}
\end{figure}

In order to demonstrate the role of incumbent-based sampling in \algo{}'s design, Figure~\ref{fig:abl-to-inc-or-not-to-inc} shows a version of \algo{} where \pinc{} is set to $0$ with \pp{} inheriting all its probability (\algo{}(w/o inc)).
Thereby, switching off the incumbent sampling, while keeping all other designs intact. 
We see that for the chosen geometric trade-off between prior and random sampling (Section~\ref{sec:decay-random-sampling}, Appendix~\ref{app:algo-abl-wp}), the inclusion of incumbent-based local search is essential.

\subsubsection{Hyperparameters of incumbent-based sampling} \label{app:algo-mutation-hps}

For the incumbent-based sampling procedure described in Section~\ref{sec:priorband-mutation}, we employ a local perturbation around the incumbent using sampling from a Gaussian, $\mathcal{N}(\hat{\config}, \sigma^2)$.
Here, we assume a fixed standard deviation (with $\sigma=0.25$) for the distribution.
However, to make the perturbation more local, especially in higher dimensional spaces, we randomly choose if an \hp{} will be perturbed or not, with a fixed probability of $p=0.5$, an unbiased coin toss.

For the discrete hyperparameters, given $k$ categorical choices $C = \{c_1, c_2, \ldots, c_k\}$, we use a discrete distribution to perturb the category in the incumbent configuration. We give a weight of $k$ to the current incumbent's categorical choice $c_j$ and weight of $1$ to each other choice, giving us sampling probabilities of $p(c_j) = \frac{k}{2k - 1}$ and $p(c_i) = \frac{1}{2k - 1}, \forall c_i \in C, c_i \neq c_j$.
This sampling procedure is hyperparameter free.

The goal of \algo{} is to adapt to use good priors but recover from bad priors.
To establish this through different experiments on different task settings and scenarios, we
reduce confounding factors, keeping the aforementioned design for local search fixed.

Instead, hyperparameters could dynamically be adjusted to control \algo{}'s behavior, for example:
\begin{itemize}
    \item A schedule to reduce the standard deviation of the Gaussian to sample the perturbation noise.
    \item Dynamically adjusting the probability $p$ for \hp{} perturbation as optimization proceeds, potentially yielding better exploitation.
    \item Expert prior insight into HPO landscapes can allow more custom setting of the standard deviation for the Gaussian distribution.
    \item Expert knowledge of interaction effect or the importance of the hyperparameters in the search space could also allow for a tuned setting of $p$ for the selection of hyperparameters to perturb.
\end{itemize}
There are many possibilities for design, however, as our goal was to design the simplest approach that fulfills our desiderata, we choose a reasonable fixed setting and show it to be a robust choice in our experiments.

\subsubsection{Role of sampling the prior mode} \label{app:algo-abl-default}

\begin{figure} 
    \centering
    \begin{tabular}{c}
         \includegraphics[width=0.9\textwidth]{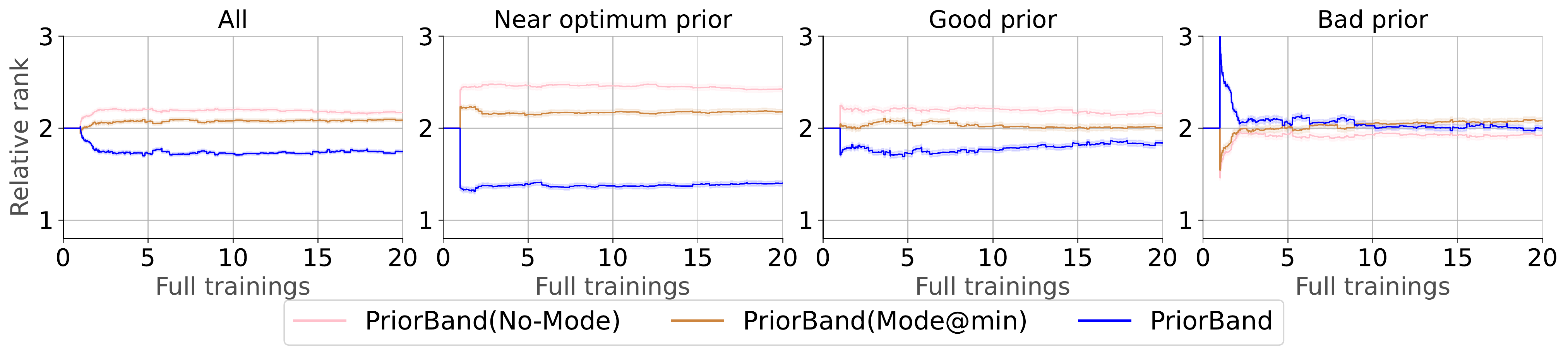} \\
         \includegraphics[width=0.9\textwidth]{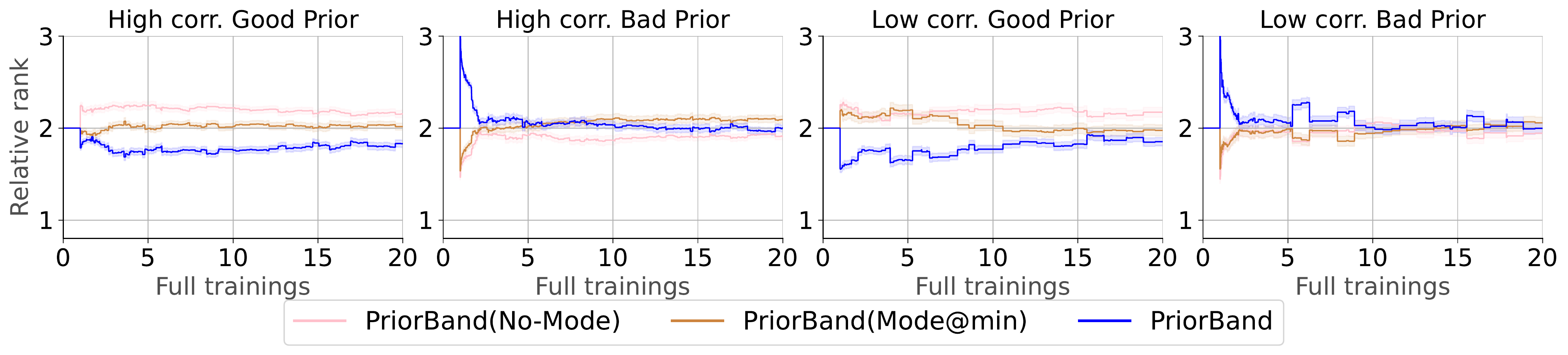}
    \end{tabular}
    \caption{We show 3 possible methods of how to take the first evaluation. \algo{} by default choosing to sample the prior's mode at the maximum fidelity, with (Mode@min) doing so at the minimum fidelity and (No-Mode) simply beginning with a random sample at the lowest fidelity. The most prominent failure case of \algo{} happens in the bad prior settings, as no exploration occurs until after 1 full training worth of budget is exhausted. 
    However, the algorithm still recovers rapidly. 
    }
    \label{fig:abl-default}
\end{figure}

Comparing \algo{}, where the prior mode is sampled at the maximum fidelity, with a version of \algo{} where the prior mode is sampled at the minimum fidelity (Mode@min), and not sampling the mode at all (No-Mode), in Figure~\ref{fig:abl-default}.
It appears that sampling the mode, in the beginning, provides huge gains especially if the prior is of good quality. 
In the case when the prior is not informative, \algo{} shows it can recover well even then.
Given that most practical settings have multiple workers available, the fact that \algo{} can recover rapidly from misleading priors, and ultimately the utility of a good expert prior, it is reasonable to simply evaluate the expert default as the first evaluation.
Under a multi-worker setup, the cost of this evaluation is amortized with the cheaper evaluations proceeding with the other workers. 
Optionally, the choice of whether the prior mode should be evaluated as the first evaluation or not can be toggled. 
For example, when the expert is confident of a good prior configuration and has knowledge of its performance from previous runs.

\subsection{Post-hoc view of \algo{} for interpretability} \label{app:algo-interpret}

\begin{figure} 
    \centering
    \includegraphics[width=0.9\textwidth]{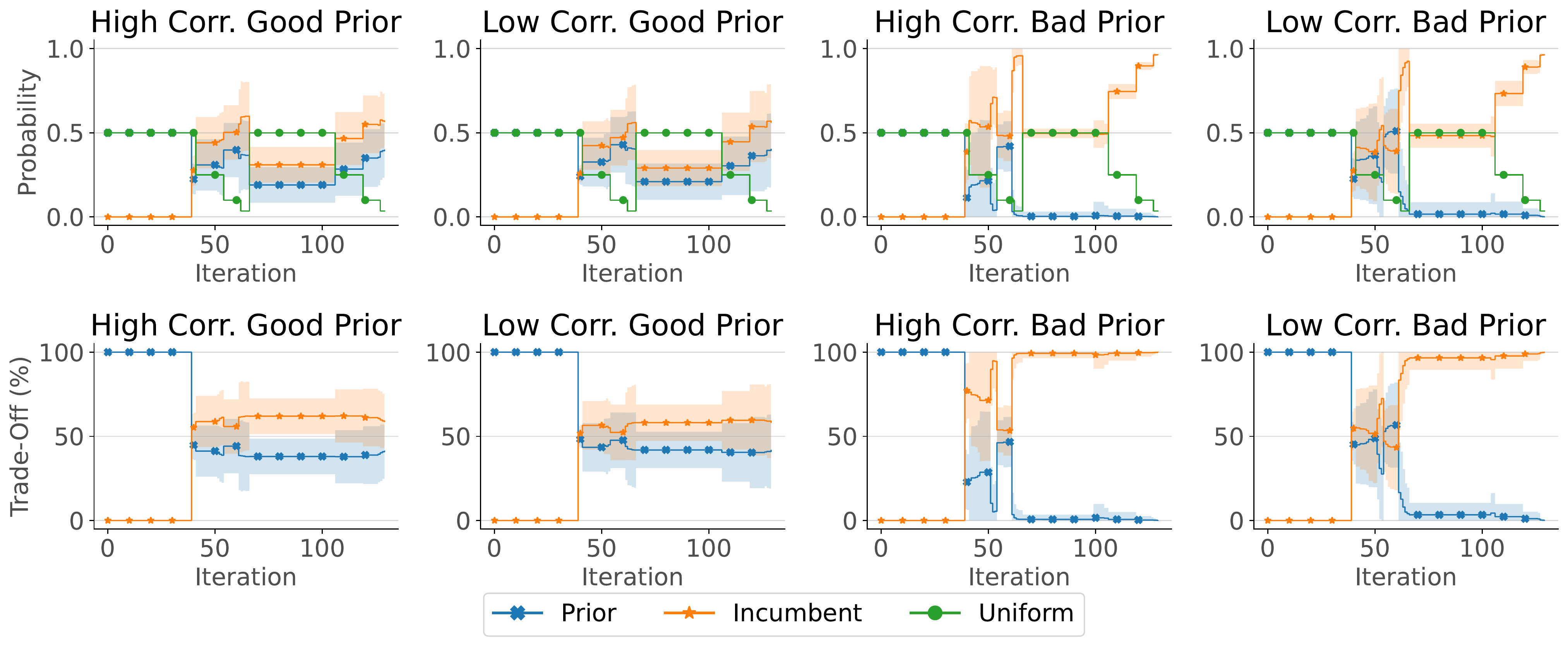}
    \caption{
        Evolution of the Ensemble Sampling policy \E{} and its probabilities \pu{}, \pp{} and \pinc{} for runs on the multi-fidelity Hartmann-3 benchmark, for all combinations of high/low fidelity correlation and good/bad prior strength. 
        The x-axis counts the number of multi-fidelity evaluations made by \hb{}.
        [\textbf{Top}] The y-axis shows the exact probability assigned to each sampling strategy, \pu{}, \pp{}, and \pinc{}. 
        Uniform sampling follows a repeating pattern as a function of fidelity. 
        Incumbent sampling remains inactive at the beginning, but once activated, it dynamically adjusts depending on the quality of the prior given.
        [\textbf{Bottom}] We visualize the dynamic trade-off between prior and incumbent-based sampling, showing \pinc{} and \pp{} as a percentage of \pinc{} $+$ \pp{} on the y-axis.
        Under good priors, the probability of sampling from around the incumbent and sampling from priors is almost similar.
        Whereas under bad priors, the prior is discarded almost completely after one complete round of \hb{}.
        Differences across high-low correlation setups exist, in the period between activation of the incumbent and one complete iteration of \hb{}, but values stabilize after this iteration for all cases.
    }
    \label{fig:e-weights}
\end{figure}

A \dl{} expert can often provide a prior \prior{} but is often unsure as to its merits. 
By tracking the evolution of \algo{}'s sampling probabilities during the optimization process, we can get an indicator of the strength of the \prior{}. 
The more useful the prior is for performance, the higher the probability of sampling from \prior{} and thus a higher \pp{}.
We illustrate this in Figure \ref{fig:e-weights}, where we plot the sampling probabilities \pu{}, \pp{} and \pinc{} of \algo{} during an \hpo{} run with $50$ seeds on the synthetic multi-fidelity Hartmann-3 function (Appendix \ref{app:exp-benchmarks-mfh}). 
We clearly see that our motivation for the design choice of \algo{} is well supported as the probability of sampling from the incumbent is not suppressing sampling from prior under good priors, but are aggressively affecting the chance of sampling from bad priors.
Such a post-hoc analysis offers a \dl{} expert insights into their own prior knowledge, allowing them to update or re-enforce their beliefs about what a good prior is for the problem at hand.




\subsection{Model extensions} \label{app:algo-model}

In this section, we elucidate the model-based components when extending algorithms with \E{} as shown in Figure~\ref{fig:exp-model}.
Firstly, we briefly explain BOHB and its modeling choice.
BOHB subsumes the hyperparameter of the initial design size by setting it to $N_{dim} + 2$ where $N_{dim}$ is the dimensionality of the search space of a task. 
To activate model-based search, BOHB uses the following criteria: find the highest fidelity level with at least $N_{dim} + 2$ evaluated observations.
If no such fidelity level exists, continue with uniform random sampling.
If such fidelity exists, use all the observations at that fidelity to build a TPE as the surrogate. 
Since a model is built at a fidelity level, the fidelity variable is not part of the feature set for the surrogate.
During acquisition, EI is used to obtain a configuration that approximately maximizes the TPE surrogate.
\begin{equation} \label{eq:ei}
    \text{EI}_{z}(\bm{x} | \mathcal{D}) = \mathbb{E}[\text{max}\{ f_{z}(\bm{x}) - y_{z}^{min}, 0 \}],
\end{equation}
where $y_{z}^{min}$ is the best score seen at the fidelity $z$.
Given $f$ is a model built at a fidelity level $z$, the EI acquisition estimates the improvement of the suggested configuration at fidelity $z$. 
When the number of observations at $z_\text{max}$ is $N_{dim} + 2$, the EI acquisition performs similarly to vanilla-BO.
Mobster or asynchronous-BOHB follows the exact BO loop as BOHB, except that it uses a GP instead of a TPE and uses asynchronous \hb{} for scheduling and not vanilla-\hb{}.

\paragraph{Model extension to \algo{}.} In our experiments to extend \algo{} with a model in \algo{}+BO, the above procedure of automatically switching to model-search lead to \E{} not taking action and affecting optimization as we desire.

Similar to the initial design size in BO, incumbent-based sampling has an `activation criteria' of one full \sh{} bracket being evaluated, and at least one configuration evaluated at $z_\text{max}$, after which incumbent-based search begins.
We could default to BOHB's `activation criteria' but for certain problems, the number of observations would satisfy the  $N_{dim} + 2$ criteria even before the $1^\text{st}$ \sh{} bracket is over.
This implies that incumbent-based sampling, a crucial component of \E{}, is never activated. 
Hence, we follow an approach that is more in line with BO and \pibo{} in which $10$ function evaluations are used as the initial design budget before switching to model search. This is most evident in Figure~\ref{fig:exp-model}, where \pibo{} behaves identically to only performing Random Search on the prior distribution, diverging at $10$ full function evaluations. 

We treat $10\times$ as ($10 \cdot z_\text{max}$) the total budget (in epochs) exhausted during multi-fidelity optimization.
After which, a GP model is activated for search that models \textit{all} the observations made during the optimization, across any fidelity available.
That is, the fidelity is an extra dimension modeled along with the search space.
During the acquisition, since it is known from the optimization state \state{} which fidelity $z$ the new sample will be evaluated at, a 2-step optimization is performed when maximizing EI.
In the first step, a set of configurations ($10$ in our experiments) is extracted for fidelity $z$ through Monte Carlo estimates of Equation~\ref{eq:ei} returning configurations likely to improve over the best configuration found so far at $z$. 
Following this, the EI score is calculated for this selected set of configurations using Equation~\ref{eq:ei} but with $z=z_\text{max}$. 
At this stage, $y^{min}$ is chosen to be the best score obtained across all observations.
The idea is to choose a configuration that is likely to maximize performance at the fidelity level where it is being queried and is also likely to improve at the target fidelity $z_\text{max}$.

\begin{figure}
    \centering
    \begin{tabular}{c}
         \includegraphics[width=0.9\textwidth]{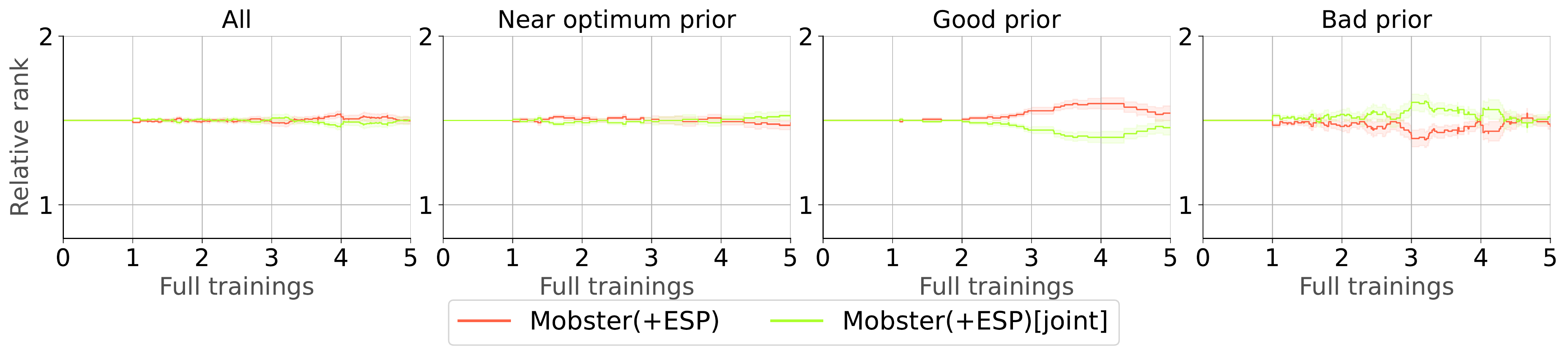} \\
         \includegraphics[width=0.9\textwidth]{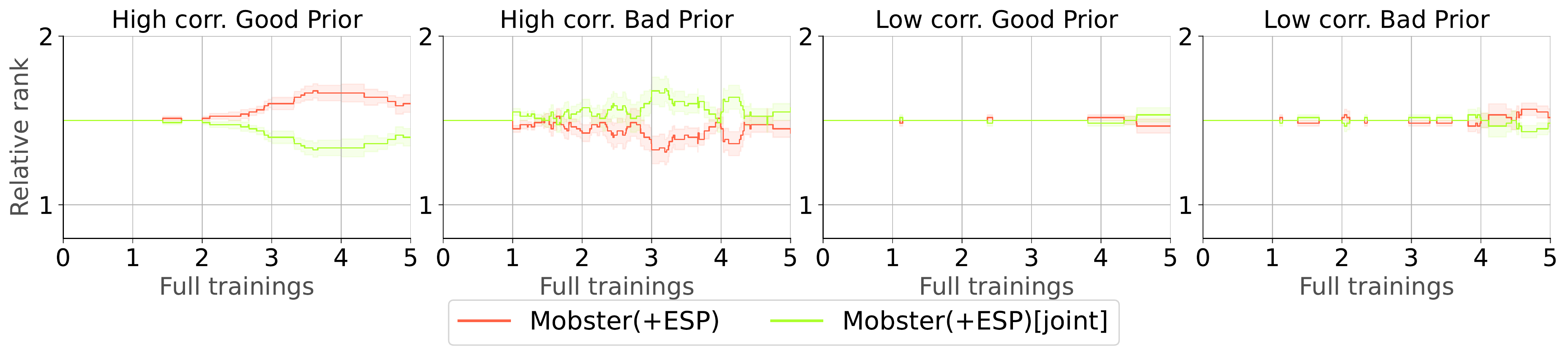}
    \end{tabular}
    \caption{Comparing multi-fidelity modelling on asynchronous-\hb{} per \textit{rung} and \textit{jointly} over the search space and fidelity ([j]). (Top) Compares the performances over different qualities of priors. (Bottom) Compares the good (at25) and bad prior cases, under benchmarks grouped into good-bad correlation. All algorithms were run for a total of $20 \times$ over $4$ workers.}
    \label{fig:async-bo-compare}
\end{figure}

\paragraph{Contextualizing model-search under \E{}.}
During the typical initial design phase of \algo{}+BO, it is \algo{} that runs with the defined \E{} comprising of \uniform{}, \prior{}, \inc{} as actions.
At the end of the initial design phase of \algo{}+BO, the action set \actionset{} updates from \{ \prior{}, \uniform{}, \inc{} \} to \{ \prior{}, \uniform{}, \inc{}, \model{} \} with new weights as \pp{} = \pu{} = \pinc{} = 0 and $p_\mathcal{M}$ = 1, where $\mathcal{M}$ denotes model-based search. 

\paragraph{Comparing multi-fidelity modeling.} In the previous section we motivate criterion and acquisition that allows \E{} to influence search and the initial design space. 
For \algo{}, we note that the incumbent activation criteria are not fulfilled if modeling per fidelity with dimensionality as a criterion, as is done in BOHB and Mobster.
In contrast, asynchronous-\hb{} samples at random fidelities instead of the lowest fidelity first.
When used with the ESP this allows the possibility of incumbent-based samples being activated before the model search begins.
Hence, for asynchronous \hb{} we can apply per-fidelity modeling. 
In Figure~\ref{fig:async-bo-compare}, we thus compare the $2$ types of modeling discussed above: per-fidelity (Mobster+E) and joint modeling like \algo{}+BO (Mobster+E[joint]).
We conclude that there is no significant difference in  the performance of the two modeling choices.
Though the joint model seems to perform slightly better under good priors.


\section{More on experiments} \label{app:exp-res}

This section gives a detailed experiment analysis that expounds on the results from Section~\ref{sec:exp-pb-robust}-\ref{sec:exp-model}.
We compare similar setups but include our constructed prior-based baselines, over a different grouping of benchmarks with high and low correlations (Appendix~\ref{app:exp-res-pb},~\ref{app:exp-res-general},~\ref{app:exp-res-model}).
In Appendix~\ref{app:exp-res-priors} we compare the prior-based algorithms with each other.
We further show the final validation performance tables of algorithms for the different experiments for every benchmark-prior combination, over $2$ budget horizons in Appendix~\ref{app:exp-res-tables}.

\subsection{Robustness of \algo{}} \label{app:exp-res-pb}

\begin{figure*}
    \centering
    \begin{tabular}{c}
         \includegraphics[width=0.95\textwidth]{figs_post_rebuttal/exp-hb-At25-Prior-Incumbent-Traces-max_fidelity_loss.pdf} \\
         \includegraphics[width=0.95\textwidth]{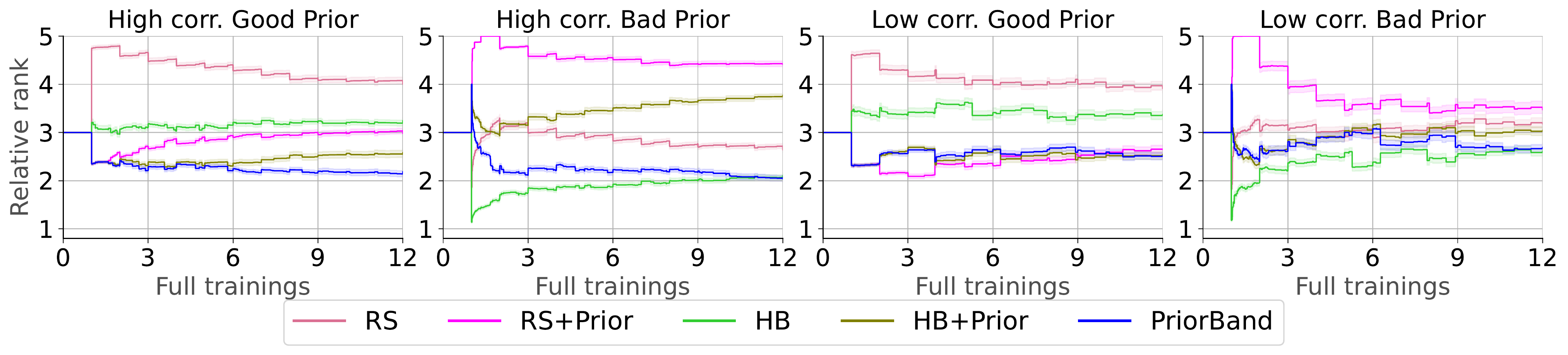} \\
         \includegraphics[width=0.95\textwidth]{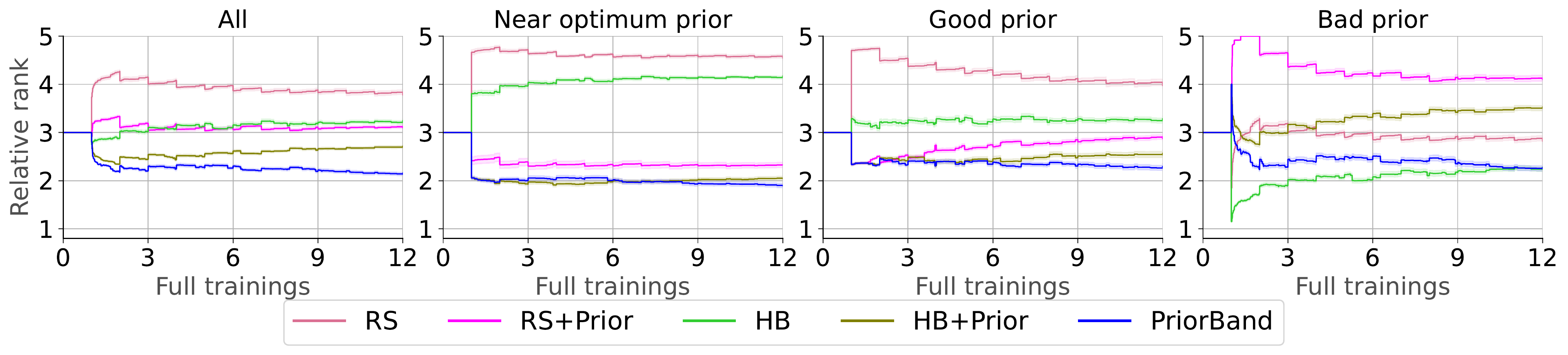} \\
    \end{tabular}
    \caption{[Top] Each plot compares the algorithms based on the average normalized regret across $50$ seeds, under the good prior setting.
    The optima for a benchmark was assumed to be the best scores achieved by all algorithms across all seeds.
    \algo{} is anytime best across in all cases.
    [+] denotes the benchmark tends to have a strong correlation across fidelities, [-] denotes a weak or poor correlation across fidelities;
    [Bottom] Comparing PriorBand with other baselines on single-worker runs for $50$ seeds.
    The \textit{top} row compares the average relative ranks across all benchmarks under different prior qualities, where the \textit{All} averages over the $3$ prior strengths too.
    The \textit{bottom} row groups the benchmarks into high-low correlation sets based on fidelity correlations and creates $4$ scenarios when combined with good-bad priors.
    \algo{} emerges as the most consistent performer across all $7$ scenarios.
    Every other baseline has at least one setting where it is one of the $2$ worst algorithms.
    Prior-based methods show improved performance with a good prior but suffer with a bad prior, as expected. 
    Likewise, multi-fidelity methods benefit from high correlation. 
    Under low-correlation settings, the ranking gap between RS and \hb{} expectedly is lower.}
    \label{fig:exp-pb-corr}
\end{figure*}

More supporting results for Figure~\ref{fig:exp-pb} in Section~\ref{sec:exp-pb-robust}.
we show the same results but with the addition of \textit{near optimum} priors.
Figure~\ref{fig:exp-pb-corr} shows the relative rank comparison over near optimum, good, and bad priors, as well as the aggregate (\textit{All}).
Most notably, we see the relation between \algo{} and \hb{} with respect to prior strength, showcasing substantial benefits of strong priors while recovering in the bad prior setting.

To further illustrate how different tasks affect optimizer behavior, we cluster the set of $12$ benchmarks we've chosen into $8$ good correlation benchmarks and $4$ bad correlation benchmarks as defined in Appendix~\ref{app:cluster-benchmarks}. 
Figure~\ref{fig:exp-pb-corr} shows the relative ranks for the same set of algorithms when grouped and aggregated along these criteria.
\algo{} is the most robust algorithm shown here.

\subsection{Generality of \E{}} \label{app:exp-res-general}

\begin{figure*}
    \centering
    \begin{tabular}{c}
         \includegraphics[width=0.95\textwidth]{figs_post_rebuttal/exp-async-hb-At25-Prior-Incumbent-Traces-max_fidelity_loss.pdf} \\
         \includegraphics[width=0.95\textwidth]{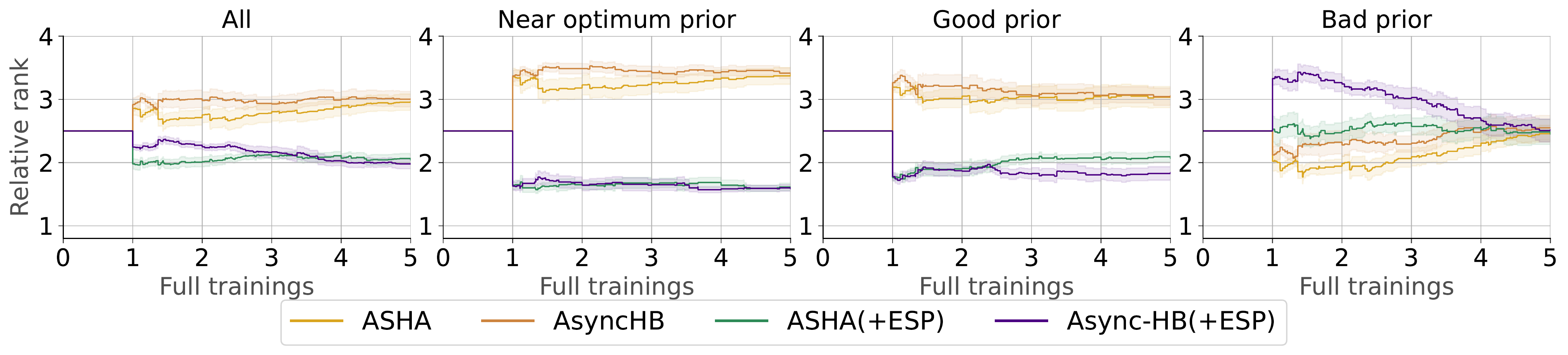} \\
         \includegraphics[width=0.95\textwidth]{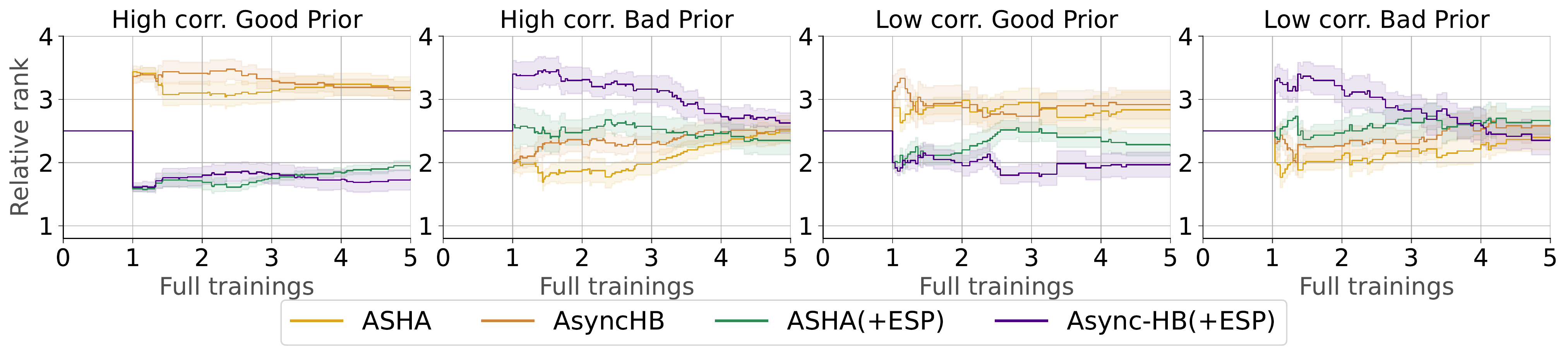} \\  
    \end{tabular}
    \caption{[Top] Each plot compares the algorithms based on the average normalized regret across $50$ seeds, under the good prior setting, run over $4$ workers each.
    The optima for a benchmark was assumed to be the best scores achieved by all algorithms across all seeds.
    [+] denotes the benchmark tends to have a strong correlation across fidelities, [-] denotes a weak or poor correlation across fidelities.
    Refer to Appendix~\ref{app:failing-benchmarks} for missing benchmark;
    [Bottom] Comparing ASHA, AsynchronousHB with their ESP augmented versions (+E) and \algo{} when distributed over $4$ workers for a total budget of $20\times$.
    The \textit{top} row compares the average relative ranks across all benchmarks under different prior qualities, where the \textit{All} averages over the $3$ prior strengths.
    The \textit{bottom} row groups the benchmarks into high-low correlation sets based on fidelity correlations and creates $4$ scenarios when combined with good-bad priors.
    All $3$ algorithms with the ESP \E{} show superior performance under good priors and strong recovery under bad priors.
    Under a bad prior for good correlation benchmarks, the recovery of ESP-based asynchronous methods is slower since the vanilla algorithms perform better under good correlation.
    Unlike the low correlation set under bad priors, where all \E{}-based algorithms show faster recovery.
    Overall, \algo{} remains a strong performer even in the parallel setup.}
    \label{fig:exp-gen-strong}
\end{figure*}

Figure~\ref{fig:exp-gen-strong} (top) compares $3$ different prior qualities for more supporting results for Figure~\ref{fig:exp-generality} in Section~\ref{sec:exp-generality}.
Figure~\ref{fig:exp-gen-strong} (bottom) splits the set of $12$ benchmarks into high-low correlations and plots their interaction with good-bad priors.

\subsection{Extensibility of \E{} with models} \label{app:exp-res-model}

\begin{figure*}
    \centering
    \begin{tabular}{c}
         \includegraphics[width=0.95\textwidth]{figs_post_rebuttal/exp-bo-At25-Prior-Incumbent-Traces-max_fidelity_loss.pdf} \\
         \includegraphics[width=0.95\textwidth]{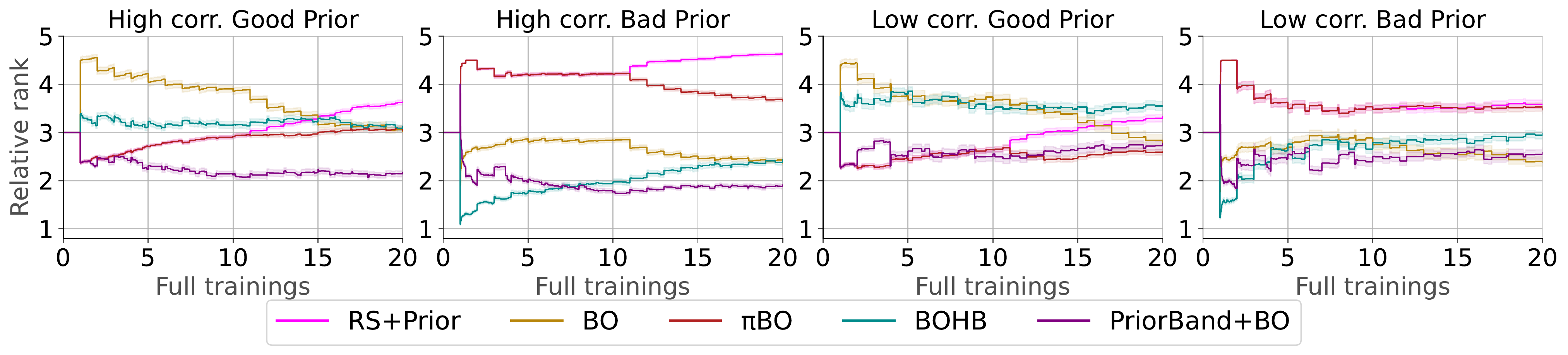} \\
         \includegraphics[width=0.95\textwidth]{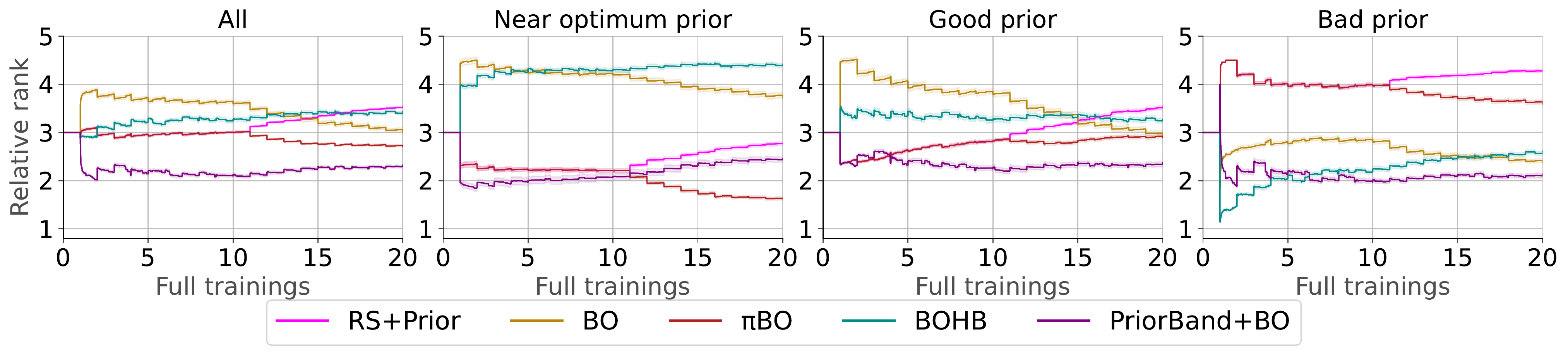} \\
    \end{tabular}
    \caption{
    [Top] Each plot compares the algorithms based on the average normalized regret across $50$ seeds, under the good prior setting.
    The optima for a benchmark was assumed to be the best scores achieved by all algorithms across all seeds.
    \algo{}+\bo{} is anytime best across in all cases.
    [+] denotes the benchmark tends to have a strong correlation across fidelities, [-] denotes a weak or poor correlation across fidelities;
    [Bottom] Comparing model extensions over $3$ sets of prior qualities.
    Using PriorBand with BO shows a dominant performance in almost all cases. In the first row, we show that $\pi$BO does outperform all other methods in the near optimum setting, once model sampling activates. This is likely due to $\pi$BO's emphasis on the prior which is the same cause for its poor performance in the bad prior setting. Both BO and BOHB suffer from having no access to the prior in the near optimum/good prior setting while only BOHB really has an advantage over PriorBand+BO early on in the bad prior setting. In the second row, we see that extending Mobster with the ensembling policy \E{} imbues Mobster with an effective mechanism for exploiting priors. However, this exploitation has a more consistent negative impact when the prior is bad, at least when compared to vanilla Mobster and our model-based PriorBand+BO.}
    \label{fig:exp-model-strong}
\end{figure*}

\begin{figure*}
    \centering
    \begin{tabular}{c}
         \includegraphics[width=0.95\textwidth]{figs_post_rebuttal/async-pi-bo-At25-Prior-Incumbent-Traces-max_fidelity_loss.pdf} \\
         \includegraphics[width=0.95\textwidth]{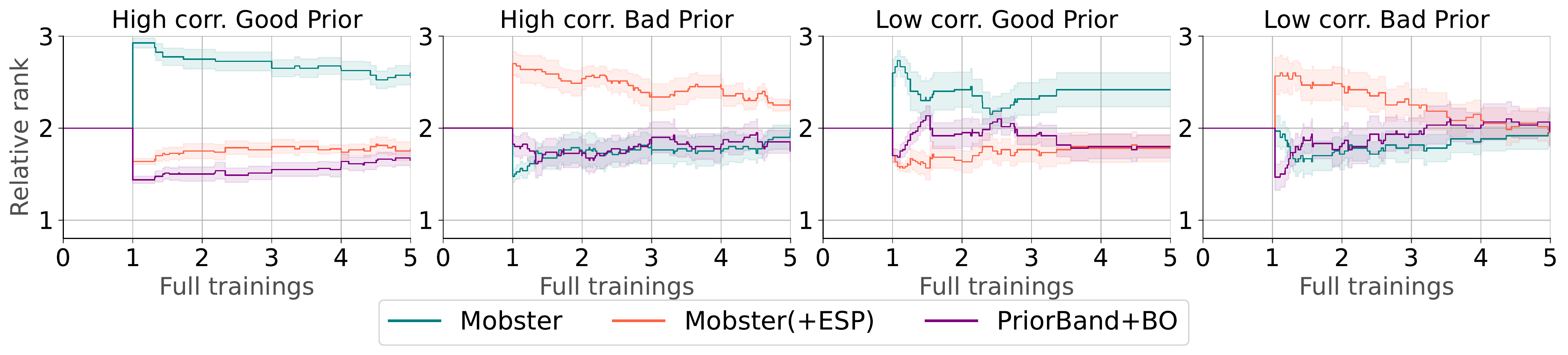} \\
         \includegraphics[width=0.95\textwidth]{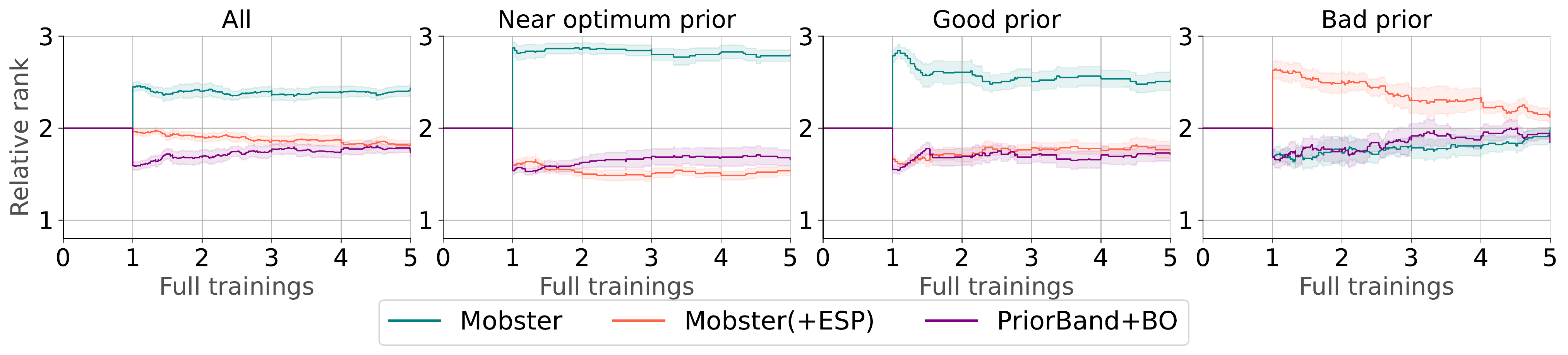} \\
    \end{tabular}
    \caption{
    [Top] Each plot compares the algorithms based on the average normalized regret across $50$ seeds, under the good prior setting, run over $4$ workers each.
    The optima for a benchmark was assumed to be the best scores achieved by all algorithms across all seeds.
    [+] denotes the benchmark tends to have a strong correlation across fidelities, [-] denotes a weak or poor correlation across fidelities.
    Refer to Appendix~\ref{app:failing-benchmarks} for missing benchmark; 
    [Bottom] Comparing model extensions over $3$ sets of prior qualities.
    Using PriorBand with BO shows a dominant performance in almost all cases. In the first row, we show that $\pi$BO does outperform all other methods in the near optimum setting, once model sampling activates. This is likely due to $\pi$BO's emphasis on the prior which is the same cause for its poor performance in the bad prior setting. Both BO and BOHB suffer from having no access to the prior in the near optimum/good prior setting while only BOHB really has an advantage over PriorBand+BO early on in the bad prior setting. In the second row, we see that extending Mobster with the ensembling policy \E{} imbues Mobster with an effective mechanism for exploiting priors. However, this exploitation has a more consistent negative impact when the prior is bad, at least when compared to vanilla Mobster and our model-based PriorBand+BO.}
    \label{fig:exp-model-async}
\end{figure*}

More supporting results for Figure~\ref{fig:exp-model} in Section~\ref{sec:exp-model}.
In Figure~\ref{fig:exp-model-strong}, which compares $3$ different prior qualities, we assess how the quality of the prior effects model-based methods.
We also compare the model-based extension of \algo{} (\algo{}+BO) against Mobster (asynchronous BOHB) in the distributed setting, in Figure~\ref{fig:exp-model-async}.

\subsection{Comparing prior-based baselines} \label{app:exp-res-priors}
\begin{figure*}
    \centering
    \begin{tabular}{c}
         \includegraphics[width=0.95\textwidth]{figs_post_rebuttal/pi-bo-At25-Prior-Incumbent-Traces-max_fidelity_loss.pdf} \\
         \includegraphics[width=0.95\textwidth]{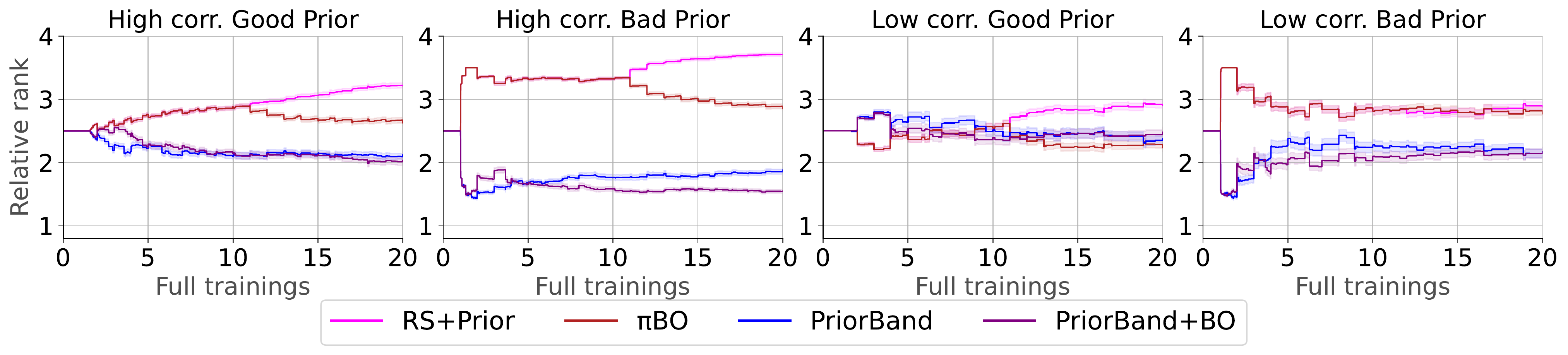} \\
         \includegraphics[width=0.95\textwidth]{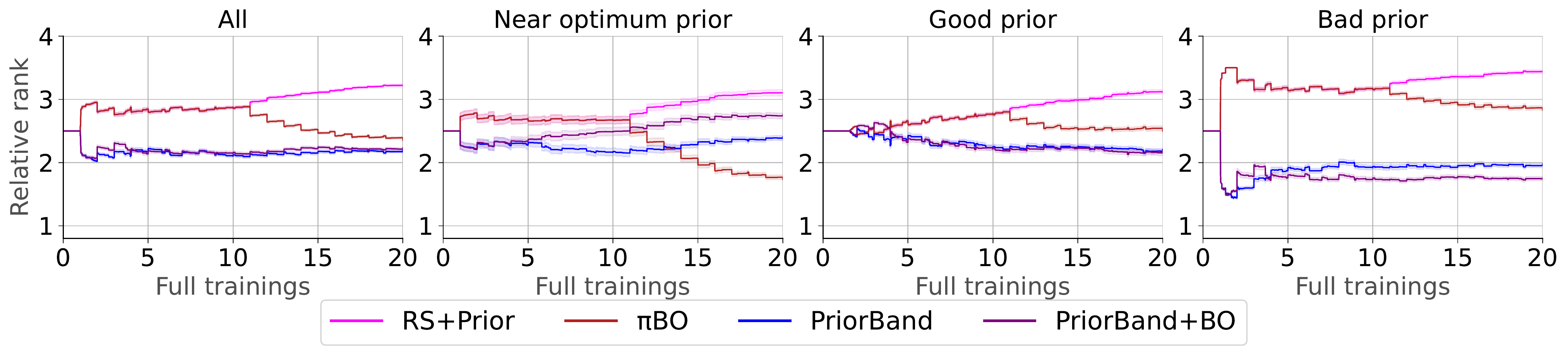} \\
    \end{tabular}
    \caption{[Top] Each plot compares the algorithms based on the average normalized regret across $50$ seeds, under the good prior setting.
    The optima for a benchmark was assumed to be the best scores achieved by all algorithms across all seeds.
    \algo{}+\bo{} offers its model benefits over \algo{} under longer budgets but has mixed performance under poor correlation of fidelity performance. 
    [+] denotes the benchmark tends to have a strong correlation across fidelities, [-] denotes a weak or poor correlation across fidelities;
    [Bottom] Comparing model-based methods which incorporate the prior. In the near optimum case, $\pi$BO's strong performance only occurs once its model-based sampling occurs. This strong emphasis on searching around the prior is also what causes $\pi$BO's weaker performance in the bad prior case. When comparing PriorBand against its model counterpart, PriorBand+BO, we see that PriorBand utilizes the near optimum prior better while PriorBand+BO with its model-based sampling leads to better performance in the long run if the prior is bad. This trade-off is averaged off in the case of a good prior which is neither overly optimistic as the near optimum prior is, nor neither as pessimistic as the bad prior is.}
    \label{fig:exp-all-priors}
\end{figure*}

We would like to determine if \algo{} is sufficient to outperform existing prior-based baselines as well as our own extension of \algo{} which is model-based, namely PriorBand+BO. Figure~\ref{fig:exp-all-priors} show the results of this comparison.  

\subsection{Final performance tables} \label{app:exp-res-tables}

In this section, Tables~\ref{table:mssg-hb-table-max_fidelity_loss},~\ref{table:mssg-async-table-max_fidelity_loss},~\ref{table:exp-async-hb-table-max_fidelity_loss},~\ref{table:table-bo-table-max_fidelity_loss},~\ref{table:async-pi-bo-table-max_fidelity_loss}, compare the performance of \algo{}, its BO extension and other applications of \E{}, to commonly used algorithms from the literature for the different classes of optimizers. The tables show the final performance at $2$ budget horizons, highlighting the robustness of the \esp{}-based algorithms.

\begin{table}
\caption{\protect Table comparing Random Search (RS), HyperBand (HB), and \algo{}'s final validation errors of the current incumbent at the highest fidelity at $2$ budget horizons of $5\times$ and $12\times$. 
Runs are averaged over $50$ seeds on $1$ worker each.
\algo{} shows superior anytime performance under informative priors.
Under extremely bad priors, for shorter compute budgets ($5\times$) \algo{} has a poor start, however, given an adequate budget ($12\times$) \algo{} can match \hb{}'s performance on average.}
\label{table:mssg-hb-table-max_fidelity_loss}
\begin{center}
\scalebox{0.55}{
\centering\begin{tabular}{l ccc c ccc}
\toprule

\textbf{Benchmark}  & \multicolumn{3}{c}{5x} & & \multicolumn{3}{c}{12x} \\ \cline{2-4} \cline{6-8}
\\
 & \multicolumn{7}{c}{\textbf{Near Optimum Prior}} \\
\\
 &      RS &                     HB &             PriorBand &        &        RS &                    HB &              PriorBand \\ \cline{2-4} \cline{6-8}

\textbf{JAHS-C10 }     &      $11.454\pm1.935$ &       $11.129\pm1.256$ &  $\bm{8.252\pm0.215}$ & & $10.451\pm1.055$ &      $10.077\pm0.769$ &   $\bm{8.236\pm0.197}$ \\
\textbf{JAHS-CH }      &       $6.752\pm0.943$ &         $6.332\pm1.12$ &  $\bm{4.449\pm0.124}$ & &  $5.977\pm0.646$ &       $5.704\pm0.659$ &   $\bm{4.445\pm0.127}$ \\
\textbf{JAHS-FM }      &       $5.497\pm0.396$ &        $5.846\pm0.823$ &   $\bm{4.724\pm0.06}$ & &  $5.282\pm0.299$ &       $5.306\pm0.267$ &   $\bm{4.719\pm0.055}$ \\
\textbf{LC-126026 }    &       $0.053\pm0.011$ &         $0.046\pm0.01$ &  $\bm{0.024\pm0.008}$ & &  $0.048\pm0.008$ &       $0.044\pm0.008$ &   $\bm{0.023\pm0.006}$ \\
\textbf{LC-167190 }    &       $0.214\pm0.021$ &        $0.193\pm0.025$ &  $\bm{0.136\pm0.017}$ & &     $0.2\pm0.02$ &       $0.187\pm0.023$ &   $\bm{0.133\pm0.013}$ \\
\textbf{LC-168330 }    &       $0.444\pm0.046$ &         $0.41\pm0.035$ &   $\bm{0.29\pm0.041}$ & &  $0.406\pm0.038$ &       $0.397\pm0.032$ &   $\bm{0.275\pm0.028}$ \\
\textbf{LC-168910 }    &        $0.38\pm0.093$ &        $0.314\pm0.021$ &    $\bm{0.2\pm0.021}$ & &  $0.324\pm0.032$ &       $0.308\pm0.018$ &   $\bm{0.194\pm0.019}$ \\
\textbf{LC-189906 }    &        $0.222\pm0.09$ &        $0.151\pm0.021$ &  $\bm{0.113\pm0.015}$ & &  $0.166\pm0.033$ &       $0.145\pm0.019$ &   $\bm{0.107\pm0.011}$ \\
\textbf{PD1-Cifar100 } &       $0.378\pm0.156$ &        $0.338\pm0.091$ &  $\bm{0.257\pm0.051}$ & &  $0.302\pm0.074$ &        $0.29\pm0.039$ &    $\bm{0.238\pm0.04}$ \\
\textbf{PD1-ImageNet } &       $0.333\pm0.108$ &        $0.306\pm0.041$ &  $\bm{0.239\pm0.038}$ & &  $0.282\pm0.049$ &       $0.268\pm0.022$ &   $\bm{0.228\pm0.026}$ \\
\textbf{PD1-LM1B }     &       $0.681\pm0.034$ &        $0.651\pm0.013$ &  $\bm{0.637\pm0.015}$ & &  $0.658\pm0.013$ &       $0.648\pm0.012$ &   $\bm{0.632\pm0.015}$ \\
\textbf{PD1-WMT }      &       $0.454\pm0.097$ &        $0.402\pm0.032$ &  $\bm{0.357\pm0.029}$ & &  $0.403\pm0.038$ &        $0.384\pm0.02$ &   $\bm{0.347\pm0.025}$ \\
\\

{} & \multicolumn{7}{c}{\textbf{Good Prior}} \\
\\
{} &                    RS &                     HB &             PriorBand &        &        RS &                    HB &              PriorBand \\ \cline{2-4} \cline{6-8}

\textbf{JAHS-C10 }     &      $11.454\pm1.935$ &       $11.129\pm1.256$ &  $\bm{9.986\pm0.369}$ & & $10.451\pm1.055$ &      $10.077\pm0.769$ &   $\bm{9.457\pm0.415}$ \\
\textbf{JAHS-CH }      &       $6.752\pm0.943$ &         $6.332\pm1.12$ &  $\bm{5.974\pm0.535}$ & &  $5.977\pm0.646$ &      $5.704\pm0.659$ &   $\bm{5.589\pm0.413}$ \\
\textbf{JAHS-FM }      &       $5.497\pm0.396$ &        $5.846\pm0.823$ &  $\bm{5.037\pm0.021}$ & &  $5.282\pm0.299$ &       $5.306\pm0.267$ &   $\bm{5.026\pm0.044}$ \\
\textbf{LC-126026 }    &       $0.053\pm0.011$ &         $0.046\pm0.01$ &  $\bm{0.045\pm0.006}$ & &  $0.048\pm0.008$ &       $0.044\pm0.008$ &   $\bm{0.043\pm0.007}$ \\
\textbf{LC-167190 }    &       $0.214\pm0.021$ &   $\bm{0.193\pm0.025}$ &       $0.198\pm0.026$ & &     $0.2\pm0.02$ &       $0.187\pm0.023$ &   $\bm{0.186\pm0.026}$ \\
\textbf{LC-168330 }    &       $0.444\pm0.046$ &    $\bm{0.41\pm0.035}$ &       $0.411\pm0.024$ & &  $0.406\pm0.038$ &       $0.397\pm0.032$ &   $\bm{0.387\pm0.035}$ \\
\textbf{LC-168910 }    &        $0.38\pm0.093$ &        $0.314\pm0.021$ &  $\bm{0.313\pm0.019}$ & &  $0.324\pm0.032$ &       $0.308\pm0.018$ &   $\bm{0.295\pm0.023}$ \\
\textbf{LC-189906 }    &        $0.222\pm0.09$ &        $0.151\pm0.021$ &  $\bm{0.147\pm0.013}$ & &  $0.166\pm0.033$ &       $0.145\pm0.019$ &   $\bm{0.134\pm0.013}$ \\
\textbf{PD1-Cifar100 } &       $0.378\pm0.156$ &        $0.338\pm0.091$ &  $\bm{0.258\pm0.002}$ & &  $0.302\pm0.074$ &        $0.29\pm0.039$ &   $\bm{0.248\pm0.013}$ \\
\textbf{PD1-ImageNet } &       $0.333\pm0.108$ &        $0.306\pm0.041$ &  $\bm{0.224\pm0.001}$ & &  $0.282\pm0.049$ &       $0.268\pm0.022$ &   $\bm{0.219\pm0.008}$ \\
\textbf{PD1-LM1B }     &       $0.681\pm0.034$ &        $0.651\pm0.013$ &  $\bm{0.647\pm0.008}$ & &  $0.658\pm0.013$ &       $0.648\pm0.012$ &   $\bm{0.642\pm0.009}$ \\
\textbf{PD1-WMT }      &       $0.454\pm0.097$ &        $0.402\pm0.032$ &   $\bm{0.37\pm0.008}$ & &  $0.403\pm0.038$ &        $0.384\pm0.02$ &   $\bm{0.367\pm0.009}$ \\
\\

{} & \multicolumn{7}{c}{\textbf{Bad Prior}} \\
\\
{} &                    RS &                     HB &             PriorBand &        &        RS &                    HB &              PriorBand \\  \cline{2-4} \cline{6-8}
\textbf{JAHS-C10 }      &      $11.454\pm1.935$ &  $\bm{11.129\pm1.256}$ &      $11.729\pm1.795$ & & $10.451\pm1.055$ &      $10.077\pm0.769$ &  $\bm{10.075\pm0.777}$ \\
\textbf{JAHS-CH }       &       $6.752\pm0.943$ &    $\bm{6.332\pm1.12}$ &       $6.553\pm1.284$ & &  $5.977\pm0.646$ &       $5.704\pm0.659$ &   $\bm{5.551\pm0.516}$ \\
\textbf{JAHS-FM }       &  $\bm{5.497\pm0.396}$ &        $5.846\pm0.823$ &       $5.962\pm0.973$ & &  $5.282\pm0.299$ &       $5.306\pm0.267$ &   $\bm{5.277\pm0.248}$ \\
\textbf{LC-126026 }     &       $0.053\pm0.011$ &    $\bm{0.046\pm0.01}$ &       $0.051\pm0.011$ & &   $0.048\pm0.008$ &  $\bm{0.044\pm0.008}$ &         $0.045\pm0.01$ \\
\textbf{LC-167190 }     &       $0.214\pm0.021$ &   $\bm{0.193\pm0.025}$ &       $0.196\pm0.024$ & &     $0.2\pm0.02$ &       $0.187\pm0.023$ &   $\bm{0.183\pm0.026}$ \\
\textbf{LC-168330 }     &       $0.444\pm0.046$ &    $\bm{0.41\pm0.035}$ &       $0.424\pm0.035$ & &  $0.406\pm0.038$ &  $\bm{0.397\pm0.032}$ &        $0.401\pm0.042$ \\
\textbf{LC-168910 }     &        $0.38\pm0.093$ &        $0.314\pm0.021$ &  $\bm{0.309\pm0.019}$ & &   $0.324\pm0.032$ &       $0.308\pm0.018$ &   $\bm{0.297\pm0.016}$ \\
\textbf{LC-189906 }     &        $0.222\pm0.09$ &   $\bm{0.151\pm0.021}$ &        $0.167\pm0.03$ & &  $0.166\pm0.033$ &  $\bm{0.145\pm0.019}$ &         $0.15\pm0.027$ \\
\textbf{PD1-Cifar100 }  &       $0.378\pm0.156$ &   $\bm{0.338\pm0.091}$ &         $0.473\pm0.2$ & &  $0.302\pm0.074$ &   $\bm{0.29\pm0.039}$ &        $0.297\pm0.058$ \\
\textbf{PD1-ImageNet }  &       $0.333\pm0.108$ &   $\bm{0.306\pm0.041}$ &        $0.315\pm0.04$ & &   $0.282\pm0.049$ &  $\bm{0.268\pm0.022}$ &          $0.28\pm0.03$ \\
\textbf{PD1-LM1B }      &       $0.681\pm0.034$ &   $\bm{0.651\pm0.013}$ &       $0.656\pm0.014$ & &  $0.658\pm0.013$ &  $\bm{0.648\pm0.012}$ &        $0.648\pm0.013$ \\
\textbf{PD1-WMT }       &       $0.454\pm0.097$ &   $\bm{0.402\pm0.032}$ &       $0.425\pm0.057$ & &   $0.403\pm0.038$ &        $0.384\pm0.02$ &   $\bm{0.383\pm0.021}$ \\

\bottomrule
\end{tabular}
}
\end{center}
\end{table}

\begin{table}
\caption{\protect Table comparing Asynchronous Successive Halving (ASHA), Asynchronous HyperBand (AsyncHB) and \algo{} final validation errors of the current incumbent at the highest fidelity at $2$ budget horizons of $1\times$ and $5\times$. 
Runs are averaged over $10$ seeds where each run is with $4$ workers.
Unlike the other $2$ algorithms, \algo{} does not performance ASHA-like asynchronous promotions to minimize idle workers.
\algo{} simply starts the next \sh{} bracket if a worker is free.
The table shows that \algo{} can maintain its robust performance when run in a parallel setting while being competitive to related asynchronous algorithms.}
\label{table:mssg-async-table-max_fidelity_loss}
\begin{center}
\scalebox{0.55}{
\centering\begin{tabular}{l ccc c ccc}
\toprule
\textbf{Benchmark}  & \multicolumn{3}{c}{1x} & & \multicolumn{3}{c}{5x} \\ \cline{2-4} \cline{6-8}

\\
 & \multicolumn{7}{c}{\textbf{Near Optimum Prior}} \\
 \\
{} &                  ASHA &               AsyncHB &              PriorBand &     &             ASHA &          AsyncHB &             PriorBand \\ \cline{2-4} \cline{6-8}

\textbf{JAHS-C10} &      $11.382\pm2.523$ &      $12.192\pm1.558$ &   $\bm{8.281\pm0.221}$ & &       $9.739\pm0.716$ &  $9.901\pm0.598$ &  $\bm{8.281\pm0.221}$ \\
\textbf{JAHS-CH} &       $6.325\pm1.116$ &       $6.771\pm1.296$ &   $\bm{4.466\pm0.188}$ & &       $5.492\pm0.362$ &  $5.565\pm0.342$ &  $\bm{4.399\pm0.143}$ \\
\textbf{JAHS-FM} &       $6.627\pm3.384$ &       $6.245\pm0.999$ &   $\bm{4.713\pm0.052}$ & &       $5.127\pm0.235$ &  $5.134\pm0.274$ &  $\bm{4.709\pm0.049}$ \\
\textbf{PD1-Cifar100} &       $0.411\pm0.152$ &       $0.434\pm0.172$ &   $\bm{0.243\pm0.056}$ & &        $0.271\pm0.04$ &  $0.279\pm0.032$ &  $\bm{0.221\pm0.043}$ \\
\textbf{PD1-ImageNet} &       $0.366\pm0.049$ &       $0.333\pm0.056$ &   $\bm{0.259\pm0.059}$ & &       $0.265\pm0.019$ &  $0.255\pm0.011$ &  $\bm{0.218\pm0.021}$ \\
\textbf{PD1-LM1B} &       $0.665\pm0.016$ &       $0.663\pm0.015$ &   $\bm{0.643\pm0.027}$ & &       $0.649\pm0.008$ &  $0.655\pm0.016$ &  $\bm{0.628\pm0.008}$ \\
\textbf{PD1-WMT} &        $0.446\pm0.09$ &        $0.443\pm0.06$ &   $\bm{0.362\pm0.033}$ & &       $0.396\pm0.026$ &  $0.402\pm0.027$ &  $\bm{0.349\pm0.025}$ \\

\\
{} & \multicolumn{7}{c}{\textbf{Good Prior}} \\
\\
{} &                  ASHA &               AsyncHB &              PriorBand &        &          ASHA &          AsyncHB &             PriorBand \\ \cline{2-4} \cline{6-8}

\textbf{JAHS-C10} &      $11.382\pm2.523$ &      $12.192\pm1.558$ &    $\bm{10.194\pm0.0}$ & &       $9.739\pm0.716$ &  $9.901\pm0.598$ &   $\bm{9.413\pm0.42}$ \\
\textbf{JAHS-CH} &  $\bm{6.325\pm1.116}$ &       $6.771\pm1.296$ &          $6.603\pm0.0$ & &       $5.492\pm0.362$ &  $5.565\pm0.342$ &  $\bm{5.238\pm0.325}$ \\
\textbf{JAHS-FM} &       $6.627\pm3.384$ &       $6.184\pm1.009$ &     $\bm{5.042\pm0.0}$ & &       $5.127\pm0.235$ &  $5.126\pm0.211$ &   $\bm{5.024\pm0.04}$ \\
\textbf{PD1-Cifar100} &       $0.411\pm0.152$ &       $0.434\pm0.172$ &     $\bm{0.259\pm0.0}$ & &        $0.271\pm0.04$ &  $0.279\pm0.032$ &  $\bm{0.237\pm0.016}$ \\
\textbf{PD1-ImageNet} &       $0.366\pm0.049$ &       $0.333\pm0.056$ &     $\bm{0.224\pm0.0}$ & &       $0.265\pm0.019$ &  $0.254\pm0.011$ &  $\bm{0.217\pm0.008}$ \\
\textbf{PD1-LM1B} &       $0.665\pm0.016$ &       $0.665\pm0.013$ &    $\bm{0.65\pm0.002}$ & &       $0.649\pm0.008$ &  $0.655\pm0.013$ &  $\bm{0.638\pm0.013}$ \\
\textbf{PD1-WMT} &        $0.446\pm0.09$ &        $0.45\pm0.065$ &     $\bm{0.372\pm0.0}$ & &       $0.396\pm0.026$ &  $0.398\pm0.019$ &  $\bm{0.356\pm0.014}$ \\

\\
{} & \multicolumn{7}{c}{\textbf{Bad Prior}} \\
\\
{} &                  ASHA &               AsyncHB &              PriorBand &      &            ASHA &          AsyncHB &             PriorBand \\ \cline{2-4} \cline{6-8}

\textbf{JAHS-C10} &      $11.382\pm2.523$ &       $12.135\pm1.62$ &  $\bm{11.055\pm1.768}$ & &  $\bm{9.739\pm0.716}$ &  $9.869\pm0.586$ &       $9.806\pm0.632$ \\
\textbf{JAHS-CH} &  $\bm{6.325\pm1.116}$ &       $6.771\pm1.296$ &        $6.622\pm0.877$ & &       $5.492\pm0.362$ &  $5.565\pm0.342$ &  $\bm{5.316\pm0.667}$ \\
\textbf{JAHS-FM} &       $6.627\pm3.384$ &       $6.184\pm1.009$ &   $\bm{5.715\pm0.485}$ & &  $\bm{5.127\pm0.235}$ &  $5.175\pm0.274$ &       $5.333\pm0.332$ \\
\textbf{PD1-Cifar100} &  $\bm{0.411\pm0.152}$ &       $0.434\pm0.172$ &        $0.657\pm0.266$ & &        $0.271\pm0.04$ &  $0.279\pm0.032$ &  $\bm{0.256\pm0.036}$ \\
\textbf{PD1-ImageNet} &       $0.366\pm0.049$ &       $0.335\pm0.058$ &   $\bm{0.324\pm0.062}$ & &       $0.265\pm0.019$ &  $0.258\pm0.015$ &  $\bm{0.257\pm0.027}$ \\
\textbf{PD1-LM1B} &       $0.665\pm0.016$ &  $\bm{0.665\pm0.013}$ &        $0.668\pm0.016$ & &       $0.649\pm0.008$ &  $0.654\pm0.013$ &  $\bm{0.646\pm0.012}$ \\
\textbf{PD1-WMT} &   $\bm{0.446\pm0.09}$ &        $0.45\pm0.065$ &        $0.485\pm0.141$ & &       $0.396\pm0.026$ &  $0.398\pm0.019$ &  $\bm{0.375\pm0.014}$ \\

\bottomrule
\end{tabular}
}
\end{center}
\end{table}

\begin{table}
\caption{\protect Table comparing Asynchronous Successive Halving (ASHA), Asynchronous HyperBand (AsyncHB), ASHA+\E{} and Asynchronous-HyperBand+\E{} final validation errors of the current incumbent at the highest fidelity at $2$ budget horizons of $1\times$ and $5\times$. 
Runs are averaged over $10$ seeds where each run is with $4$ workers.
This table showcases the flexibility of the \esp{}, as it can be applied off-the-shelf to other multi-fidelity algorithms too.
Under informative priors, ASHA(+\esp{}) performs marginally better than Asynchronous-\hb{}(+\esp{}) but wanes for longer budgets.
Interestingly, ASHA(+\esp{}) retains a better performance under the bad priors.
Since ASHA(+\esp{}) samples only at the \rung{}$=0$, the \esp{} fixes \pu{} to $50\%$.
This could explain the increased exploration under bad priors and reduced exploitation under the good priors for ASHA(+\esp{}).}
\label{table:exp-async-hb-table-max_fidelity_loss}
\begin{center}
\scalebox{0.55}{
\centering\begin{tabular}{l cccc c cccc}
\toprule
\textbf{Benchmark}  & \multicolumn{4}{c}{1x} & & \multicolumn{4}{c}{5x} \\ \cline{2-5} \cline{7-10}

\\
 & \multicolumn{9}{c}{\textbf{Near Optimum Prior}} \\
 \\
  {} &                   ASHA &               AsyncHB &              ASHA(+ESP) &          Async-HB(+ESP) & &                  ASHA &               AsyncHB &              ASHA(+ESP) &          Async-HB(+ESP) \\ \cline{2-5} \cline{7-10}

\textbf{JAHS-C10} &       $11.382\pm2.523$ &      $12.192\pm1.558$ &  $\bm{8.281\pm0.221}$ &       $8.281\pm0.221$ & &       $9.739\pm0.716$ &       $9.901\pm0.598$ &       $8.223\pm0.183$ &   $\bm{8.18\pm0.177}$ \\
\textbf{JAHS-CH} &        $6.325\pm1.116$ &       $6.771\pm1.296$ &  $\bm{4.466\pm0.188}$ &       $4.466\pm0.188$ & &       $5.492\pm0.362$ &       $5.565\pm0.342$ &  $\bm{4.466\pm0.188}$ &       $4.466\pm0.188$ \\
\textbf{JAHS-FM} &        $6.627\pm3.384$ &       $6.245\pm0.999$ &  $\bm{4.713\pm0.052}$ &       $4.713\pm0.052$ & &       $5.127\pm0.235$ &       $5.134\pm0.274$ &       $4.709\pm0.049$ &   $\bm{4.707\pm0.05}$ \\
\textbf{PD1-Cifar100} &        $0.411\pm0.152$ &       $0.434\pm0.172$ &  $\bm{0.243\pm0.056}$ &       $0.259\pm0.083$ & &        $0.271\pm0.04$ &       $0.279\pm0.032$ &  $\bm{0.221\pm0.037}$ &        $0.227\pm0.04$ \\
\textbf{PD1-ImageNet} &        $0.366\pm0.049$ &       $0.333\pm0.056$ &        $0.269\pm0.07$ &   $\bm{0.261\pm0.06}$ & &       $0.265\pm0.019$ &       $0.255\pm0.011$ &        $0.22\pm0.021$ &  $\bm{0.214\pm0.017}$ \\
\textbf{PD1-LM1B} &        $0.665\pm0.016$ &       $0.663\pm0.015$ &       $0.644\pm0.025$ &  $\bm{0.639\pm0.018}$ & &       $0.649\pm0.008$ &       $0.655\pm0.016$ &        $0.629\pm0.01$ &  $\bm{0.628\pm0.009}$ \\
\textbf{PD1-WMT} &         $0.446\pm0.09$ &        $0.443\pm0.06$ &  $\bm{0.364\pm0.036}$ &       $0.373\pm0.053$ & &       $0.396\pm0.026$ &       $0.402\pm0.027$ &  $\bm{0.344\pm0.035}$ &       $0.353\pm0.026$ \\

\\
 & \multicolumn{9}{c}{\textbf{Good Prior}} \\
 \\
  {} &                   ASHA &               AsyncHB &              ASHA(+ESP) &          Async-HB(+ESP) & &                  ASHA &               AsyncHB &              ASHA(+ESP) &          Async-HB(+ESP) \\ \cline{2-5} \cline{7-10}

\textbf{JAHS-C10} &       $11.382\pm2.523$ &      $12.192\pm1.558$ &   $\bm{10.194\pm0.0}$ &        $10.194\pm0.0$ & &       $9.739\pm0.716$ &       $9.901\pm0.598$ &       $9.603\pm0.412$ &   $\bm{9.365\pm0.51}$ \\
\textbf{JAHS-CH} &   $\bm{6.325\pm1.116}$ &       $6.771\pm1.296$ &         $6.603\pm0.0$ &         $6.603\pm0.0$ & &       $5.492\pm0.362$ &       $5.565\pm0.342$ &        $5.592\pm0.31$ &  $\bm{5.386\pm0.327}$ \\
\textbf{JAHS-FM} &        $6.627\pm3.384$ &       $6.184\pm1.009$ &    $\bm{5.042\pm0.0}$ &         $5.042\pm0.0$ & &       $5.127\pm0.235$ &       $5.126\pm0.211$ &  $\bm{4.995\pm0.088}$ &        $5.013\pm0.06$ \\
\textbf{PD1-Cifar100} &        $0.411\pm0.152$ &       $0.434\pm0.172$ &    $\bm{0.259\pm0.0}$ &         $0.259\pm0.0$ & &        $0.271\pm0.04$ &       $0.279\pm0.032$ &       $0.241\pm0.014$ &   $\bm{0.24\pm0.017}$ \\
\textbf{PD1-ImageNet} &        $0.366\pm0.049$ &       $0.333\pm0.056$ &    $\bm{0.224\pm0.0}$ &         $0.224\pm0.0$ & &       $0.265\pm0.019$ &       $0.254\pm0.011$ &        $0.22\pm0.005$ &  $\bm{0.217\pm0.005}$ \\
\textbf{PD1-LM1B} &        $0.665\pm0.016$ &       $0.665\pm0.013$ &  $\bm{0.649\pm0.004}$ &         $0.651\pm0.0$ & &       $0.649\pm0.008$ &       $0.655\pm0.013$ &  $\bm{0.646\pm0.006}$ &       $0.646\pm0.005$ \\
\textbf{PD1-WMT} &         $0.446\pm0.09$ &        $0.45\pm0.065$ &    $\bm{0.372\pm0.0}$ &         $0.372\pm0.0$ & &       $0.396\pm0.026$ &       $0.398\pm0.019$ &       $0.371\pm0.004$ &  $\bm{0.368\pm0.009}$ \\

\\
 & \multicolumn{9}{c}{\textbf{Bad Prior}} \\
 \\
  {} &                   ASHA &               AsyncHB &              ASHA(+ESP) &          Async-HB(+ESP) & &                  ASHA &               AsyncHB &              ASHA(+ESP) &          Async-HB(+ESP) \\ \cline{2-5} \cline{7-10}

\textbf{JAHS-C10} &  $\bm{11.382\pm2.523}$ &       $12.135\pm1.62$ &      $12.697\pm2.049$ &      $13.857\pm3.841$ & &       $9.739\pm0.716$ &       $9.869\pm0.586$ &  $\bm{9.669\pm0.502}$ &       $9.809\pm0.774$ \\
\textbf{JAHS-CH} &   $\bm{6.325\pm1.116}$ &       $6.771\pm1.296$ &       $6.778\pm0.904$ &       $9.416\pm2.559$ & &  $\bm{5.492\pm0.362}$ &       $5.565\pm0.342$ &       $5.856\pm0.401$ &       $5.506\pm0.703$ \\
\textbf{JAHS-FM} &        $6.627\pm3.384$ &  $\bm{6.184\pm1.009}$ &       $7.091\pm3.287$ &       $7.244\pm3.419$ & &       $5.127\pm0.235$ &       $5.175\pm0.274$ &  $\bm{5.113\pm0.299}$ &       $5.134\pm0.264$ \\
\textbf{PD1-Cifar100} &   $\bm{0.411\pm0.152}$ &       $0.434\pm0.172$ &       $0.479\pm0.226$ &       $0.665\pm0.186$ & &   $\bm{0.271\pm0.04}$ &       $0.279\pm0.032$ &        $0.29\pm0.041$ &        $0.293\pm0.02$ \\
\textbf{PD1-ImageNet} &        $0.366\pm0.049$ &  $\bm{0.335\pm0.058}$ &       $0.401\pm0.177$ &       $0.558\pm0.155$ & &       $0.265\pm0.019$ &  $\bm{0.258\pm0.015}$ &       $0.291\pm0.098$ &       $0.288\pm0.052$ \\
\textbf{PD1-LM1B} &        $0.665\pm0.016$ &  $\bm{0.665\pm0.013}$ &       $0.687\pm0.046$ &       $0.705\pm0.042$ & &       $0.649\pm0.008$ &       $0.654\pm0.013$ &  $\bm{0.647\pm0.018}$ &       $0.652\pm0.012$ \\
\textbf{PD1-WMT} &    $\bm{0.446\pm0.09}$ &        $0.45\pm0.065$ &        $0.54\pm0.172$ &       $0.621\pm0.152$ & &       $0.396\pm0.026$ &       $0.398\pm0.019$ &  $\bm{0.383\pm0.022}$ &       $0.395\pm0.029$ \\
\bottomrule
\end{tabular}
}
\end{center}
\end{table}

\begin{table}
\caption{\protect Table comparing Bayesian Optimization (BO), \pibo{}, BOHB and \algo{}+BO final validation errors of the current incumbent at the highest fidelity at $2$ budget horizons of $10\times$ and $20\times$. 
Runs are averaged over $50$ seeds on $1$ worker each. 
BO, \pibo{} and \algo{}+BO use an initial design size of $10$.
This table highlights the model extensibility of \algo{}.
Under near-optimum priors, \pibo{} emerges as the best, especially after $10\times$.
Knowledge of a near-optimum region is not available in practice and \pibo{}'s rate of recovery from bad priors is costlier for DL.
\algo{}+BO importantly is better than BOHB in almost all settings and better than vanilla-BO. }
\label{table:table-bo-table-max_fidelity_loss}

\begin{center}
\scalebox{0.55}{
\centering\begin{tabular}{l cccc c cccc}
\toprule
\textbf{Benchmark}  & \multicolumn{4}{c}{10x} & & \multicolumn{4}{c}{20x} \\ \cline{2-5} \cline{7-10}

\\
 & \multicolumn{9}{c}{\textbf{Near Optimum Prior}} \\
 \\
{} &                     BO &                   \pibo{} &                  BOHB &          PriorBand+BO & &                    BO &                   \pibo{} &                  BOHB &          PriorBand+BO \\  \cline{2-5} \cline{7-10}

\textbf{JAHS-C10} &        $9.723\pm0.949$ &   $\bm{8.23\pm0.229}$ &       $10.281\pm0.88$ &       $8.252\pm0.215$ & &       $9.115\pm0.725$ &   $\bm{8.07\pm0.125}$ &       $9.969\pm0.688$ &        $8.248\pm0.21$ \\
\textbf{JAHS-CH} &        $5.445\pm0.493$ &       $4.481\pm0.112$ &       $5.628\pm0.613$ &  $\bm{4.452\pm0.127}$ & &       $4.964\pm0.426$ &  $\bm{4.402\pm0.089}$ &       $5.393\pm0.442$ &       $4.452\pm0.127$ \\
\textbf{JAHS-FM} &        $5.097\pm0.223$ &   $\bm{4.71\pm0.048}$ &       $5.406\pm0.323$ &        $4.724\pm0.06$ & &       $4.917\pm0.172$ &  $\bm{4.687\pm0.042}$ &       $5.227\pm0.255$ &        $4.724\pm0.06$ \\
\textbf{LC-126026} &        $0.042\pm0.007$ &       $0.026\pm0.006$ &       $0.043\pm0.009$ &  $\bm{0.024\pm0.007}$ & &       $0.031\pm0.008$ &  $\bm{0.017\pm0.006}$ &       $0.038\pm0.008$ &       $0.023\pm0.005$ \\
\textbf{LC-167190} &        $0.188\pm0.021$ &       $0.138\pm0.014$ &       $0.191\pm0.022$ &  $\bm{0.136\pm0.016}$ & &       $0.164\pm0.023$ &  $\bm{0.119\pm0.012}$ &       $0.175\pm0.023$ &       $0.134\pm0.014$ \\
\textbf{LC-168330} &         $0.38\pm0.043$ &   $\bm{0.289\pm0.03}$ &       $0.412\pm0.032$ &       $0.295\pm0.044$ & &       $0.322\pm0.044$ &  $\bm{0.258\pm0.024}$ &       $0.395\pm0.033$ &       $0.281\pm0.035$ \\
\textbf{LC-168910} &        $0.294\pm0.033$ &  $\bm{0.198\pm0.021}$ &        $0.31\pm0.033$ &         $0.2\pm0.021$ & &       $0.266\pm0.042$ &  $\bm{0.171\pm0.022}$ &       $0.297\pm0.025$ &          $0.2\pm0.02$ \\
\textbf{LC-189906} &         $0.14\pm0.021$ &       $0.112\pm0.011$ &       $0.141\pm0.014$ &  $\bm{0.111\pm0.012}$ & &       $0.117\pm0.015$ &  $\bm{0.101\pm0.008}$ &        $0.13\pm0.011$ &       $0.106\pm0.011$ \\
\textbf{PD1-Cifar100} &        $0.284\pm0.041$ &       $0.239\pm0.035$ &       $0.291\pm0.039$ &  $\bm{0.238\pm0.035}$ & &       $0.238\pm0.022$ &   $\bm{0.224\pm0.03}$ &        $0.262\pm0.03$ &       $0.227\pm0.029$ \\
\textbf{PD1-ImageNet} &        $0.261\pm0.033$ &   $\bm{0.222\pm0.02}$ &       $0.289\pm0.035$ &       $0.238\pm0.035$ & &        $0.234\pm0.02$ &  $\bm{0.215\pm0.017}$ &       $0.278\pm0.029$ &       $0.228\pm0.028$ \\
\textbf{PD1-LM1B} &         $0.66\pm0.015$ &        $0.644\pm0.02$ &       $0.643\pm0.014$ &  $\bm{0.637\pm0.014}$ & &       $0.646\pm0.014$ &  $\bm{0.631\pm0.018}$ &       $0.634\pm0.014$ &       $0.633\pm0.012$ \\
\textbf{PD1-WMT} &        $0.387\pm0.026$ &  $\bm{0.349\pm0.025}$ &       $0.382\pm0.022$ &       $0.353\pm0.027$ & &       $0.362\pm0.021$ &  $\bm{0.335\pm0.022}$ &       $0.363\pm0.017$ &       $0.346\pm0.021$ \\

\\
 & \multicolumn{9}{c}{\textbf{Good Prior}} \\
 \\
{} &                     BO &                   \pibo{} &                  BOHB &          PriorBand+BO &   &                 BO &                   \pibo{} &                  BOHB &          PriorBand+BO \\  \cline{2-5} \cline{7-10}

\textbf{JAHS-C10} &       $10.049\pm0.983$ &   $\bm{9.52\pm0.427}$ &       $10.281\pm0.88$ &        $9.887\pm0.39$ & &       $9.451\pm0.806$ &  $\bm{9.222\pm0.287}$ &       $9.969\pm0.688$ &       $9.706\pm0.416$ \\
\textbf{JAHS-CH} &        $5.743\pm0.482$ &       $5.632\pm0.323$ &       $5.628\pm0.613$ &    $\bm{5.56\pm0.51}$ & &  $\bm{5.158\pm0.387}$ &       $5.439\pm0.256$ &       $5.393\pm0.442$ &       $5.393\pm0.417$ \\
\textbf{JAHS-FM} &        $5.257\pm0.245$ &       $5.038\pm0.015$ &       $5.406\pm0.323$ &  $\bm{4.996\pm0.091}$ & &       $5.047\pm0.205$ &       $4.965\pm0.063$ &       $5.227\pm0.255$ &  $\bm{4.964\pm0.114}$ \\
\textbf{LC-126026} &        $0.049\pm0.007$ &       $0.044\pm0.003$ &       $0.043\pm0.009$ &  $\bm{0.039\pm0.006}$ & &        $0.04\pm0.008$ &        $0.04\pm0.003$ &       $0.038\pm0.008$ &  $\bm{0.036\pm0.006}$ \\
\textbf{LC-167190} &        $0.204\pm0.019$ &       $0.209\pm0.017$ &       $0.191\pm0.022$ &  $\bm{0.184\pm0.027}$ & &       $0.175\pm0.023$ &       $0.203\pm0.018$ &       $0.175\pm0.023$ &  $\bm{0.169\pm0.026}$ \\
\textbf{LC-168330} &        $0.412\pm0.032$ &       $0.404\pm0.029$ &       $0.412\pm0.032$ &  $\bm{0.374\pm0.051}$ & &        $0.361\pm0.04$ &       $0.366\pm0.039$ &       $0.395\pm0.033$ &  $\bm{0.344\pm0.052}$ \\
\textbf{LC-168910} &        $0.318\pm0.028$ &       $0.313\pm0.024$ &        $0.31\pm0.033$ &  $\bm{0.302\pm0.032}$ & &        $0.295\pm0.03$ &       $0.301\pm0.022$ &       $0.297\pm0.025$ &  $\bm{0.289\pm0.035}$ \\
\textbf{LC-189906} &         $0.153\pm0.02$ &       $0.148\pm0.011$ &       $0.141\pm0.014$ &  $\bm{0.126\pm0.016}$ & &       $0.124\pm0.013$ &       $0.141\pm0.009$ &        $0.13\pm0.011$ &  $\bm{0.115\pm0.014}$ \\
\textbf{PD1-Cifar100} &        $0.282\pm0.039$ &        $0.25\pm0.014$ &       $0.291\pm0.039$ &  $\bm{0.248\pm0.015}$ & &       $0.242\pm0.028$ &   $\bm{0.23\pm0.017}$ &        $0.262\pm0.03$ &       $0.234\pm0.016$ \\
\textbf{PD1-ImageNet} &        $0.266\pm0.029$ &  $\bm{0.223\pm0.005}$ &       $0.289\pm0.035$ &       $0.223\pm0.003$ & &       $0.237\pm0.017$ &  $\bm{0.213\pm0.009}$ &       $0.278\pm0.029$ &        $0.22\pm0.008$ \\
\textbf{PD1-LM1B} &        $0.663\pm0.016$ &        $0.65\pm0.003$ &       $0.643\pm0.014$ &  $\bm{0.641\pm0.011}$ & &       $0.648\pm0.013$ &       $0.646\pm0.006$ &  $\bm{0.634\pm0.014}$ &       $0.636\pm0.011$ \\
\textbf{PD1-WMT} &        $0.391\pm0.018$ &       $0.372\pm0.003$ &       $0.382\pm0.022$ &  $\bm{0.367\pm0.014}$ & &       $0.372\pm0.014$ &       $0.368\pm0.009$ &       $0.363\pm0.017$ &  $\bm{0.356\pm0.018}$ \\

\\
 & \multicolumn{9}{c}{\textbf{Bad Prior}} \\
 \\
{} &                     BO &                   \pibo{} &                  BOHB &          PriorBand+BO &        &            BO &                   \pibo{} &                  BOHB &          PriorBand+BO \\  \cline{2-5} \cline{7-10}

\textbf{JAHS-C10} &  $\bm{10.132\pm1.143}$ &       $10.24\pm1.242$ &       $10.281\pm0.88$ &      $10.213\pm0.846$ & &  $\bm{9.354\pm0.722}$ &       $9.534\pm0.857$ &       $9.969\pm0.688$ &       $9.716\pm0.597$ \\
\textbf{JAHS-CH} &        $5.874\pm0.803$ &       $6.572\pm1.379$ &       $5.628\pm0.613$ &  $\bm{5.608\pm0.582}$ & &  $\bm{5.316\pm0.537}$ &       $6.204\pm1.057$ &       $5.393\pm0.442$ &       $5.429\pm0.531$ \\
\textbf{JAHS-FM} &         $5.323\pm0.38$ &       $5.425\pm0.343$ &       $5.406\pm0.323$ &  $\bm{5.261\pm0.291}$ & &  $\bm{5.088\pm0.227}$ &       $5.212\pm0.276$ &       $5.227\pm0.255$ &       $5.138\pm0.199$ \\
\textbf{LC-126026} &        $0.051\pm0.007$ &       $0.059\pm0.009$ &       $0.043\pm0.009$ &   $\bm{0.039\pm0.01}$ & &       $0.042\pm0.008$ &       $0.051\pm0.005$ &       $0.038\pm0.008$ &   $\bm{0.034\pm0.01}$ \\
\textbf{LC-167190} &         $0.21\pm0.024$ &       $0.228\pm0.024$ &       $0.191\pm0.022$ &  $\bm{0.185\pm0.028}$ & &       $0.184\pm0.026$ &       $0.216\pm0.019$ &       $0.175\pm0.023$ &  $\bm{0.163\pm0.024}$ \\
\textbf{LC-168330} &        $0.427\pm0.043$ &       $0.463\pm0.042$ &       $0.412\pm0.032$ &   $\bm{0.395\pm0.04}$ & &   $\bm{0.35\pm0.052}$ &        $0.429\pm0.04$ &       $0.395\pm0.033$ &       $0.366\pm0.053$ \\
\textbf{LC-168910} &        $0.323\pm0.028$ &       $0.321\pm0.045$ &        $0.31\pm0.033$ &  $\bm{0.304\pm0.032}$ & &       $0.291\pm0.028$ &        $0.29\pm0.012$ &       $0.297\pm0.025$ &  $\bm{0.288\pm0.029}$ \\
\textbf{LC-189906} &        $0.217\pm0.076$ &       $0.365\pm0.139$ &       $0.141\pm0.014$ &  $\bm{0.126\pm0.019}$ & &       $0.129\pm0.016$ &       $0.185\pm0.057$ &        $0.13\pm0.011$ &  $\bm{0.115\pm0.017}$ \\
\textbf{PD1-Cifar100} &        $0.408\pm0.155$ &       $0.586\pm0.195$ &       $0.291\pm0.039$ &  $\bm{0.282\pm0.054}$ & &       $0.251\pm0.031$ &       $0.336\pm0.164$ &        $0.262\pm0.03$ &  $\bm{0.241\pm0.021}$ \\
\textbf{PD1-ImageNet} &        $0.355\pm0.135$ &       $0.619\pm0.188$ &       $0.289\pm0.035$ &  $\bm{0.285\pm0.032}$ & &        $0.26\pm0.109$ &        $0.41\pm0.233$ &       $0.278\pm0.029$ &  $\bm{0.249\pm0.025}$ \\
\textbf{PD1-LM1B} &        $0.688\pm0.034$ &        $0.782\pm0.06$ &  $\bm{0.643\pm0.014}$ &       $0.644\pm0.013$ & &       $0.653\pm0.015$ &        $0.667\pm0.03$ &  $\bm{0.634\pm0.014}$ &       $0.636\pm0.012$ \\
\textbf{PD1-WMT} &        $0.523\pm0.124$ &       $0.751\pm0.099$ &       $0.382\pm0.022$ &   $\bm{0.38\pm0.025}$ & &       $0.375\pm0.024$ &       $0.497\pm0.133$ &       $0.363\pm0.017$ &  $\bm{0.361\pm0.018}$ \\

\bottomrule
\end{tabular}
}
\end{center}
\end{table}

\begin{table}
\caption{\protect Table comparing Asynchronous-HyperBand+BO (Mobster), Mobster+\esp{} and \algo{}+BO final validation errors of the current incumbent at the highest fidelity at $2$ budget horizons of $1\times$ and $5\times$. 
Runs are averaged over $10$ seeds where each run is with $4$ workers.
This table shows the extensibility of ESP to asynchronous model-based \hb{} and also verifies the effectiveness of running \algo{} in a parallel setting.
The key differences between Mobster+\esp{} and \algo{}+BO are their choice of initial design and the nature of multi-fidelity scheduling.}
\label{table:async-pi-bo-table-max_fidelity_loss}
\begin{center}
\scalebox{0.55}{
\centering\begin{tabular}{l ccc c ccc}
\toprule
\textbf{Benchmark} & \multicolumn{3}{c}{1x} & & \multicolumn{3}{c}{5x} \\ \cline{2-4} \cline{6-8}

\\
{} & \multicolumn{7}{c}{\textbf{Good Prior}} \\
\\
{} &               Mobster &           Mobster(+ESP) &           PriorBand+BO & &               Mobster &           Mobster(+ESP) &          PriorBand+BO \\ \cline{2-4} \cline{6-8}

\textbf{JAHS-C10} &      $12.192\pm1.558$ &  $\bm{8.281\pm0.221}$ &        $8.281\pm0.221$ & &       $9.872\pm0.579$ &   $\bm{8.18\pm0.177}$ &       $8.281\pm0.221$ \\
\textbf{JAHS-CH} &       $6.771\pm1.296$ &  $\bm{4.466\pm0.188}$ &        $4.466\pm0.188$ & &       $5.565\pm0.342$ &       $4.466\pm0.188$ &  $\bm{4.424\pm0.169}$ \\
\textbf{JAHS-FM} &       $6.172\pm1.015$ &  $\bm{4.713\pm0.052}$ &        $4.713\pm0.052$ & &       $5.177\pm0.279$ &  $\bm{4.713\pm0.052}$ &       $4.713\pm0.052$ \\
\textbf{PD1-Cifar100} &       $0.434\pm0.172$ &       $0.259\pm0.083$ &   $\bm{0.243\pm0.056}$ & &       $0.252\pm0.029$ &  $\bm{0.217\pm0.033}$ &       $0.217\pm0.033$ \\
\textbf{PD1-ImageNet} &       $0.338\pm0.056$ &        $0.261\pm0.06$ &   $\bm{0.258\pm0.053}$ & &       $0.249\pm0.023$ &  $\bm{0.214\pm0.017}$ &       $0.234\pm0.029$ \\
\textbf{PD1-LM1B} &       $0.663\pm0.012$ &  $\bm{0.639\pm0.018}$ &        $0.642\pm0.022$ & &        $0.647\pm0.01$ &       $0.634\pm0.012$ &   $\bm{0.63\pm0.018}$ \\
\textbf{PD1-WMT} &        $0.45\pm0.065$ &        $0.368\pm0.04$ &   $\bm{0.358\pm0.032}$ & &       $0.374\pm0.015$ &  $\bm{0.348\pm0.023}$ &       $0.353\pm0.027$ \\

\\
{} & \multicolumn{7}{c}{\textbf{Good Prior}} \\
\\
{} &               Mobster &           Mobster(+ESP) &           PriorBand+BO & &               Mobster &           Mobster(+ESP) &          PriorBand+BO \\ \cline{2-4} \cline{6-8}

\textbf{JAHS-C10} &      $12.192\pm1.558$ &   $\bm{10.194\pm0.0}$ &         $10.194\pm0.0$ & &       $9.952\pm0.637$ &   $\bm{9.297\pm0.39}$ &       $9.609\pm0.576$ \\
\textbf{JAHS-CH} &       $6.768\pm1.295$ &    $\bm{6.603\pm0.0}$ &          $6.603\pm0.0$ & &       $5.668\pm0.371$ &       $5.492\pm0.312$ &  $\bm{5.321\pm0.553}$ \\
\textbf{JAHS-FM} &       $6.245\pm0.999$ &    $\bm{5.042\pm0.0}$ &          $5.042\pm0.0$ & &       $5.134\pm0.274$ &        $5.013\pm0.06$ &  $\bm{5.008\pm0.063}$ \\
\textbf{PD1-Cifar100} &       $0.434\pm0.172$ &    $\bm{0.259\pm0.0}$ &          $0.259\pm0.0$ & &       $0.252\pm0.029$ &  $\bm{0.226\pm0.017}$ &        $0.234\pm0.02$ \\
\textbf{PD1-ImageNet} &       $0.333\pm0.056$ &    $\bm{0.224\pm0.0}$ &          $0.224\pm0.0$ & &       $0.244\pm0.013$ &  $\bm{0.222\pm0.005}$ &         $0.224\pm0.0$ \\
\textbf{PD1-LM1B} &       $0.663\pm0.012$ &         $0.651\pm0.0$ &   $\bm{0.641\pm0.009}$ & &       $0.649\pm0.009$ &       $0.646\pm0.008$ &  $\bm{0.635\pm0.011}$ \\
\textbf{PD1-WMT} &        $0.45\pm0.065$ &    $\bm{0.372\pm0.0}$ &          $0.372\pm0.0$ & &       $0.377\pm0.017$ &        $0.37\pm0.006$ &   $\bm{0.358\pm0.02}$ \\

\\
{} & \multicolumn{7}{c}{\textbf{Bad Prior}} \\
\\
{} &               Mobster &           Mobster(+ESP) &           PriorBand+BO & &               Mobster &           Mobster(+ESP) &          PriorBand+BO \\ \cline{2-4} \cline{6-8}

\textbf{JAHS-C10} &      $12.375\pm1.474$ &      $13.857\pm3.841$ &  $\bm{10.252\pm1.069}$ & &       $9.792\pm0.624$ &       $9.744\pm0.643$ &  $\bm{9.614\pm0.622}$ \\
\textbf{JAHS-CH} &  $\bm{6.768\pm1.295}$ &       $9.218\pm2.769$ &         $7.291\pm1.25$ & &       $5.668\pm0.371$ &        $5.56\pm0.707$ &  $\bm{5.407\pm0.483}$ \\
\textbf{JAHS-FM} &  $\bm{6.245\pm0.999}$ &       $7.244\pm3.419$ &        $6.845\pm3.545$ & &  $\bm{5.134\pm0.274}$ &       $5.181\pm0.266$ &        $5.17\pm0.227$ \\
\textbf{PD1-Cifar100} &  $\bm{0.434\pm0.172}$ &       $0.678\pm0.195$ &        $0.657\pm0.266$ & &       $0.252\pm0.029$ &       $0.258\pm0.025$ &   $\bm{0.244\pm0.03}$ \\
\textbf{PD1-ImageNet} &       $0.333\pm0.056$ &       $0.558\pm0.155$ &   $\bm{0.328\pm0.035}$ & &  $\bm{0.244\pm0.013}$ &       $0.248\pm0.028$ &       $0.266\pm0.026$ \\
\textbf{PD1-LM1B} &       $0.663\pm0.012$ &       $0.699\pm0.034$ &   $\bm{0.661\pm0.022}$ & &       $0.649\pm0.009$ &       $0.657\pm0.006$ &   $\bm{0.639\pm0.01}$ \\
\textbf{PD1-WMT} &   $\bm{0.45\pm0.065}$ &        $0.609\pm0.16$ &         $0.47\pm0.088$ & &       $0.375\pm0.013$ &       $0.372\pm0.017$ &  $\bm{0.371\pm0.016}$ \\

\bottomrule
\end{tabular}
}
\end{center}
\end{table}

\end{document}